\documentclass{emyxpaper}

\usepackage[utf8]{inputenc}

\usepackage{amsmath}
\usepackage{amsfonts}
\usepackage{amssymb}
\usepackage{mathtools}
\usepackage{bm}

\usepackage{booktabs}
\usepackage{longtable}
\usepackage{multirow}
\usepackage{wrapfig}

\usepackage{url}
\usepackage{nicefrac}
\usepackage[table]{xcolor}
\usepackage{xspace}
\usepackage{enumitem}
\definecolor{editgreen}{RGB}{40,150,60}
\newcommand{\edit}[1]{#1}

\usepackage{algorithm}
\usepackage{algorithmic}
\usepackage{float}
\makeatletter
\newenvironment{algindent}{\begin{ALC@g}}{\end{ALC@g}}
\makeatother

\usetikzlibrary{calc, positioning, arrows.meta, fit}

\newcommand{\emyx}{{Emyx}\xspace}

\newcommand{\proteina}{Prote\'{i}na\xspace}

\newcommand{\proteinacomplexa}{Prote\'{i}na-Complexa\xspace}
\newcommand{\rfdthree}{RFdiffusion3\xspace}
\newcommand{\xt}{\bm{x}_t}

\newcommand{\R}{\mathbb{R}}
\newcommand{\ie}{\emph{i.e.}\xspace}

\newcommand{\angstrom}{\text{\AA}}
\newcommand{\todo}[1]{\textcolor{red}{\textbf{TODO\if\relax\detokenize{#1}\relax\else: #1\fi}}}

\definecolor{figmotif}{HTML}{5B2C8C}    
\definecolor{figflow}{HTML}{1B6B2A}     
\definecolor{figligand}{HTML}{D94E1F}   

\newcommand{\repdir}{figures/representations}
\newcommand{\iconnode}[1]{\raisebox{-0.6ex}{\includegraphics[height=1.2em]{\repdir/#1}}}
\newcommand{\iconpair}[1]{\raisebox{-0.9ex}{\includegraphics[height=3em]{\repdir/#1}}}
\newcommand{\atomrep}{\iconnode{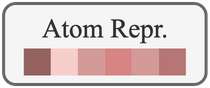}}
\newcommand{\tokenrep}{\iconnode{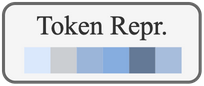}}
\newcommand{\atomcond}{\iconnode{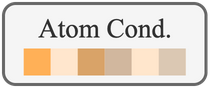}}
\newcommand{\tokencond}{\iconnode{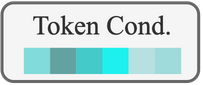}}
\newcommand{\atompair}{\iconpair{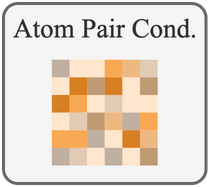}}
\newcommand{\tokenpair}{\iconpair{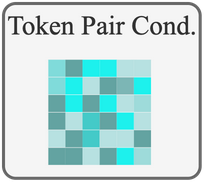}}
\newcommand{\graphrep}{\raisebox{-1.2ex}{\includegraphics[height=2.4em]{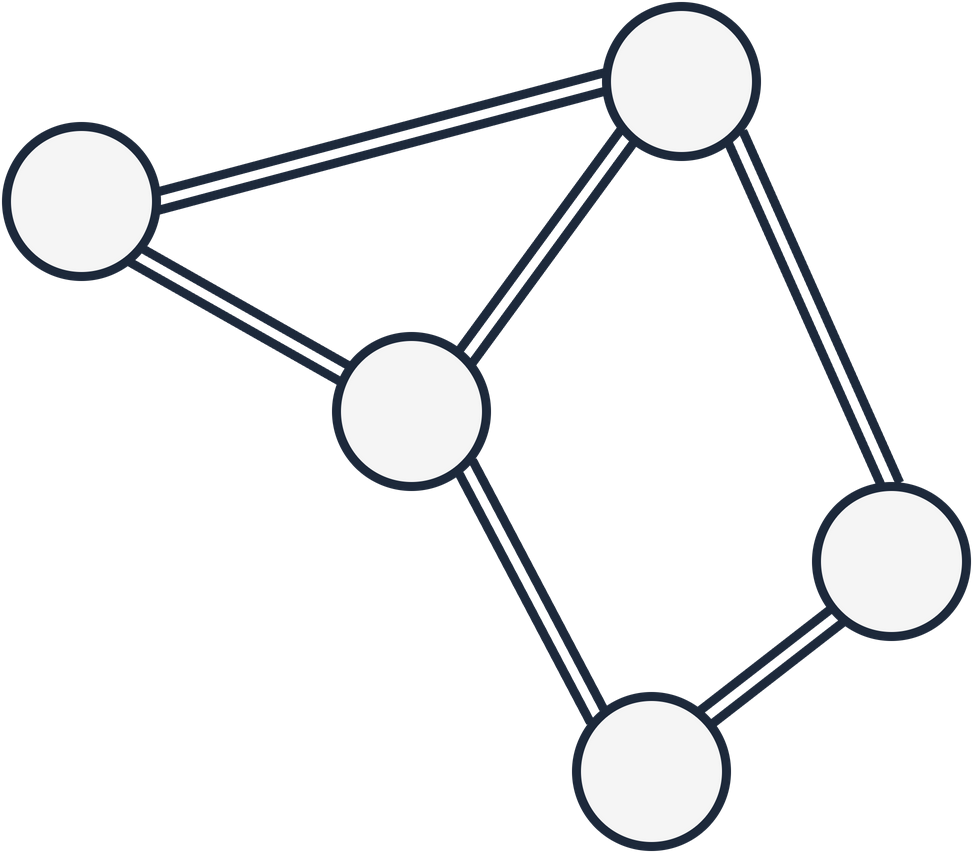}}}

\newenvironment{ack}{\paragraph*{Acknowledgments}}{\par}

\title{Emyx: Fast and efficient all-atom protein generation}
\author[\dagger]{Nicholas J. Williams}
\author[\dagger,*]{Ward Haddadin}
\author{Matteo P. Ferla}
\author{Constantin Schneider}
\author{Nicholas B. Woodall}
\author{Ruby Sedgwick}
\author{Christian D. Madsen}
\author{Andrew L. Hopkins}
\author{Douglas E. V. Pires}
\author{Edward O. Pyzer-Knapp}
\affiliation{Xyme, Oxford, UK}
\contribution[\dagger]{Equal contribution}
\contribution[*]{Correspondence: \email{doug@xyme.ai}}

\abstract{

Computational enzyme design requires generating proteins that scaffold catalytic residues and ligands, a task that demands both geometric accuracy and structural diversity from the underlying generative model. Current all-atom generators inherit expensive architectures from structure prediction, leading to high training costs and limited sample diversity. We argue that much of this complexity is unnecessary for generators, which condition on sparse geometric constraints rather than rich co-evolutionary signals. \emyx is a 140M-parameter conditional flow matching model that concentrates capacity within standard transformer blocks, replacing heavy embedding stacks with lightweight conditional representations and sparse connectivity. We additionally derive an exact reparametrisation of the flow matching interpolant into the EDM noise-level framework, bridging flow matching training efficiency with state-of-the-art sampling methods designed for diffusion models without retraining. Despite being the smallest model, \emyx outperforms both \proteinacomplexa and \rfdthree against the AME enzyme design benchmark across success rate under strict evaluation requiring both global fold recovery and catalytic geometry accuracy, structural novelty, scaffold diversity, and geometric validity, while training in just $682$ GPU-hours, roughly $4\times$ less than \rfdthree.
}
\date{\sffamily\today}

\begin{document}
\addtocontents{toc}{\protect\setcounter{tocdepth}{-1}} 

\maketitle

\section{Introduction}
\label{sec:introduction}

\begin{figure}[H]
    \centering
    \includegraphics[width=\linewidth]{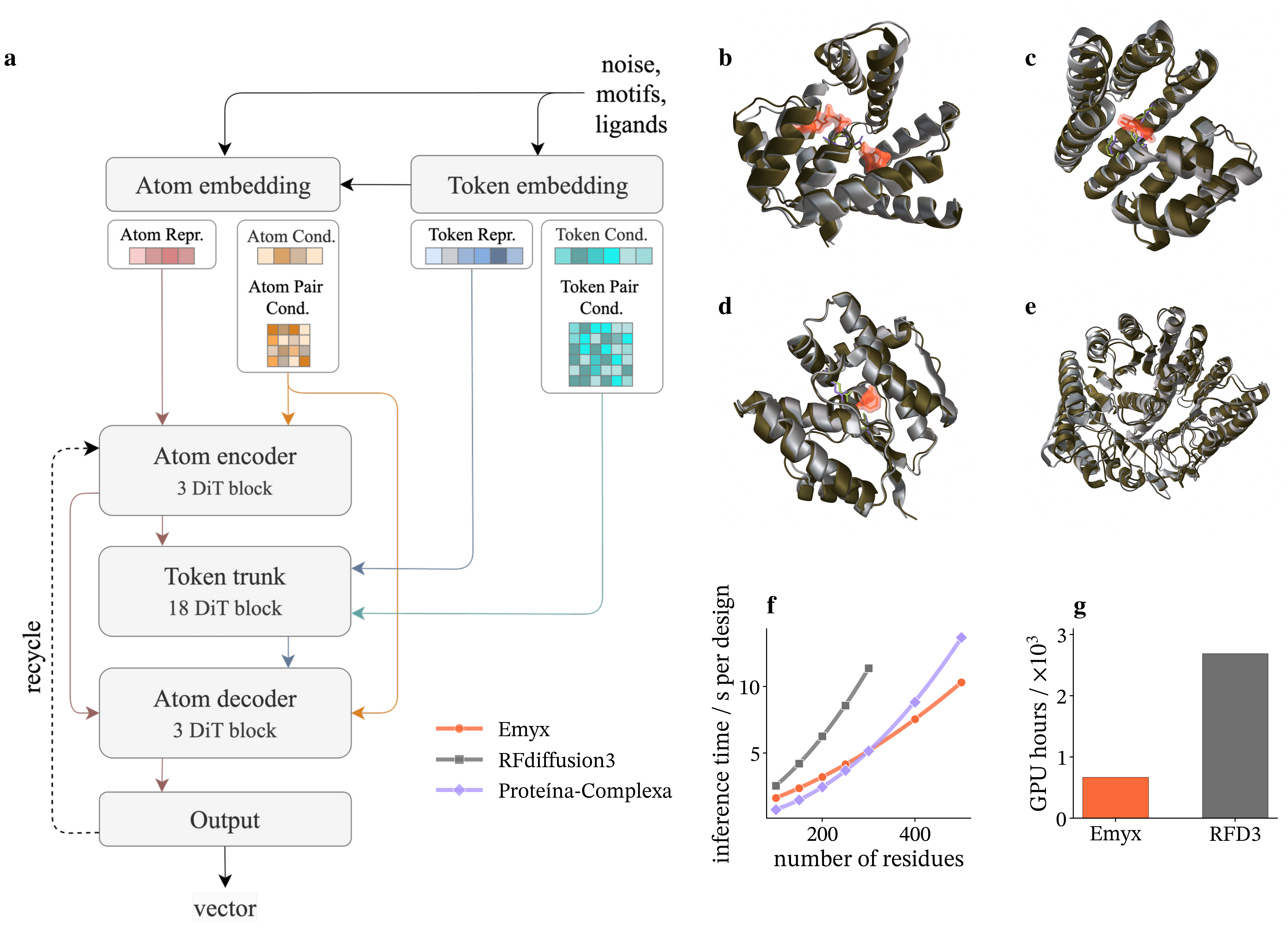}
    \caption{%
        \textbf{\emyx overview.}
        \textbf{a}, \emyx architecture overview. Token and atom embeddings combine pair features, coordinate embeddings, and categorical features into representations and pair conditioning. Atom-level encoder blocks process local atomic detail within each token's 14-atom representation. Gated cross-attention pools atom representations into token-level embeddings, which are refined by a deep transformer trunk with sparse self-attention and learned pair bias. A second cross-attention broadcasts token context back to atoms for the final atom-level decoder and output velocity head. Full component definitions and pseudocode in \S\ref{si:architecture}.
        \textbf{b}--\textbf{d}, Representative \emyx designs for AME benchmark active sites; generated backbone (dark green) overlaid with the Boltz-2 re-prediction (grey), motif residues highlighted (green/purple), bound ligand as orange sticks.
        \textbf{e}, Unconditional \emyx design, 512 residues.
        \textbf{f}, Inference time per design on a single NVIDIA A10G (200 sampling steps; \proteinacomplexa timing includes its all-atom decoder).
        \textbf{g}, Training cost in GPU-hrs (\emyx and \rfdthree only).\protect\footnotemark
    }
    \label{fig:samples_efficiency}
\end{figure}

Nature produces molecules that are near-inaccessible to synthetic chemistry, largely through the action of enzymes, which catalyse complex reactions with exceptional precision and selectivity by tightly controlling their local environment~\citep{arnold1998directed}. Engineering natural enzymes expands the reachable chemical space, but is typically confined to reactions chemically similar to those that already exist; accessing genuinely novel reaction space requires novel enzymes. \emph{De novo} enzyme design is therefore a long-standing goal of protein engineering, with applications spanning industrial and medical biocatalysis~\citep{listov2026de-b39}. One of the key computational challenges is generating scaffolds around catalytic sites of interest (Figure~\ref{fig:samples_efficiency}), producing stable protein folds that position catalytic residues around a bound ligand. Recent advances in generative models have made this more tractable~\citep{Yeh-luciferases,lauko-serine-hydrolases,kim-metallohydrolases}, but current methods remain expensive to train, produce limited structural diversity, and result in few novel successful designs on complex tasks~\citep{watson2023rfdiffusion,krishna2024rfdiffusionaa,Ahern2026-ob-rfdiffusion2,geffner2026proteinacomplexa}. All of these factors make computational enzyme design not only expensive but also limited in its exploration of novel proteins.

\paragraph{From backbone to all-atom generation.}
Early deep-learning methods for protein generation, such as RFDiffusion~\citep{watson2023rfdiffusion}, operated solely on the backbone and required a separate inverse folding model, such as ProteinMPNN~\citep{dauparas2022proteinmpnn} or LigandMPNN~\citep{dauparas2025ligandmpnn}, to obtain the sequence; that is, the protein is modelled at residue-level rather than atomic resolution. A limitation of using such methods for enzyme design is that motif residues must be manually placed at pre-specified indices along the chain, thereby reducing the success rate of generated proteins and increasing the computational cost due to manual index searches. More recent models~\citep{Ahern2026-ob-rfdiffusion2, geffner2025laproteina, rfdiffusion3} overcome this by using unindexed motif scaffolding, allowing the model to select motif indices along the chain, thus substantially improving success rates. Generating proteins in an all-atom coordinate system is inherently challenging due to the thousands of atoms comprising a typical protein, and only became tractable with the advent of modern architectures and scalable transformer designs. Models such as RFDiffusion3~\citep{rfdiffusion3} and BoltzGen~\citep{stark2025boltzgen} moved to architectures analogous to patch tokenisation in Vision Transformers~\citep{dosovitskiy2020vit}, where atom representations are pooled into per-residue tokens which are then processed in a main transformer block. This allows attention to operate in the compressed token space while retaining atomic detail in the input featurisation and coordinate decoding. Although these models generate all-atom coordinates and hence implicitly produce sequences, the generated sequences are typically low quality and necessitate re-design with an inverse folding step. This is not the only route to all-atom structures: an alternative family of protein models, La Prote\'{i}na~\citep{geffner2025laproteina}, and \proteinacomplexa~\citep{geffner2026proteinacomplexa}, uses joint flow matching in Euclidean and latent space, which are then decoded into all-atom structures and sequences. While these advances have brought all-atom protein generation within reach, current approaches remain bound to the expensive architectures inherited from structure prediction.

\paragraph{Generators do not need expensive architectures.}
All of the aforementioned models tend to share similar architectures built around dense pairwise representations and domain-specific embedding layers drawn from the mature field of structure prediction~\citep{jumper2021alphafold, wohlwend2024boltz1, chaidiscovery2024chai1}. These pair representations impose substantial memory overhead and limit scalability~\citep{abramson2024alphafold3, passaro2025boltz2}: \rfdthree utilises node-to-edge updates during recycling, while the \proteina family optionally adds triangle multiplicative pair updates~\citep{geffner2025proteina, geffner2025laproteina}, though their recommended models omit these for scalability. SimpleFold~\citep{wang2025simplefold} demonstrated that general-purpose Diffusion Transformer (DiT) blocks~\citep{peebles2023dit} can compete with domain-specific architectures for protein structure predictors at scale without pair representations or pair-attention modules. We argue that this finding carries a stronger implication for structure generators than for predictors, and that architectural simplicity can be actively beneficial. Predictors are conditioned on rich sequence-level context: multiple sequence alignments (MSA)~\citep{rao2021msa}, templates from known homologous structures, and embeddings from protein language models such as ESM~\citep{lin2023esm2, hayes2025esm3}. Together, these initialise both token and pair representations with semantically meaningful information. Generators, by contrast, condition on geometric constraints, such as chain connectivity and motif and ligand geometries, which are inductively weak and only acquire relational structure after information is mixed across tokens, lending to our hypothesis that architectural simplicity is beneficial for generators.


\footnotetext{\proteinacomplexa uses a stagewise training procedure across 48-96 A100 GPUs totalling ${\approx}1$M gradient steps~\citep{geffner2026proteinacomplexa}; wall-clock training time is not reported, precluding a direct comparison, but the number of GPUs and gradient steps suggest a compute budget that substantially exceeds both \rfdthree and \emyx.}

\paragraph{Contributions.}
We introduce \emyx, a conditional flow matching model for all-atom protein structure generation that concentrates model capacity almost entirely within standard transformer blocks, dispensing with the expensive architectures inherited from structure prediction. Our main contributions are as follows:       


\begin{itemize}
\item \textbf{Efficient all-atom generator,} trained on experimental PDB data in just 3.55 days on 8 H200 GPUs ($681.6$ GPU-hours), roughly $4\times$ faster to train than \rfdthree.


\item \textbf{Strict sc-RMSD evaluation} that requires both global fold recovery (C$_\alpha$ RMSD $< 2$\,\angstrom) and catalytic geometry accuracy (motif RMSD $< 1.5$\,\angstrom), revealing that the original AME benchmark heavy-atom sc-RMSD overstates success by roughly $2\times$ (\S\ref{sec:backbone_eval}).
\item \textbf{State-of-the-art on AME} with fewer parameters, achieving $13.4\%$ success under strict sc-RMSD, outperforming \proteinacomplexa ($8.8\%$) and \rfdthree ($6.7\%$) (\S\ref{sec:res_benchmark}).
\item \textbf{Structurally diverse scaffolds,} with the lowest median TM-score to the closest PDB hit (\edit{$0.48$} via Foldseek) and the most unique structural clusters among successful designs.

\item \textbf{EDM reparametrisation} of the flow matching interpolant into the EDM noise-level framework~\citep{karras2022edm}, a model-agnostic mapping that enables Karras-schedule sampling without retraining and substantially improves success rates over SDE Euler--Maruyama integration (\S\ref{sec:res_edm_vs_euler}).
\end{itemize}

\section{Methods}
\label{sec:methods}
%

\emyx is an all-atom protein structure generator that uses conditional flow matching over Cartesian coordinates. In this section, we briefly summarise the flow matching formulation (\S\ref{sec:fm}), protein representation (\S\ref{sec:representation}), network architecture (\S\ref{sec:arch}), training procedure (\S\ref{sec:training}), and sampling algorithms (\S\ref{sec:sampling}); full details can be found in the supplementary material.

\subsection{Conditional flow matching}
\label{sec:fm}

\emyx learns a time-dependent velocity field $\bm{v}_\theta(\xt,\, t,\, \mathbf{c},\, \mathcal{G})$ that transports samples from a noise distribution $\bm\epsilon \sim \mathcal{N}(\bm{\mu}, \sigma_{\text{data}}^2\bm{I})$ to the data distribution $\bm{x}\sim p_\text{data}$, where $\mathbf{c}$ denotes conditioning (\emph{e.g.}\ motif geometry) and $\mathcal{G}$ is a sparse connectivity graph built from the interpolated coordinates. We use a linear interpolant between noise ($t=0$) and data ($t=1$),
\begin{equation}
    \xt = (1 - t)\, \bm\epsilon + t\,\bm{x}, \qquad t \in [0,\, 1],
    \label{eq:interpolant}
\end{equation}
with target velocity $\bm{v}_t = \bm{x} - \bm\epsilon$. The model is trained via an $\ell_2$ regression objective $\mathbb{E}\bigl[\|\bm{v}_\theta(\xt,\, t,\, \mathbf{c},\, \mathcal{G}) - \bm{v}_t\|^2\bigr]$ over non-motif atoms\footnote{Motif atoms are frozen at their input coordinates throughout generation and excluded from the training loss, so the model learns to build the scaffold around a fixed catalytic geometry.} combined with an auxiliary loss based on a differentiable local
distance difference test (lDDT) (full specification in \S\ref{si:training_loss}). The denoised endpoint is recovered as
\begin{equation}
    \hat{\bm{x}} = \xt + (1 - t)\,\bm{v}_\theta(\xt,\, t,\, \mathbf{c},\, \mathcal{G}).
\end{equation}
At inference, structures are generated by integrating the ODE, $d\xt/dt = \bm{v}_\theta(\xt, t, \mathbf{c}, \mathcal{G})$, or its corresponding SDE (\S\ref{si:sampling}), from $t{=}0$ to $t{=}1$. Although flow matching and diffusion models define equivalent probability paths under Gaussian assumptions~\citep{albergo2023building, lipman2023flow}, the linear interpolant yields lower-variance training gradients and faster convergence than variance-preserving diffusion schedules~\citep{lipman2023flow, ma2024sit}, motivating our choice of formulation. We exploit the theoretical equivalence between the two frameworks in the EDM sampler (\S\ref{sec:sampling}).

\subsection{Architecture}
\label{sec:arch}
\label{sec:representation}

Each token (residue or HETATM) is represented by a fixed-size matrix of 14 atom slots (Rep14\footnote{Rep14: each residue or ligand atom occupies a fixed $14 \times 3$ coordinate matrix, with unused positions filled by ghost atoms that inherit backbone N or O coordinates in residues or masked out in ligands. The name reflects the 14-atom budget per token.}; \S\ref{si:rep14}), connected via a sparse edge graph with a fixed budget assigned by importance (\S\ref{si:features_edges}). The architecture (Figure~\ref{fig:samples_efficiency}\textbf{a}, \S\ref{si:architecture}) uses lightweight atom-level transformer blocks to encode local atomic detail into token representations, where a deep transformer module captures long-range structural context, with cross-attention modules bridging the two levels. By concentrating model capacity almost entirely in the transformer blocks and computing conditional pair representations only along sparse edges, \emyx avoids the quadratic memory overhead of pairformer blocks while retaining the ability to represent arbitrary structural relationships.

\paragraph{DiT block modifications.}
Each transformer layer follows the DiT~\citep{peebles2023dit} design (adaLN-Zero, SwiGLU~\citep{shazeer2020glu}) with three modifications targeting the low-signal generative regime: (i) a \emph{bottleneck projection} that compresses the conditioning to a small intermediate dimension before expansion to shift/scale/gate, reducing modulation parameters; (ii) \emph{sigmoid gating} bounding gate values to $[0, 1]$; and (iii) \emph{stochastic depth}~\citep{huang-stochastic-dropout} via drop path masking with probability increasing linearly from $0$ to $p_\text{path}$ across the layer stack. Full pseudocode is in \S\ref{si:architecture}.
\subsection{Training}
\label{sec:training}

We train \emyx using the flow matching loss (Eq.~\ref{eq:fm_loss}) combined with an auxiliary loss of a differentiable local distance difference test (lDDT) ~\citep{mariani2013lddt}. Conditioning features (such as secondary structure composition, radius of gyration, RASAs) and motif residue indices are stochastically masked during training to allow for both unconditional and conditional generation at inference time to steer toward desired fold properties (\S\ref{si:conditioning}). Full loss definitions, dataset preparation, timestep sampling, data augmentations, dataset preparation pipeline, batch sampling, and optimiser configuration are in \S\ref{si:training}. \emyx trains in $3.55$ days on 8 NVIDIA H200 GPUs ($681.6$ GPU-hours; Table~\ref{tab:training_cost}, \S\ref{si:efficiency}).

\subsection{Sampling}
\label{sec:sampling}

The flow matching interpolant in Eq.~\ref{eq:interpolant} admits an exact, model-agnostic reparametrisation into the EDM framework~\citep{karras2022edm}, mapping the flow time $t$ to an EDM noise level $\sigma$ and expressing the velocity as an equivalent denoiser. This enables Karras-schedule sampling~\cite{karras2022edm} and stochastic churn without retraining or architectural changes, and applies to any flow matching model using a linear interpolant. Full algorithms, derivations including the EDM reparametrisation (\S\ref{si:sampling_edm}), and a visualisation of the progressive denoising trajectory (Figure~\ref{fig:flow_sampling}) are given in \S\ref{si:sampling}.


\section{Experiments and discussion}
\label{sec:results}
We evaluate \emyx against the AME benchmark for motif-conditioned scaffolding (\S\ref{sec:experimental_setup}), propose strict sc-RMSD (\S\ref{sec:backbone_eval}), benchmark performance (\S\ref{sec:res_benchmark}), compare sampling strategies (\S\ref{sec:res_edm_vs_euler}), and characterise weight utilisation across generators and predictors (\S\ref{sec:res_weight_quality}).

\subsection{Experimental setup}
\label{sec:experimental_setup}
We evaluate against the AME benchmark~\citep{Ahern2026-ob-rfdiffusion2, rfdiffusion3}: 41 catalytic active sites spanning EC classes 1--5, with 1--7 residue islands per target (\S\ref{si:ame_benchmark}). For each motif, we generate 200 structures, redesign with LigandMPNN~\citep{dauparas2025ligandmpnn} (8 sequences each), and re-predict the structures using Boltz-2~\citep{passaro2025boltz2} (\S\ref{si:sc_protocol}). Success criteria are defined in \S\ref{sec:backbone_eval}. We additionally report structural novelty (TM-score to PDB via Foldseek~\citep{vanKempen2024foldseek}), diversity (Foldseek cluster count), and geometry validity against PeptideBuilder~\citep{peptidebuilder} ideal bond lengths and angles (\S\ref{si:novelty_diversity}). We compare against \rfdthree and \proteinacomplexa~\citep{geffner2026proteinacomplexa}, the two other all-atom generators that support ligand-constrained, unindexed, and atomic motif scaffolding against the AME benchmark. All three models are evaluated through the same pipeline (LigandMPNN redesign + Boltz-2)\footnote{Even though \proteinacomplexa reports good performance with its original generated sequences, sequence redesign with $8$ LigandMPNN sequences improves success rates for all models; results without redesign are in \S\ref{si:orig_vs_redes}.}. \emyx and \rfdthree use EDM sampling~\citep{karras2022edm} (\rfdthree natively; \emyx via reparametrisation, \S\ref{sec:sampling}); \proteinacomplexa uses SDE sampling as in its public release. Full evaluation details are in \S\ref{si:eval_details}.

\subsection{Strict sc-RMSD protocol}
\label{sec:backbone_eval}

The AME success criterion used in previous studies~\cite{geffner2026proteinacomplexa, rfdiffusion3}, hereafter \emph{heavy-atom sc-RMSD},\footnote{Self-consistency RMSD: the generated structure is redesigned with an inverse folding model and the resulting sequence is re-predicted by a structure predictor. The RMSD between generated and re-predicted coordinates measures whether the design is self-consistent, \ie whether the sequence implied by the structure folds back to the intended geometry.} aligns the generated and re-predicted structures on the \emph{motif backbone atoms} and checks that the motif heavy-atom RMSD falls below $1.5$\,\angstrom. Additionally, the criterion checks for any ligand clashes (all inter-atomic distances > $1.5$\,\angstrom). We argue that this criterion overestimates the extent to which the re-predicted structures respect the input motif (see Figure~\ref{fig:sc_rmsd_schematic} for a visual example). \emph{Tip-atom sc-RMSD}, which rectifies this overestimate, uses the same motif-backbone alignment but checks the RMSD of the re-predicted structure compared only to the motif tip atoms specified in the input motif. Both test whether the catalytic geometry is recovered locally, but neither verifies that the overall protein fold is predicted to be consistent with the design.

\begin{figure}[H]
    \centering
    \includegraphics[width=0.5\linewidth]{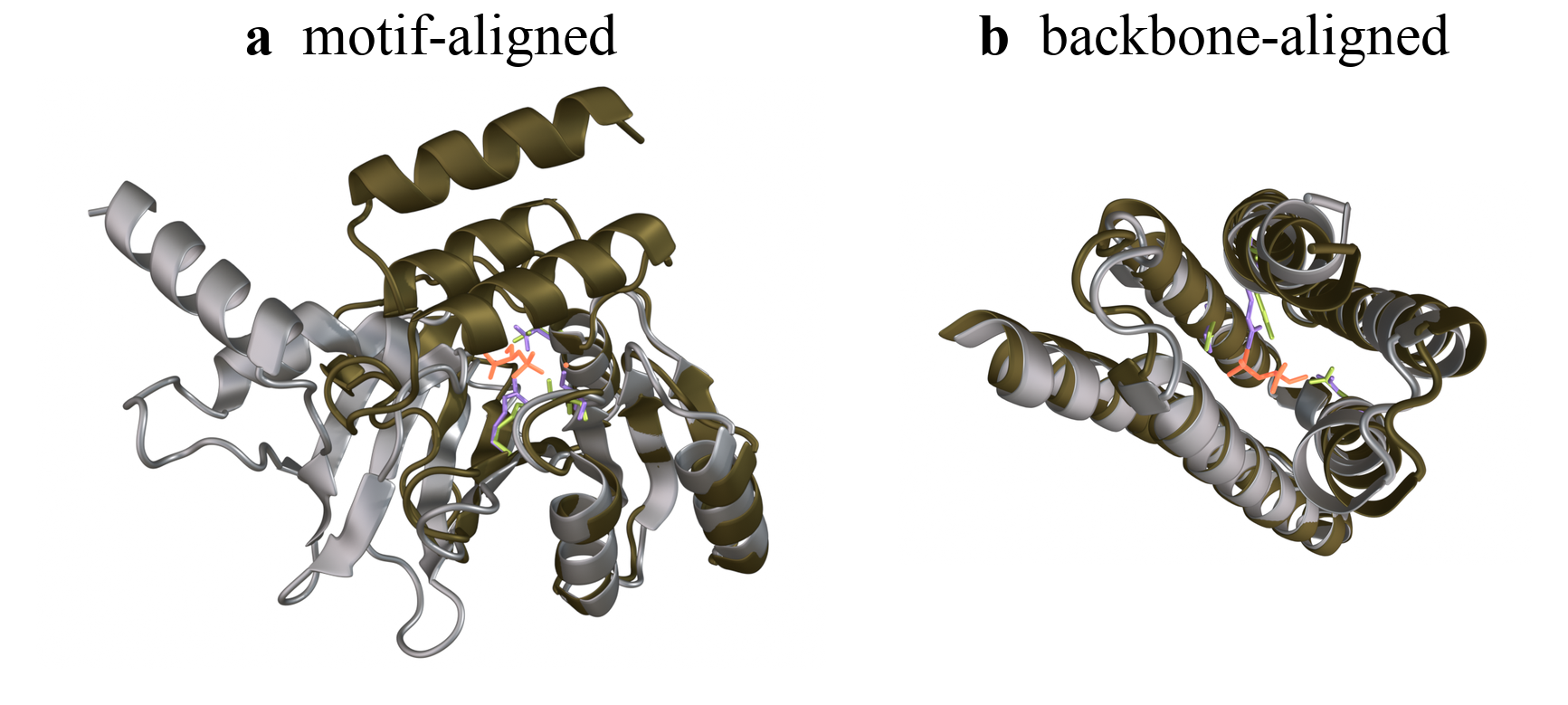}
    \caption{%
        \textbf{Strict sc-RMSD examples} (\emyx design for M0738\protect\footnotemark).
        \textbf{a}, Motif-aligned (used by heavy-atom sc-RMSD); passes motif heavy-atom RMSD $<1.5$\,\angstrom{} but global fold is incorrect.
        \textbf{b}, Same target, backbone-aligned (used by strict sc-RMSD); passes both backbone RMSD $<2.0$\,\angstrom{} and motif tip-atom RMSD $<1.5$\,\angstrom.
    }
    \label{fig:sc_examples}
\end{figure}

We propose a stricter, global backbone-aligned evaluation that requires both global fold recovery and local motif accuracy, hereafter called \emph{strict sc-RMSD}. After Kabsch alignment~\citep{kabsch1976solution} on the \emph{full backbone} of the generated and re-predicted structures, a design is successful if:
\begin{enumerate}[label=\roman*., nosep, leftmargin=*, topsep=2pt]
    \item global backbone RMSD $< 2.0$\,\angstrom, confirming the predicted structure recovers the designed fold;
    \item motif tip-atom RMSD $< 1.5$\,\angstrom{} between the set of re-predicted motif atoms and the input motif atoms, confirming catalytic geometry;
    \item no ligand clashes (all inter-atomic distances $> 1.5$\,\angstrom).
\end{enumerate}
Criterion (i) is absent from heavy and tip-atom sc-RMSD, and criterion (ii) is made more precise by comparing to the input motif atoms only. Figure~\ref{fig:sc_examples} shows two real designs from the same target where heavy-atom sc-RMSD passes but only one recovers the global fold (schematic overview in Figure~\ref{fig:sc_rmsd_schematic}). Per-target breakdowns under all three metrics are in \S\ref{si:ame_per_target}.

\subsection{Emyx outperforms on AME}
\label{sec:res_benchmark}

\footnotetext{Generated backbone (dark green) overlaid on re-prediction (grey); motif sticks in light green (generated) and purple (re-predicted); ligand in orange.}

\begin{figure}[H]
    \centering
    \includegraphics[width=\linewidth]{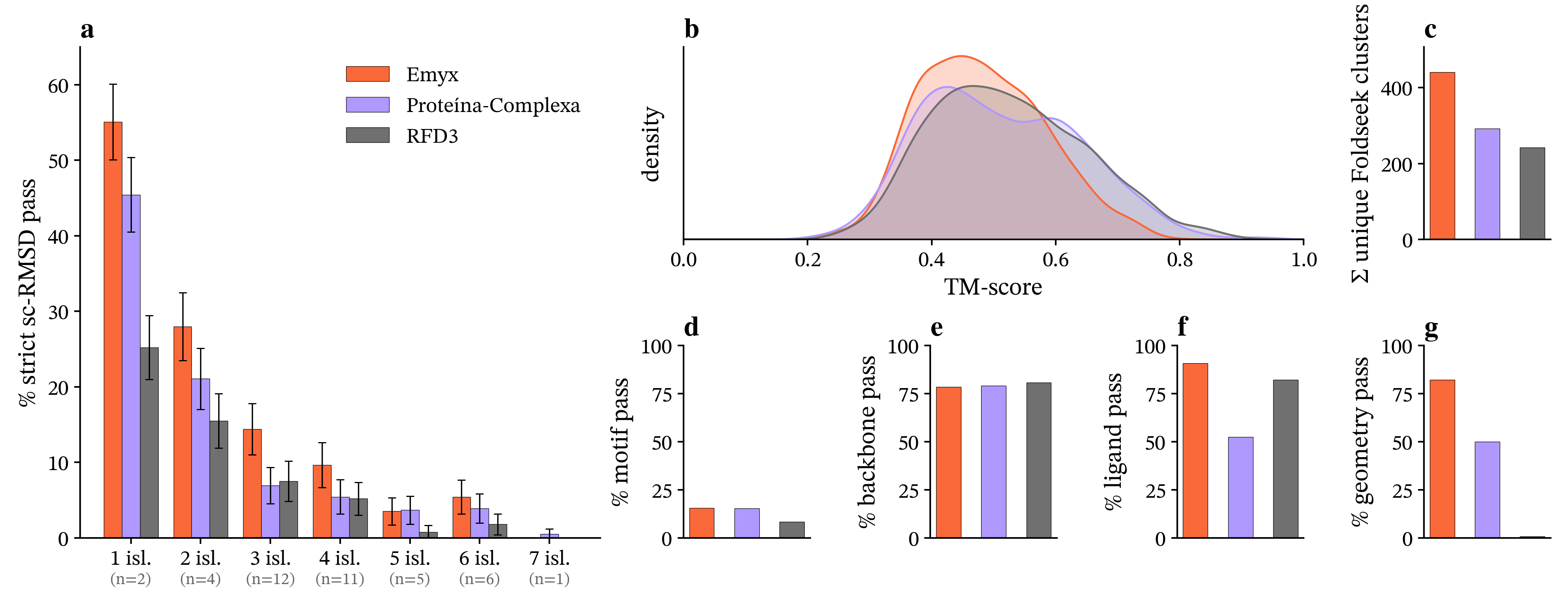}
    \caption{%
        \textbf{AME benchmark} for \emyx, \rfdthree, and \proteinacomplexa (strict sc-RMSD; \S\ref{sec:backbone_eval}).
        \textbf{a}, Strict sc-RMSD success rate, by motif island count (bootstrap mean $\pm 1\sigma$ over $1{,}000$ resamples of 100 designs).
        \textbf{b}, TM-score to the closest PDB hit (Foldseek) for all successful designs; lower is more novel.
        \textbf{c}, Raw count of unique Foldseek clusters across successful designs.
        \textbf{d}, Fraction of designs with any prediction at motif tip-atom RMSD $< 1.5$\,\angstrom.
        \textbf{e}, Fraction of designs with any prediction at backbone RMSD $< 2.0$\,\angstrom.
        \textbf{f}, Fraction of designs passing the ligand-clash filter (all inter-atomic distances $> 1.5$\,\angstrom{}).
        \textbf{g}, Fraction of designs passing the PeptideBuilder geometry check (\S\ref{si:geom_validity}).
        Per-target breakdowns in Fig.~\ref{fig:ame_success_comparison}. Efficiency reported separately in Fig.~\ref{fig:samples_efficiency}.
    }
    \label{fig:ame_combined}
\end{figure}

\paragraph{Motif scaffolding success.}
Under strict sc-RMSD (\S\ref{sec:backbone_eval}), \emyx achieves $13.4\%$ success ($39/41$ targets solved) compared to $8.8\%$ for \proteinacomplexa ($37/41$) and $6.7\%$ for \rfdthree ($35/41$; Figure~\ref{fig:ame_combined}a, Figure~\ref{fig:ame_success_comparison}c). We also report performance under the heavy-atom sc-RMSD (the original AME benchmark metric), where \emyx reaches $35.8\%$, \proteinacomplexa $21.0\%$, and \rfdthree $27.3\%$ (Figure~\ref{fig:ame_success_comparison}a), and under tip-atom sc-RMSD, where \emyx reaches $21.5\%$, \proteinacomplexa $12.0\%$, \rfdthree $12.6\%$ (Figure~\ref{fig:ame_success_comparison}b). The gap between heavy-atom sc-RMSD and strict sc-RMSD reflects designs where the active site is locally recovered but the global fold is not: roughly two thirds of designs that pass heavy-atom sc-RMSD fail the backbone check, indicating that heavy-atom sc-RMSD substantially overstates the fraction of designs that would be expected to fold correctly. Consistent with \rfdthree and \proteinacomplexa, we observe degrading performance as island count increases, but \emyx is competitive or superior on every group (per-target bars under all three metrics in \S\ref{si:ame_per_target}).
Decomposing strict sc-RMSD into its individual components is revealing: \emyx and \proteinacomplexa achieve nearly identical motif RMSD pass rates ($15.5\%$ vs.\ $15.3\%$ below $1.5$\,\angstrom) and backbone RMSD pass rates ($78.3\%$ vs.\ $79.1\%$ below $2.0$\,\angstrom), yet \proteinacomplexa passes the ligand-clash filter on only $52.3\%$ of designs compared to $90.7\%$ for \emyx (Figure~\ref{fig:ame_combined}f). Nearly half of all \proteinacomplexa designs place backbone atoms within $1.5$\,\angstrom{} of the ligand, suggesting that while its scaffold geometry is competitive, the model does not adequately respect the steric constraints imposed by the bound ligand.

\paragraph{Novelty and diversity.}

Among successful designs under strict sc-RMSD, \emyx produces structures with lower similarity to existing PDB entries (median TM-score \edit{$0.475$} vs.\ \edit{$0.499$} for \proteinacomplexa and $0.512$ for \rfdthree; Figure~\ref{fig:ame_combined}b), indicating greater structural novelty, and yields more diverse successful solutions (raw counts shown in Figure~\ref{fig:ame_combined}c): \edit{$441$} unique Foldseek clusters vs.\ \edit{$292$} for \proteinacomplexa and \edit{$242$} for \rfdthree; \emyx remains the most diverse. To control for sample-volume bias, we also report the \emph{unique-clusters per 100 designs} metric (\S\ref{si:clusters_per_100}, Table~\ref{tab:ame_results}), which credits each Foldseek cluster at most once per bootstrap draw. Notably, despite being trained on the smallest and purely experimental dataset,\footnote{\emyx is trained on experimental PDB structures (${\approx}629$K chains; \S\ref{si:dataset_preparation}), whereas \proteinacomplexa augments PDB data with ${\approx}345$K AlphaFold Database structures and ${\approx}500$K synthetic dimer pairs (${\approx}1.3$M total)~\citep{geffner2026proteinacomplexa}, and \rfdthree supplements PDB with AlphaFold2 distillation structures~\citep{rfdiffusion3}.} \emyx achieves the highest novelty and diversity scores, suggesting that architectural inductive biases and sampling quality may matter more than training set scale for scaffold diversity. The TM-score densities for each target in the AME benchmark can be found in \S\ref{si:ame_per_protein}. Full summary tables are in \S\ref{si:ame_summary}.

\paragraph{Geometric validity.}
We also examine the geometric quality of the generated samples. \emyx shows a geometry pass rate of $82.2\%$ against PeptideBuilder~\citep{peptidebuilder} ideal bond lengths and angles (Figure~\ref{fig:ame_combined}g), indicating that most structures have locally correct covalent geometry.
Among the $17.8\%$ that fail, Figure~\ref{fig:geometry_validity} (\S\ref{si:geom_validity}) shows the breakdown by check type: omega dihedrals ($11.0\%$) and intra-residue bond angles ($6.7\%$) are the dominant failure modes. \proteinacomplexa passes $49.9\%$ of geometry checks, with intra-residue bond lengths ($42.6\%$) as the primary failure mode.
\rfdthree fails $99.3\%$ of geometry checks, with intra-residue bond lengths and angles each failing in ${\approx}97\%$ of designs. We note that this is likely due to a post-processing artifact rather than an intrinsic issue with the model: \rfdthree's inference script overwrites the generated motif coordinates with the input motif coordinates whenever the generation does not exactly overlap the specified motif, producing unphysical bonds and angles (Fig.~\ref{fig:m0157_motif_closeup}, \S\ref{si:geom_validity}). Full per-check breakdown in Figure~\ref{fig:geometry_validity} (\S\ref{si:geom_validity}).

\subsection{EDM reparameterisation improves sampling}
\label{sec:res_edm_vs_euler}
As described in \S\ref{sec:sampling}, we reparametrise the flow matching interpolant in Eq.~\ref{eq:interpolant} into the EDM noise-level framework~\citep{karras2022edm}, gaining access to the Karras noise schedule and stochastic churn mechanisms. We are not aware of prior work that explicitly reparametrises a flow-matching velocity model into the EDM framework for protein structure generation. This reparametrisation is model-agnostic; it requires no retraining or architectural changes, and could be applied to other flow matching models in the field. We note that \rfdthree is trained using the EDM framework.

On a 6-target subset shared across all sampler configurations (per-target breakdown in Fig.~\ref{fig:sampler_per_target}, \S\ref{si:edm_comparison}), the EDM sampler achieves the highest strict sc-RMSD success rate, exceeding both the Euler SDE sampler and the two external baselines. The deterministic Euler ODE sampler fails entirely ($0\%$ on all 6 targets). We hypothesise that the advantage of EDM over SDE stems from the churn and overstep mechanisms, which are more effective at injecting noise and correcting trajectory errors than the corresponding Wiener process and score terms in Euler-Maruyama integration.\footnote{The EDM reparametrisation casts \emyx as a diffusion-equivalent model. The superior performance of \emyx-EDM compared to \rfdthree (itself an EDM diffusion model) isolates the architectural contribution from the sampling-paradigm advantage, providing a direct architectural evaluation.}

\subsection{Weight utilisation reveals a generator--predictor gap}
\label{sec:res_weight_quality}

Spectral analysis of trained weights across three generators and three predictors reveals an intrinsic information gap between the two model types (methodology in \S\ref{si:spectral_analysis}). Predictors initialise from MSA and template-derived representations (32--35\% effective rank; \S\ref{si:spectral_embedding}), while generators initialise from noised coordinates (7--17\% rank). This gap propagates through the transformer trunk, where all three generators converge to 15--19\% while all three predictors achieve 29--35\% (Table~\ref{tab:trunk_gap}, Figure~\ref{fig:spectral_combined}a), regardless of model capacity. Without the rich post-MSA embeddings that naturally regularise predictor trunks, generator weights are more prone to overtraining: 20--47\% of trunk layers fall below the overtrained boundary ($\alpha < 2$), compared with 3--16\% for predictors (Table~\ref{tab:alpha_stats}, Figure~\ref{fig:spectral_combined}b,c). Among generators, \emyx has the lowest overtrained fraction ($20.0\%$, mean $\alpha = 2.97$), compared with \rfdthree ($42.2\%$) and \proteinacomplexa ($46.8\%$).
We attribute this to aggressive bottlenecking in the adaLN-modulation layers (Algorithm~\ref{alg:dit_mod}), which compress the low-signal conditioning information into a lower-dimensional subspace, reducing the number of underutilised weight matrices in the main transformer block. We emphasise that this attribution is correlative rather than causal. Controlled ablations isolating each design choice are left to future work.

\section{Conclusions}
\label{sec:conclusion}

\paragraph{Fast, accurate, and diverse.}
With 140M parameters, \emyx achieves the highest strict sc-RMSD success rate against the AME benchmark ($13.4\%$ vs.\ $8.8\%$ for \proteinacomplexa and $6.7\%$ for \rfdthree), while producing more structurally novel scaffolds (median TM-score \edit{$0.48$} vs.\ \edit{$0.50$} and $0.51$). These gains come at $4\times$ lower training cost ($682$ GPU-hours vs.\ $2{,}688$ for \rfdthree). At inference, \emyx is $>2\times$ faster than \rfdthree across all chain lengths on a single A10G GPU (Figure~\ref{fig:speed}), with the gap widening at longer sequences. \proteinacomplexa shows a heavily quadratic relationship between sampling speed and chain length due to dense $O(N^2)$ attention, while \emyx operates on a fixed-budget sparse edge graph.


\paragraph{Flow matching training efficiency meets diffusion sampling.}
The linear interpolant used by \emyx is known to converge faster than the variance-preserving schedules used by diffusion-based generators~\cite{ma2024sit}, complementing the architectural simplifications described above. The exact reparametrisation of this interpolant into the EDM framework then allows \emyx to use EDM-style stochastic sampling without retraining or architectural changes, bridging the training efficiency of flow matching with state-of-the-art sampling methods developed for diffusion models. EDM gives higher success rates than standard SDE Euler--Maruyama integration (\S\ref{sec:res_edm_vs_euler}; per-target breakdown in Fig.~\ref{fig:sampler_per_target}, \S\ref{si:edm_comparison}), confirming that diffusion sampling innovations transfer directly to flow matching generators. The reparametrisation is model-agnostic and could be applied to other flow matching models in the field.

\paragraph{Limitations and future work.}

Strict sc-RMSD enforces global backbone agreement between generated and re-predicted structures. This can over-penalise functionally valid designs that contain unstructured or flexible regions far from the active site. While such designs can be valid, unstructured and flexible regions fall outside of the scope of the single state structure predictors and can be liabilities for desirable protein properties like stability.

Although the generated amino acid distribution matches the natural PDB distribution more closely than both \rfdthree and LigandMPNN, \emyx still requires a separate inverse folding step for sequence redesign; joint sequence-structure generation is the most impactful next step. The self-consistency evaluation pipeline is a significant computational bottleneck: each design requires LigandMPNN sequence redesign followed by full MSA-conditioned structure prediction with Boltz-2, making large-scale sampler comparisons expensive. For this reason, the EDM versus SDE comparison (\S\ref{sec:res_edm_vs_euler}) was conducted on 6 of 41 AME targets rather than the full benchmark; the consistent advantage of EDM across these targets was sufficient to justify adopting it for the main evaluation. The spectral analysis (\S\ref{sec:res_weight_quality}) characterises a generator-predictor gap in weight utilisation but does not include ablation studies isolating which architectural choices are responsible for \emyx's lower overtraining fraction; controlled ablations would strengthen the architectural argument.

\paragraph{Broader impact.}
\textit{De novo} enzyme design has direct applications in industrial biocatalysis, pharmaceutical synthesis, and sustainable chemistry. Lower training cost and faster inference reduce barriers to entry, potentially broadening participation beyond well-resourced institutions. The strict sc-RMSD evaluation protocol introduced here also addresses a gap in the field's standard success metric, and can help to diagnose additional failure modes. 

As with all generative models for protein structure, dual-use risks exist. In this case they are partially mitigated by the nature of the task: Emyx is a scaffold generator conditioned on a pre-specified catalytic motif, and the design of such motifs with novel or harmful catalytic function is a separate and substantially harder problem.

\begin{ack}

We thank Thomas Blaschke, and Luke Dicks for helpful discussions and feedback. We also thank Simon Blount, Harry Jubb, Abhinav Yadav, Mark Szepieniec, Stanley Hill, and Emlyn Clay for engineering support. We thank Tom Sayers for producing the protein illustrations. We thank the UK Sovereign AI Unit for access to IsambardAI for model training. Nicholas Williams and Ward Haddadin contributed equally to this work. Author order was decided by Nicholas Woodall's fair toss of a fifty pence piece.

\end{ack}

\bibliographystyle{unsrtnat}
\bibliography{references}

\clearpage
\appendix
\addtocontents{toc}{\protect\setcounter{tocdepth}{2}}  
\setcounter{tocdepth}{2}

\begin{center}
    {\LARGE\bfseries Supplementary Information}
\end{center}
\vspace{6pt}
\renewcommand{\contentsname}{}
{\small\setlength{\parskip}{0pt}\setlength{\lineskiplimit}{0pt}\tableofcontents}
\vspace{1em}

%

\section{Protein representation}
\label{si:representation}

\subsection{Rep14 tokenisation}
\label{si:rep14}

\emyx represents the entire structure (protein, ligands, and metals) with a unified two-level representation. At the \emph{atom level}, heavy atoms carry 3D coordinates and categorical node and bond features. At the \emph{token level}, whole residues and individual HETATMs constitute single tokens, each carrying its own set of 3D coordinates and categorical node and bond features. The Rep14 representation provides a fixed-size interface for cross-attention between atom and token levels and enables vectorised conversion between the two resolutions.

Token coordinates are represented by a fixed-size matrix $\bm{x} \in \mathbb{R}^{14 \times 3}$. For residues, the first four positions (0--3) correspond to backbone atoms (N, C$_\alpha$, C, O) and the remaining positions (4--13) to sidechain atoms. Following \citep{stark2025boltzgen}, residues with fewer than 14 heavy atoms fill the remaining positions with \emph{ghost atoms} whose coordinates are set to those of either the backbone nitrogen or the backbone oxygen. Critically, the assignment of each ghost position to N or O is deterministic and unique per residue type (Table~\ref{tab:rep14}), allowing generated residues to be readily identified by counting the number of atoms coinciding with their N and O positions. For HETATM tokens, the $x \in \mathbb{R}^{14 \times 3}$  coordinates contain a single real atom at the index $=1$ position, with all other positions masked as invalid. Ghost atoms in residue tokens are treated as valid atoms (their coordinates carry geometric information) and participate in distance calculations and edge construction using backbone N or O coordinates, whereas empty positions in HETATM tokens are masked out during cross-attention.

\definecolor{ghostN}{HTML}{4A90D9}
\definecolor{ghostO}{HTML}{D94A4A}
\definecolor{realatom}{HTML}{2E8B57}
\newcommand{\gn}{\cellcolor{ghostN!15}\textcolor{ghostN}{\scriptsize N}}
\newcommand{\go}{\cellcolor{ghostO!15}\textcolor{ghostO}{\scriptsize O}}
\newcommand{\ga}[1]{\cellcolor{realatom!10}\textcolor{realatom}{#1}}
\newcommand{\gcn}[1]{\textcolor{ghostN}{\textbf{#1}}}
\newcommand{\gco}[1]{\textcolor{ghostO}{\textbf{#1}}}
\newcommand{\inv}{\textcolor{gray}{--}}

\begin{table}[H]
    \caption{%
        \textbf{Rep14 atom assignment.} Each residue is represented by 14 atom slots. Positions 0--3 are backbone; positions 4--13 are sidechain. Real atoms are shown in \textcolor{realatom}{green}; ghost atoms in
        \textcolor{ghostN}{blue} (inheriting backbone N coordinates) or
        \textcolor{ghostO}{red} (inheriting backbone O coordinates).
        The $n_\text{\gcn{N}}$ and $n_\text{\gco{O}}$ columns give the total ghost count on each backbone atom. Each residue has a unique $(n_\text{N}, n_\text{O})$ pair, enabling the residue type to be readily inferred from generated coordinates.
    }
    \label{tab:rep14}
    \centering
    \scriptsize
    \setlength{\tabcolsep}{2.2pt}
    \begin{tabular}{l cccc cccccccccc cc}
        \toprule
         & 0 & 1 & 2 & 3 & 4 & 5 & 6 & 7 & 8 & 9 & 10 & 11 & 12 & 13 & $n_\text{\gcn{N}}$ & $n_\text{\gco{O}}$ \\
        \midrule
        GLY & \ga{N} & \ga{CA} & \ga{C} & \ga{O} & \go & \go & \go & \go & \go & \go & \go & \go & \go & \go & \gcn{0} & \gco{10} \\
        ALA & \ga{N} & \ga{CA} & \ga{C} & \ga{O} & \ga{CB} & \go & \go & \go & \go & \go & \go & \go & \go & \go & \gcn{0} & \gco{9} \\
        SER & \ga{N} & \ga{CA} & \ga{C} & \ga{O} & \ga{CB} & \ga{OG} & \gn & \gn & \gn & \gn & \gn & \gn & \gn & \gn & \gcn{8} & \gco{0} \\
        CYS & \ga{N} & \ga{CA} & \ga{C} & \ga{O} & \ga{CB} & \go & \ga{SG} & \go & \go & \go & \go & \go & \go & \go & \gcn{0} & \gco{8} \\
        PRO & \ga{N} & \ga{CA} & \ga{C} & \ga{O} & \ga{CB} & \ga{CG} & \ga{CD} & \go & \go & \go & \go & \go & \go & \go & \gcn{0} & \gco{7} \\
        VAL & \ga{N} & \ga{CA} & \ga{C} & \ga{O} & \ga{CB} & \ga{CG1} & \ga{CG2} & \gn & \gn & \gn & \gn & \gn & \gn & \gn & \gcn{7} & \gco{0} \\
        THR & \ga{N} & \ga{CA} & \ga{C} & \ga{O} & \ga{CB} & \ga{OG1} & \ga{CG2} & \gn & \gn & \gn & \go & \go & \go & \go & \gcn{3} & \gco{4} \\
        LEU & \ga{N} & \ga{CA} & \ga{C} & \ga{O} & \ga{CB} & \ga{CG} & \ga{CD1} & \ga{CD2} & \gn & \gn & \gn & \gn & \go & \go & \gcn{4} & \gco{2} \\
        ILE & \ga{N} & \ga{CA} & \ga{C} & \ga{O} & \ga{CB} & \ga{CG1} & \ga{CG2} & \ga{CD1} & \go & \go & \go & \go & \go & \go & \gcn{0} & \gco{6} \\
        ASP & \ga{N} & \ga{CA} & \ga{C} & \ga{O} & \ga{CB} & \ga{CG} & \ga{OD1} & \gn & \ga{OD2} & \gn & \go & \go & \go & \go & \gcn{2} & \gco{4} \\
        ASN & \ga{N} & \ga{CA} & \ga{C} & \ga{O} & \ga{CB} & \ga{CG} & \ga{OD1} & \ga{ND2} & \gn & \go & \go & \go & \go & \go & \gcn{1} & \gco{5} \\
        MET & \ga{N} & \ga{CA} & \ga{C} & \ga{O} & \ga{CB} & \ga{CG} & \ga{SD} & \ga{CE} & \gn & \gn & \gn & \gn & \gn & \gn & \gcn{6} & \gco{0} \\
        GLN & \ga{N} & \ga{CA} & \ga{C} & \ga{O} & \ga{CB} & \ga{CG} & \ga{CD} & \ga{OE1} & \ga{NE2} & \go & \go & \go & \go & \go & \gcn{0} & \gco{5} \\
        GLU & \ga{N} & \ga{CA} & \ga{C} & \ga{O} & \ga{CB} & \ga{CG} & \ga{CD} & \ga{OE1} & \gn & \ga{OE2} & \gn & \go & \go & \go & \gcn{2} & \gco{3} \\
        LYS & \ga{N} & \ga{CA} & \ga{C} & \ga{O} & \ga{CB} & \ga{CG} & \ga{CD} & \ga{CE} & \ga{NZ} & \gn & \gn & \gn & \gn & \gn & \gcn{5} & \gco{0} \\
        HIS & \ga{N} & \ga{CA} & \ga{C} & \ga{O} & \ga{CB} & \ga{CG} & \ga{ND1} & \ga{CD2} & \ga{CE1} & \ga{NE2} & \go & \go & \go & \go & \gcn{0} & \gco{4} \\
        PHE & \ga{N} & \ga{CA} & \ga{C} & \ga{O} & \ga{CB} & \ga{CG} & \ga{CD1} & \ga{CD2} & \ga{CE1} & \ga{CE2} & \ga{CZ} & \go & \go & \go & \gcn{0} & \gco{3} \\
        ARG & \ga{N} & \ga{CA} & \ga{C} & \ga{O} & \ga{CB} & \ga{CG} & \ga{CD} & \ga{NE} & \ga{CZ} & \ga{NH1} & \ga{NH2} & \gn & \gn & \gn & \gcn{3} & \gco{0} \\
        TYR & \ga{N} & \ga{CA} & \ga{C} & \ga{O} & \ga{CB} & \ga{CG} & \ga{CD1} & \ga{CD2} & \ga{CE1} & \ga{CE2} & \ga{CZ} & \ga{OH} & \go & \go & \gcn{0} & \gco{2} \\
        TRP & \ga{N} & \ga{CA} & \ga{C} & \ga{O} & \ga{CB} & \ga{CG} & \ga{CD1} & \ga{CD2} & \ga{CE2} & \ga{CE3} & \ga{NE1} & \ga{CZ2} & \ga{CZ3} & \ga{CH2} & \gcn{0} & \gco{0} \\
        \midrule
        HETATM & \inv & \ga{atom} & \inv & \inv & \inv & \inv & \inv & \inv & \inv & \inv & \inv & \inv & \inv & \inv & \inv & \inv \\
        \bottomrule
    \end{tabular}
\end{table}

\subsection{Per-atom features}
\label{si:features_atom}

Each atom carries the following categorical features: element type (56 classes, spanning the periodic table including biologically relevant metals such as Fe, Zn, Cu, Mg, Mn, Co, Ni, and Ca), atom index within its token (14 positions), relative accessible surface area (RASA, binned into four categories: buried $< 0.1$, intermediate $0.1$--$0.9$, exposed $> 0.9$, and unknown/NaN), secondary structure assignment (coil, helix, sheet, unknown, non-amino-acid; 5 classes), terminus status (N-terminus, C-terminus, or other; 3 classes), and three boolean masks indicating whether the atom belongs to a standard residue, a HETATM, or a motif.

\subsection{Per-token features}
\label{si:features_token}

Each token carries: residue or HETATM type (22 classes: 20 standard amino acids, UNK for unknown or masked residues, and HETATM), secondary structure assignment (5 classes), terminus status (3 classes), and boolean flags for residue, HETATM, and motif membership.

\subsection{Global features}
\label{si:features_global}

In addition to per-atom and -token features, each structure is annotated with global features: secondary structure composition (proportions of coil, helix, sheet, and unknown, each discretised into 10 bins), radius of gyration ($R_g$, discretised into 12 bins at $0.25$\,\angstrom{} intervals from 0 to $2.5$\,\angstrom{}). These features serve as lightweight, stochastically masked conditioning signals that allow the model to bias generation toward desired structural properties.

\subsection{Bond representation}
\label{si:features_bonds}

Chemical connectivity is represented at two levels. At the atom-level, ligands receive bonds with bond order (single, double, triple, aromatic). At the token-level, we use sequence adjacency between consecutive residues on the same chain and the ligand bonds again (since each ligand atom is a token). Both are used in the sparse edge construction and pair bias computation described below.

\subsection{Sparse edge construction}
\label{si:features_edges}

\emyx constructs a sparse graph with a fixed edge-budget per node at both atom and token levels. Edges are assigned according to a five-level priority scheme (hyperparameters are found in Table~\ref{tab:graph_hyperparams}):
\begin{enumerate}
    \item \textbf{Sequence neighbours}: residues within $\pm n_\text{seq}$ of the specified residue on the same chain.
    \item \textbf{Chemical bonds}: all atom-atom and token-token bonds from the structure. Since each ligand atom is its own token, ligand atom-atom bonds directly become token-token edges at the token level, preserving ligand connectivity in both graphs.
    \item \textbf{Ligand $k$-NN}: the $k_\text{lig}$ nearest HETATM neighbours of each HETATM atom, ensuring ligand-internal connectivity is maintained throughout the generation.
    \item \textbf{Motif connectivity} (token level only): edges from every non-motif token to every motif token. This connectivity ensures that the scaffold always has direct access to the motif geometry.
    \item \textbf{$k$-NN fill}: the remaining budget per node (atom or token) is filled by nearest neighbours in Euclidean distance.
\end{enumerate}

\section{Training details}
\label{si:training}

\subsection{Base distribution}
\label{si:training_base}

The base distribution is defined as $\epsilon \sim \mathcal{N}(\bm{\mu}, \sigma_\text{data} \bm{I})$. The standard deviation is set to $\sigma_{\text{data}}=10$ to match the variance of the data. The centre of the noise distribution $\bm{\mu}$ is specified in two ways (with equal probability): either as the centre of mass of the target structure or as the centre of mass of the motif atoms. Furthermore, during training, the noise is stochastically translated from the specified centre $\bm{\mu}$ with a small deviation to avoid memorisation.

\subsection{Loss functions}
\label{si:training_loss}

\edit{The training loss combines a flow matching objective with an auxiliary loss based on a differentiable local distance difference test (lDDT),
\begin{equation}
    \mathcal{L} = \mathcal{L}_\text{FM} + \alpha(t) \mathcal{L}_\text{lDDT}.
    \label{eq:total_loss}
\end{equation}
The flow matching loss is a} per-atom mean squared error between predicted and true velocities over non-motif atoms,
\begin{equation}
    \mathcal{L}_\text{FM}
    = \mathbb{E}_{\bm{x}, \bm\epsilon, t}
    \Bigl[\,
        \frac{1}{|\mathcal{A}|} \sum_{i \in \mathcal{A}}
        \bigl\| \bm{v}_\theta^{(i)} - \bm{v}_t^{(i)} \bigr\|^2
    \,\Bigr],
    \label{eq:fm_loss}
\end{equation}
where $\mathcal{A}$ is the set of non-motif atoms. \edit{The smoothed lDDT loss~\citep{mariani2013lddt} provides a differentiable structural quality signal computed from the predicted endpoint $\hat{\bm{x}} = \bm{x}_t + (1-t)\bm{v}_\theta$. For each pair $(i, j)$ of non-motif atoms from different tokens within an inclusion radius $r$, we compute the absolute distance error $\delta_{ij} = |d_{ij}^\text{pred} - d_{ij}^\text{true}|$ and evaluate it against four thresholds $\theta_k \in \{0.5, 1.0, 2.0, 4.0\}$\,\angstrom{} using a sigmoid relaxation. The smoothed lDDT loss is,
\begin{equation}
    \mathcal{L}_\text{lDDT}
    = \mathbb{E}_{\bm{x}, \bm\epsilon, t}
    \Biggl[
        1 - \frac{1}{|\mathcal{P}|}
        \sum_{(i,j) \in \mathcal{P}}
        \frac{1}{4} \sum_{k=1}^{4}
        \sigma\!\bigl(\kappa\,(\theta_k - \delta_{ij})\bigr)
    \Biggr],
    \label{eq:lddt_loss}
\end{equation}
where $\mathcal{P}$ is the set of valid atom pairs within the inclusion radius $r = 15$\,\angstrom{}, $\sigma$ is the sigmoid function, and $\kappa=1.0$ controls steepness.

We apply a time-dependent weight to the lDDT loss that ramps up the structural supervision closer to $t=1$ where the structure is close to the data manifold.}

\begin{figure}[!htb]
\centering
\begin{minipage}[c]{0.60\textwidth}
\begin{equation}
    \alpha(t)
    = 0.25 \bigl(1 + 8 \cdot \mathrm{ReLU}(t - 0.5)\bigr),
    \label{eq:lddt_loss_scale}
\end{equation}
\end{minipage}%
\hfill
\begin{minipage}[c]{0.32\textwidth}
    \centering
    \includegraphics[width=\textwidth]{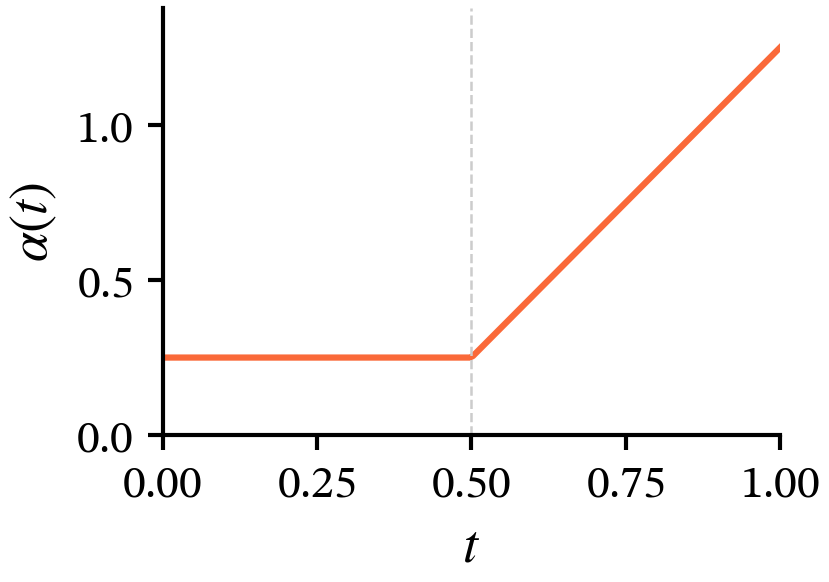}
\end{minipage}
\vspace{-6pt}
\caption{\edit{\textbf{lDDT loss scale.} $\alpha(t)$. Weighting is flat for $t < 0.5$ and increases linearly to $5\times$ at $t = 1$, concentrating structural supervision on timesteps where pairwise distances are being resolved.}}
\label{fig:lddt_scale}
\end{figure}

\subsection{Timestep sampling}
\label{si:training_timestep}

The flow time $t$ is drawn from a mixture of $\text{Uniform}(0, 1)$ and $\text{Beta}(\alpha=1.9, \beta=1.0)$ distributions.

\begin{figure}[H]
\centering
\begin{minipage}[c]{0.60\textwidth}
\begin{equation}
    t \sim (1 - p)\,\mathrm{Beta}(\alpha, \beta) + p\,\mathrm{Uniform}(0, 1),
    \quad p = 0.02,
    \label{eq:time_sampling}
\end{equation}
\end{minipage}%
\hfill
\begin{minipage}[c]{0.32\textwidth}
    \centering
    \includegraphics[width=\textwidth]{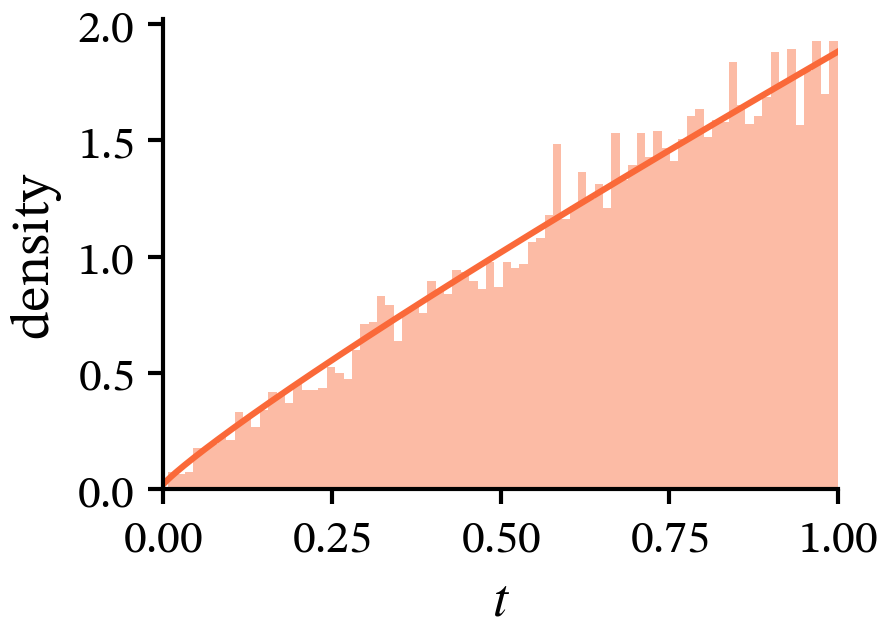}
\end{minipage}
\vspace{-6pt}
\caption{\textbf{Timestep sampling distribution.} Parameters: $\alpha = 1.9$, $\beta = 1.0$, $p = 0.02$ (Eq.~\ref{eq:time_sampling}).}
\label{fig:time_sampling}
\end{figure}

We clamp the time to $t \in [10^{-3},\, 1 - 10^{-3}]$. The right-skewed Beta component concentrates samples toward $t = 1$, where fine structural details are being learned, while the uniform component ensures adequate coverage of early timesteps (Figure~\ref{fig:time_sampling}). \edit{This biasing shares the same motivation as the time weighting in $\mathcal{L}_\text{lDDT}$: the model should invest more capacity in the final, structure-critical portion of the trajectory.}

\subsection{Data augmentations}
\label{si:training_augmentations}

During training, we augment structures in the following way (applied in order):
\begin{enumerate}
    \item \textbf{Spatial cropping}: Structures are cropped to a maximum token budget. A seed atom (from the HETATMs if available) is selected, and whole residues or whole ligands are greedily added by increasing distance from the seed. Unlike other models which crop based on spatial and sequence distance with equal probability, we only crop by spatial distance. We suspect that sequential cropping introduces artificial gaps in the proteins that detrimentally affect learning.

    \item \textbf{Motif sampling}: With some probability, between 2 and 20 residues are selected as a frozen motif. With equal probability, motif residues are sampled as contiguous 3-residue segments along the chain or as random individual residues. For each selected residue, the atoms included in the motif are determined by one of two modes: \textbf{(a)} \emph{all mode}, which includes all non-ghost atoms, or \textbf{(b)} \emph{tip mode}, which samples part of the residue starting from a seed atom and expanding $n$ bonds around that atom, where $n$ is sampled from a geometric distribution $n \sim \text{Pr}(n, p=0.25) = (1-p)^{n -1} p$. Selected atoms are duplicated and specified as \textit{motif}.

    \item \textbf{HETATM freezing}: With some probability, we freeze HETATM atoms (ligands, cofactors, metals) alongside the motif residues as part of the motif condition.

    \item \textbf{Conditional feature masking}: Features are stochastically masked during training to allow for both unconditional and conditional inference. For flowing (non-motif) atoms, all features apart from the atom index in the Rep14 representation (\S\ref{si:rep14}) are always masked. For motif residues, features are only masked half of the time. Global features are also each independently masked half of the time. In addition, the motif residue index (its position along the chain) is independently masked with probability $0.5$, decoupling motif identity from a fixed sequence location and allowing the scaffold to place the motif at any chain position at inference time.

    \item \textbf{SE(3) augmentation}: All structures are initially centred at the origin and then randomly rotated and translated stochastically for learning 3D equivariance.

    \item \textbf{Coordinate scaling}: The noise and data coordinates are scaled down by $\sigma_{\text{data}} = 10$ before being passed to the model for numerical and training stability.

\end{enumerate}

\subsection{Dataset preparation}
\label{si:dataset_preparation}

\paragraph{Sources and filtering.}
Starting protein structures were obtained from (i) PDB-REDO~\citep{joosten2014pdbredo}, which provides re-refined and rebuilt X-ray crystallographic models, and (ii) the RCSB Protein Data Bank~\citep{berman2000pdb} for entries not available in PDB-REDO (\textit{e.g.}\ NMR and cryo-EM structures). Structures were filtered using deposited metadata to retain crystallographic and cryo-EM structures solved to a resolution of $4.5$\,\angstrom{} or better. No deposition-date cutoff was applied. After filtering, 243{,}094 structures were selected for processing (71{,}391 from PDB-REDO, 171{,}703 from RCSB). Individual monomeric chains (together with their close ligands and metals) were then extracted from the processed entries, yielding 629,290 final training structures.

\paragraph{Primary conformation and biological assembly.}
Structures were collapsed to the primary conformation (altloc~A or the first deposited conformer). The first biological assembly annotated in each mmCIF file was constructed by applying the deposited symmetry operations. Atoms overlapping within $0.2$\,\angstrom{} after symmetry expansion were merged to avoid duplication. Author-assigned chain identifiers maintained a strict one-to-one correspondence between author chains and label chains.

\paragraph{Renumbering and chain separation.}
Residues were renumbered sequentially without PDB insertion codes. Polymer chains were separated from ligand entities.

\paragraph{Ligand filtering.}
The following ligands were removed:
\begin{itemize}
    \item All water molecules (HOH).
    \item An empirically curated list of 330 chemical component identifiers corresponding to common buffer components, cryoprotectants, and other non-biological small molecules (\textit{e.g.}\ DMSO, polyethylene glycol, glycerol, sulfate, acetate).
    \item Metal ions classified as crystallisation artefacts. Metals were scored by the number of protein or ligand coordination contacts, and ions with fewer than two coordination partners were removed. Alkaline earths (Sr, Rb, Ba) and various heavy metals (Ag, Au, Pt, Pd, La, Ce, Sm, Eu, Gd, Tb, Yb, Hg, Pb, Tl) were removed unconditionally, as they are contrast agents rather than biological cofactors under normal physiological conditions. Cadmium ions, occasionally used in crystallisation as isomorphic replacement for zinc, were converted to zinc.
    \item N-linked and O-linked glycan chains. Substrate sugars in enzyme active sites were retained.
    \item Ligands with any atom within $1.0$\,\angstrom{} of a polymer atom, which are likely modelling artefacts.
\end{itemize}

\paragraph{Non-canonical amino acids.}
Non-canonical amino acids (ncAAs) were converted to their nearest canonical parent using the Chemical Component Dictionary (CCD) parent mappings. Where no parent mapping was available, residues were converted to alanine. Specific conversions included:
\begin{itemize}
    \item selenomethionine (MSE) to methionine, with Se replaced by S;
    \item aminoisobutyric acid (AIB) to alanine, with the second C$\beta$ removed;
    \item phosphoserine (SEP), phosphothreonine (TPO), and phosphotyrosine (PTR) to their unmodified parents, with the phosphate group removed;
    \item various methylated lysines (MLY, M3L) to lysine;
    \item oxidised cysteines (CSO, CSD) to cysteine;
    \item hydroxyproline (HYP) to proline.
\end{itemize}
Amber force-field protonation-state residue names (ASH, GLH, HIE, HID, HIP) were disambiguated from identically named RCSB ligand codes to prevent downstream issues. Standard amino acids deposited as HETATM records (\textit{i.e.}\ as ligands rather than polymer residues) were assigned distinct three-letter codes to avoid ambiguity with their polymeric counterparts.

\paragraph{Final filtering.}
Structures without peptide polymers were discarded. Polymer chains shorter than 10 residues were excluded from the final output. Nucleic acid chains were removed from the output structures. Covalent bond records were maintained and updated to reflect chain renumbering and biological assembly expansion.

\paragraph{Chain extraction.}
Individual monomeric chains, together with any close ligands and metals, were extracted from the processed multimeric structures. Extracted chains shorter than 10 residues were discarded. Each structure was classified by oligomeric state (true monomer or extracted monomer) based on the number of polymeric chains in its source entry.

\subsection{Batch sampling and clustering}
\label{si:training_batching}

The resulting structures were clustered using Foldseek~\citep{vanKempen2024foldseek} structural clustering, yielding $\approx24{,}600$ clusters, and partitioned into training and validation sets (90/10) via a greedy stratified split that preserves category distributions across clusters. Structures are assigned to categories based on the presence of free ligands and metals (protein-monomer, protein-ligand, protein-metal, protein-ligand-metal). Batch sampling follows a three-level hierarchy: category probabilities (protein-monomer 0.30, protein-ligand 0.50, protein-metal 0.10, protein-ligand-metal 0.10), then cluster-level sampling weighted by clamped cluster size, then uniform sampling within each cluster.

\subsection{Optimisation and compute}
\label{si:training_optimisation}

We use AdamW~\citep{loshchilov2019adamw} with learning rate $2 \times 10^{-4}$, weight decay $3 \times 10^{-3}$, $(\beta_1, \beta_2) = (0.9, 0.95)$, and linear warmup from $2 \times 10^{-8}$ to $2 \times 10^{-4}$ over $1000$ steps. Gradients are clipped at norm $10$, and we use exponential moving average (EMA) weight averaging from step $1000$ onwards. \emyx is trained in bfloat16 mixed precision on a single node of 8 NVIDIA H200 GPUs with a per-GPU batch size of 64, yielding an effective batch size of 512 samples per optimiser step. Training completes in $\approx3.5$ days for a total of 248,000 steps.

\subsection{Hyperparameters}
\label{si:training_hyperparameters}

\emyx has 140M parameters in total (Table~\ref{tab:param_counts}). We summarise the hyperparameters in Tables~\ref{tab:hyperparams},~\ref{tab:graph_hyperparams}, and~\ref{tab:data_hyperparams}.

\begin{table}[H]
    \caption{\textbf{Parameter counts by module.}}
    \label{tab:param_counts}
    \centering
    \small
    \begin{tabular}{lr}
        \toprule
        Token embedding     & 1,336,111 \\
        Atom embedding      & 568,458 \\
        Time embedding      & 31,520 \\
        \midrule
        \quad Atom encoder   & 833,088 \\
        \quad Downcast (cross-attention) & 1,969,280 \\
        \quad Token trunk    & 133,721,280 \\
        \quad Upcast (cross-attention) & 258,560 \\
        \quad Atom decoder   & 833,088 \\
        \cmidrule{1-2}
        Trunk (subtotal)     & 137,615,296 \\
        \midrule
        Output head          & 33,539 \\
        Recycling layers     & 608,432 \\
        \midrule
        \textbf{Total}       & \textbf{140,193,356} \\
        \bottomrule
    \end{tabular}
\end{table}

\begin{table}[H]
    \caption{\textbf{Model hyperparameters.}}
    \label{tab:hyperparams}
    \centering
    \small
    \begin{tabular}{lc}
        \toprule
        Atom dimension & 128 \\
        Atom layers & 3 \\
        Atom heads & 4 \\
        Token dimension & 768 \\
        Token edge dimension & 256 \\
        Token layers & 18 \\
        Token heads & 16 \\
        Cross-attention heads & 4 \\
        Upcast duplicates & 14 \\
        Time dimension & 32 \\
        Fourier frequencies & 256 \\
        Dropout ($p_\text{drop}$) & 0.1 \\
        Drop-path rate ($p_\text{path}$) & 0.1 \\
        Modulation bottleneck (atom / token) & 16 / 64 \\
        Feature embedding bottleneck & 32 \\
        Recycles & 2 \\
        \bottomrule
    \end{tabular}
\end{table}

\begin{table}[H]
    \caption{\textbf{Graph construction hyperparameters.}}
    \label{tab:graph_hyperparams}
    \centering
    \small
    \begin{tabular}{lc}
        \toprule
        Atom sequence neighbours & $\pm 1$ \\
        Atom edge budget & 128 \\
        Token sequence neighbours & $\pm 32$ \\
        Token edge budget & 128 \\
        Ligand k-NN (atom / token) & 32 / 32 \\
        \bottomrule
    \end{tabular}
\end{table}

\begin{table}[H]
    \caption{%
        \textbf{Data pipeline and augmentation hyperparameters.}
        Parameters governing the training data pipeline, organised by stage.
    }
    \label{tab:data_hyperparams}
    \centering
    \small
    \begin{tabular}{lc}
        \toprule
        \multicolumn{2}{l}{\textit{Timestep sampling (\S\ref{si:training_timestep})}} \\
        Time distribution $t \sim \text{Beta}(\alpha, \beta)$ & $(1.9, 1.0)$ \\
        Uniform mixing probability $p$ & 0.02 \\
        Time clamp & $[10^{-3},\, 1{-}10^{-3}]$ \\
        \midrule
        \multicolumn{2}{l}{\textit{Cropping and batching (\S\ref{si:training_batching})}} \\
        Max tokens per structure & 512 \\
        Train batch size (per GPU) & 64 \\
        Samples per epoch & 819{,}200 \\
        \midrule
        \multicolumn{2}{l}{\textit{Geometric augmentation (\S\ref{si:training_augmentations})}} \\
        SO(3) rotation &  \\
        Translation $\mathcal{N}(0, \sigma^2 \bm{I})$, $\sigma$ (\angstrom) & 3.0 \\
        \midrule
        \multicolumn{2}{l}{\textit{Motif conditioning (\S\ref{si:training_augmentations})}} \\
        $p_\text{freeze HETATM}$ & 0.5 \\
        $p_\text{add residue motif}$ & 0.8 \\
        Max motif residues & 20 \\
        $p_\text{contiguous segment}$ (vs.\ random) & 0.5 \\
        $p_\text{tip mode}$ (vs.\ all-atom) & 0.75 \\
        Bond expansion around seed atom & Pr$(n, p=0.25)$ \\
        $p_\text{centre on motif COM}$ & 0.5 \\
        \midrule
        \multicolumn{2}{l}{\textit{Feature masking (\S\ref{si:training_augmentations})}} \\
        $p_\text{RASA mask}$ & 0.5 \\
        $p_\text{per-residue DSSP mask}$ & 0.5 \\
        $p_\text{global DSSP ratio mask}$ & 0.5 \\
        $p_{R_g\text{ mask}}$ & 0.5 \\
        
        \bottomrule
    \end{tabular}
\end{table}

\section{Training and inference efficiency}
\label{si:efficiency}

Table~\ref{tab:training_cost} compares training costs. \emyx trains in $\approx3.5$ days on a single node of 8 NVIDIA H200 GPUs ($681.6$ GPU-hours), a ${\approx}4\times$ reduction compared with \rfdthree, which requires $7$ days on 16 H200 GPUs (${\approx}2{,}688$ GPU-hours)~\citep{rfdiffusion3}. \proteinacomplexa uses a stagewise training procedure totalling ${\approx}1$M gradient steps on 48-96 A100 GPUs~\citep{geffner2026proteinacomplexa}; wall-clock time is not reported, but the number of steps and GPUs suggest a compute budget that substantially exceeds both \rfdthree and \emyx.

\begin{figure}[H]
    \centering
    \includegraphics[width=0.6\linewidth]{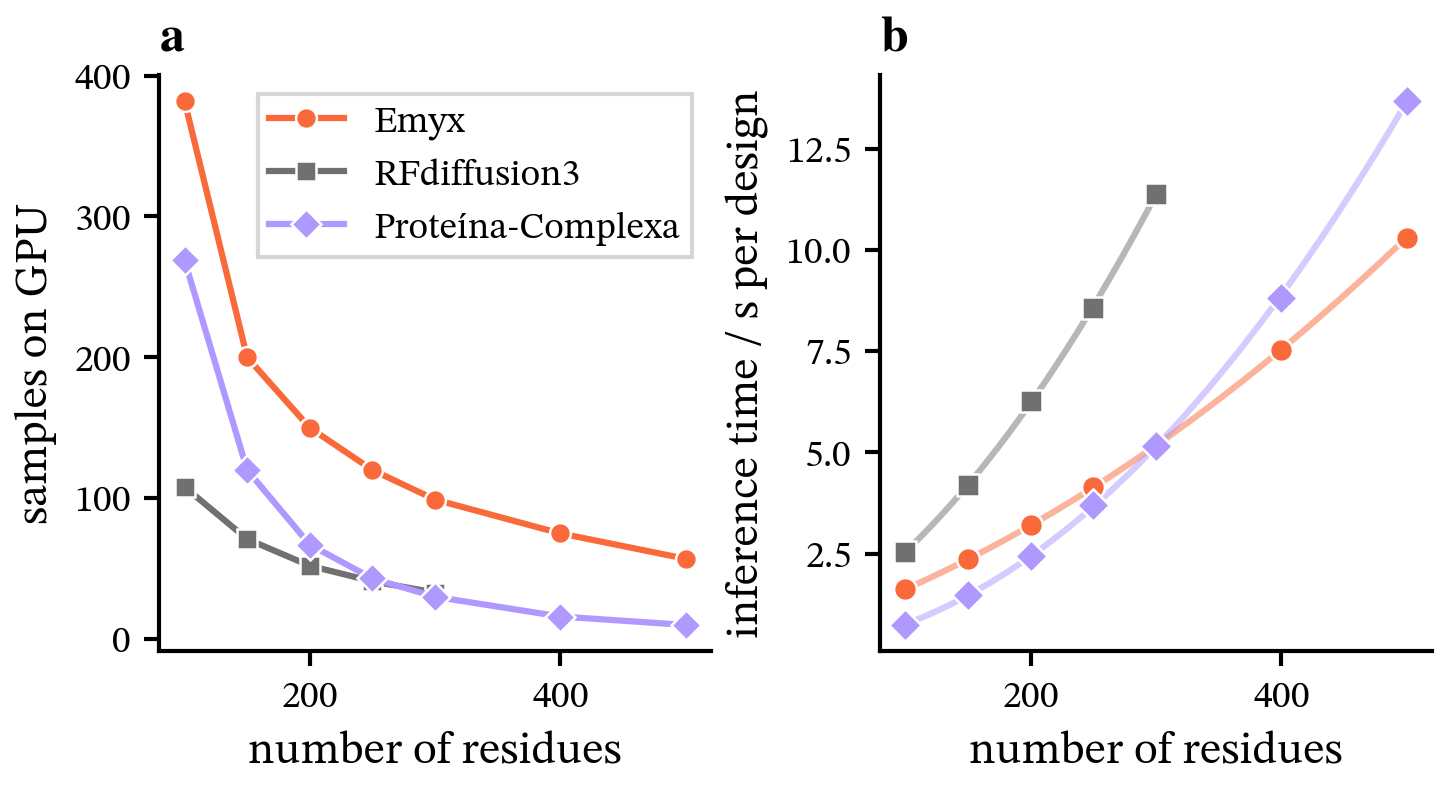}
    \caption{%
        \textbf{Inference efficiency comparison.} Single NVIDIA A10G GPU (24\,GB, bf16-mixed); all timings use unconditional sampling with random-initialised weights matching each model's published architecture. \textbf{a}, Maximum inference batch size (forward pass only, no gradient tracking). \textbf{b}, Wall-clock time for 200 sampling steps per design at GPU-saturated batch size. \emyx (140M parameters) and \rfdthree (168M) share identical trunk dimensions (18 transformer blocks, 768 token dimension); \proteinacomplexa (170M latent denoiser + 128M all-atom decoder = 298M) uses 14 transformer blocks at 768 token dimension with an additional 8-layer decoder (256 dim) to reconstruct all-atom coordinates from its latent representation. \proteinacomplexa timings include the decoder pass. Curves show quadratic fits. \rfdthree runs out of memory at ${\geq}400$ residues on an A10G.
    }
    \label{fig:speed}
\end{figure}

At inference time (Figure~\ref{fig:speed}), \emyx is faster than \rfdthree at all chain lengths on a single A10G GPU, with the gap widening at longer sequences. \proteinacomplexa's denoiser alone is comparable in speed to \emyx (both use sparse attention without pair updates), but its two-stage architecture requires an additional 8-layer decoder pass to reconstruct all-atom coordinates from the latent representation; including the decoder, \proteinacomplexa is comparable to \emyx at short lengths but slower above ${\approx}300$ residues. \rfdthree is the slowest of the three, likely due to its fully-connected token initialisation, which computes dense pair representations scaling as $O(N^2)$ in both time and memory, and its updates of the pair representation. \emyx does not update its pair representation and avoids fully-connected layers entirely; while its edge-construction module does scale quadratically with atom count, this cost is small relative to the transformer module and does not dominate until much larger sequence lengths.
\begin{table}[H]
    \caption{%
        \textbf{Training cost comparison.} GPU-hours computed as (number of GPUs) $\times$ (wall-clock hours). \rfdthree training time as reported by the authors~\citep{rfdiffusion3}. \proteinacomplexa does not report wall-clock time; ${\approx}1$M gradient steps on 48-96 A100s~\citep{geffner2026proteinacomplexa} suggest a substantially larger compute budget, but a direct GPU-hour comparison is not possible.
    }
    \label{tab:training_cost}
    \centering
    \small
    \begin{tabular}{lrrrr}
        \toprule
        Method & GPUs & Wall time & GPU-hours & Relative cost \\
        \midrule
        \proteinacomplexa & 48--96 $\times$ A100 & -- & -- & -- \\
        \rfdthree & 16 $\times$ H200 & 7 days & 2,688 & 1.0$\times$ \\
        \emyx     & 8 $\times$ H200  & 3.55 days &  681.6 & $0.254\times$ \\
        \bottomrule
    \end{tabular}
\end{table}

\section{Sampling}
\label{si:sampling}

\begin{figure}[H]
    \centering
    \includegraphics[width=\linewidth]{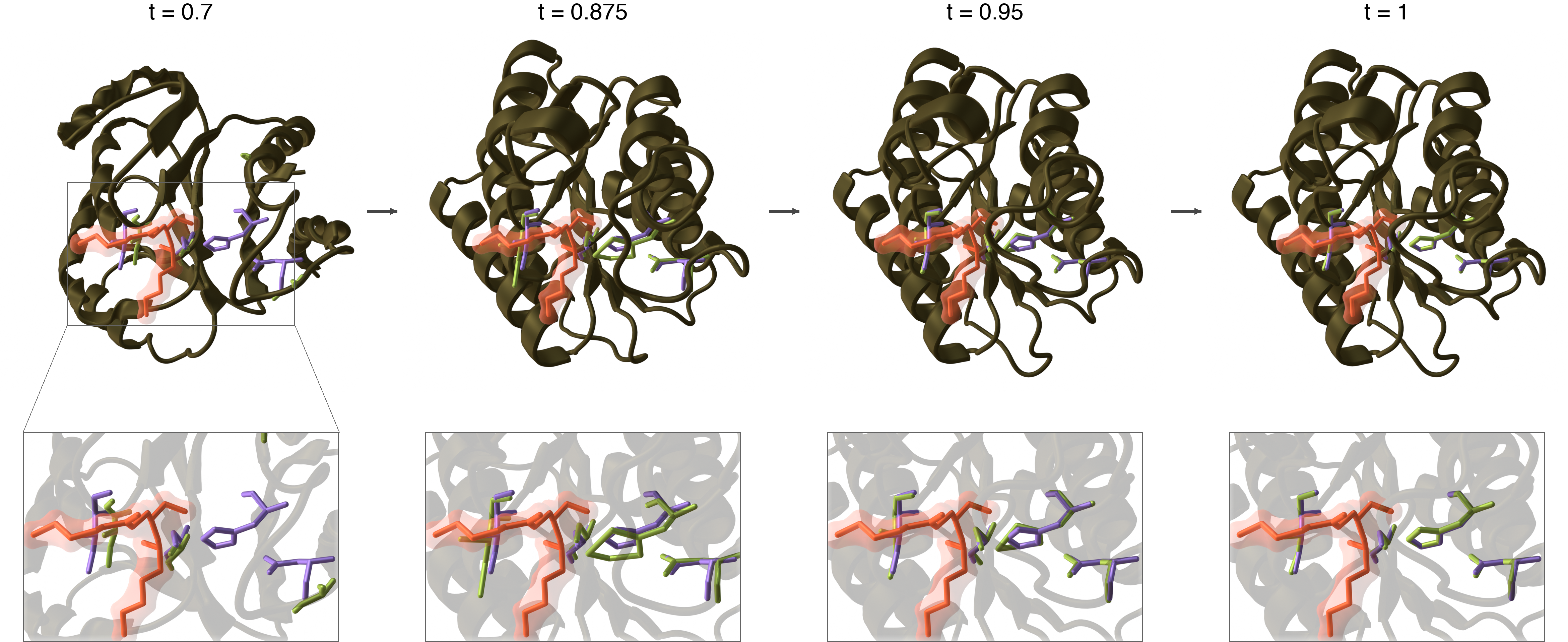}
    \caption{%
        \textbf{Flow matching $\hat{\bm{x}}$ prediction along sampling trajectory.}
        Top: full protein structure at four points during sampling ($t = 0.70$, $0.875$, $0.95$, $1.0$; 200 total steps). Motif residues (\textcolor{figmotif}{purple}), generated residues (\textcolor{figflow}{green}), ligand (\textcolor{figligand}{orange} sticks + mesh).
        Bottom: active-site detail at the same timesteps, showing progressive refinement of sidechain geometry around the fixed catalytic motif.
    }
    \label{fig:flow_sampling}
\end{figure}

This section gives the full derivations and algorithms for the samplers summarised in \S\ref{sec:sampling}: Euler integration of the flow matching ODE (\S\ref{si:sampling_ode}), Euler--Maruyama integration of an equivalent SDE (\S\ref{si:sampling_sde}), and the EDM reparametrisation (\S\ref{si:sampling_edm}). Figure~\ref{fig:flow_sampling} illustrates the progressive denoising along a sampling trajectory.

\subsection{Euler integration of the ODE}
\label{si:sampling_ode}

The simplest sampler integrates the learned velocity field directly via a simple Euler scheme,
\begin{equation}
    \bm{x}_{t + \Delta t} = \xt + \Delta t \cdot \bm{v}_\theta(\xt, t, \mathbf{c}, G),
    \label{eq:euler}
\end{equation}
with a time schedule $t(s)$ mapping uniformly-spaced pseudo-time $s$ to flow time $t$. We use a logarithmic schedule $t(s) = 1 - 10^{-2s}$, which concentrates integration steps near $t = 1$ where fine structural details emerge.

\subsection{Euler--Maruyama integration of the SDE}
\label{si:sampling_sde}

Flow matching models learn to transport a noise distribution $p_0$ to a data distribution $p_1$ via an ODE. However, as shown by~\citet{albergo2023stochastic_interpolants} and~\citet{singh2024stochastic}, there exists a family of stochastic differential equations (SDEs) that share the same marginal distributions $p_t$ as the deterministic ODE but introduce controlled stochasticity during sampling. This added noise can improve sample quality by allowing the sampler to correct accumulated integration errors and explore the distribution more broadly.

For a flow matching model, the general family of SDEs with identical marginals is ~\citep{singh2024stochastic},
\begin{equation}
    d\xt = \biggl[\bm{v}_\theta(\xt, t) + \frac{\tilde{g}(t)^2}{2} \nabla_{\xt} \log p_t(\xt)\biggr] dt + \tilde{g}(t)\, d\mathbf{W}_t,
    \label{eq:sde_general}
\end{equation}
where $\tilde{g}(t)$ is a freely chosen, time-dependent diffusion coefficient, $\nabla_{\xt} \log p_t(\xt)$ is the score function of the time-marginal, and $d\bm{W}_t$ is a Wiener process. The key observation is that the score can be recovered directly from the learned velocity field without training a separate score model. In what follows, the derivations are presented for normalised coordinates so that $\sigma_\text{data} = 1$ and where the noise is centred at the origin ($\bm{\mu} = \bm{0}$). For the linear interpolant $\xt = (1 - t)\bm\epsilon + t\bm{x}$ with $\bm\epsilon \sim \mathcal{N}(\bm{0}, \mathbf{I})$, the conditional distribution is $p(\xt \mid \bm{x}) = \mathcal{N}(t\bm{x}, (1-t)^2 \mathbf{I})$, giving the conditional score,
\begin{equation}
    \nabla_{\xt}\log p(\xt \mid \bm{x})
    = -\frac{\xt - t\bm{x}}{(1-t)^2}.
    \label{eq:cond_score}
\end{equation}
Substituting the endpoint prediction $\hat{\bm{x}} = \xt + (1-t)\bm{v}_\theta$, the marginal score estimate becomes:
\begin{equation}
    \nabla_{\xt}\log p_t(\xt)
    \approx \frac{t\,\bm{v}_\theta(\xt, t) - \xt}{1-t}.
    \label{eq:score_from_v}
\end{equation}

\paragraph{Choice of diffusion coefficient.}
The function $\tilde{g}(t)$ is a design choice that controls the noise--accuracy trade-off. Several parameterisations have been proposed in the literature. In this work, we use the SNR-proportional~\citep{wang2025simplefold} schedule $\tilde{g}(t)^2 = 2 (1-t)/(t + \delta)$ with small $\delta > 0$ for numerical stability and a cut-off at $t \geq 0.99$. This scales the noise injection inversely with the signal-to-noise ratio of the interpolant: more noise is injected early (low SNR) and less near the data (high SNR).
Substituting~Eq.~\ref{eq:score_from_v} into the general SDE (Eq.~\ref{eq:sde_general}) and discretising via Euler-Maruyama yields the update rule,
\begin{equation}
    \Delta\xt
    = \Delta t \cdot \bm{v}_\theta
    + \frac{\tilde{g}(t)^2}{2}\Delta t \cdot
        \frac{t\,\bm{v}_\theta - \xt}{1-t}
    + \tilde{g}(t)\sqrt{\eta \Delta t}\;\bm{\epsilon},
    \label{eq:sde}
\end{equation}
where $\bm{\epsilon} \sim \mathcal{N}(\bm{0}, \mathbf{I})$. A scaling factor $\eta$ is usually artificially introduced which reduces the noise scale with respect to the score term and has been shown to improve sample quality at the cost of diversity. We use $\eta=0.2$, following~\citet{wang2025simplefold} who found this value optimal for flow matching protein structure prediction.

\subsection{EDM reparametrisation}
\label{si:sampling_edm}

\paragraph{The EDM framework.}
The EDM framework of~\citet{karras2022edm} was developed for diffusion models and provides two key innovations: (i) a noise schedule that optimally distributes integration steps across noise levels, and (ii) a stochastic churn mechanism that injects controlled noise during sampling. We show that the flow matching interpolant admits a natural reparametrisation into the EDM framework, allowing us to leverage both mechanisms without any retraining or architectural changes. The reparametrisation is model-agnostic: it applies to any flow matching model using a linear interpolant, requiring only that the velocity network be evaluable at the corresponding flow time. We are not aware of prior work explicitly reparametrising a flow matching velocity model into the EDM noise-level framework for protein structure generation.

As in \S\ref{si:sampling_sde}, all derivations in this subsection are written in normalised coordinates ($\sigma_\text{data} = 1$). In the EDM framework, the generative process operates in terms of a noise level $\sigma$ that decreases from $\sigma_\text{max}$ (pure noise) to $\sigma_\text{min}$ (clean data). The central object is a \emph{denoiser} $D(\bm{y}_\sigma, \sigma)$ that takes a noisy sample $\bm{y}_\sigma = \bm{x} + \sigma\bm\epsilon$ and returns an estimate of the clean data $\hat{\bm{x}}$. The probability flow ODE in $\sigma$-space is,
\begin{equation}
    \frac{d\bm{y}_\sigma}{d\sigma}
    = \frac{\bm{y}_\sigma - D(\bm{y}_\sigma, \sigma)}{\sigma}.
    \label{eq:edm_ode}
\end{equation}
The numerator $\bm{y}_\sigma - D(\bm{y}_\sigma, \sigma)$ points from the denoised estimate back towards the current (noisy) sample. Integration proceeds by stepping from large $\sigma$ to small $\sigma$: at each step the denoiser predicts the clean data, and an Euler step moves $\bm{y}$ toward the next (lower) noise level.

\paragraph{State equivalence.}
We now establish the correspondence between the flow matching state $\xt$ and the EDM state $\bm{y}_\sigma$. The noise distribution is $\bm\epsilon \sim \mathcal{N}(\bm{0}, \mathbf{I})$. Dividing both sides of the linear interpolant by the time $t$:
\begin{equation}
    \frac{\xt}{t} = \bm{x} + \underbrace{\frac{1-t}{t}}_{\sigma(t)}\,\bm\epsilon.
    \label{eq:state_equiv}
\end{equation}
Since $\bm\epsilon \sim \mathcal{N}(\bm{0}, \mathbf{I})$, we identify:
\begin{equation}
    \bm{y}_\sigma= \frac{\xt}{t}, \qquad
    \sigma(t) = \frac{1-t}{t},
    \label{eq:sigma_edm}
\end{equation}
so that $\bm{y}_\sigma= \bm{x} + \sigma\,{\bm\epsilon}$, which is exactly the EDM noise model. As $t$ increases from 0 to 1, $\sigma$ decreases from $\infty$ to 0, matching the EDM convention of denoising from high to low noise.

\paragraph{Denoiser--velocity equivalence.}
The EDM denoiser maps a noisy sample to the predicted clean data: $D(\bm{y}_\sigma, \sigma) = \hat{\bm{x}}$. In flow matching, the endpoint prediction gives $\hat{\bm{x}} = \xt + (1-t)\,\bm{v}_\theta$. Therefore, the flow matching velocity model can be used as an EDM denoiser via:
\begin{equation}
    D(\bm{y}_\sigma, \sigma)
    = \hat{\bm{x}}
    = \xt + (1-t)\,\bm{v}_\theta(\xt, t)
    = t\,\bm{y}_\sigma+ (1-t)\,\bm{v}_\theta\!\bigl(t\,\bm{y}_\sigma,\; t\bigr),
    \label{eq:denoiser_equiv}
\end{equation}
where $t = 1 / (1 + \sigma)$ from inverting Eq.~\ref{eq:sigma_edm}. This equivalence is exact: any velocity-predicting flow matching model can be wrapped as an EDM denoiser by (i) converting $\sigma \to t$, (ii) computing $\xt = t \cdot \bm{y}_\sigma$, (iii) evaluating the velocity $\bm{v}_\theta$, and (iv) forming the endpoint prediction $\hat{\bm{x}}$. The reverse direction is equally straightforward: given a denoiser output $\hat{\bm{x}}$, the velocity is $\bm{v}_\theta = (\hat{\bm{x}} - \xt) / (1-t)$.

\paragraph{Karras noise schedule.}
The Karras schedule~\citep{karras2022edm} distributes the $n$ integration steps across noise levels as:
\begin{equation}
    \sigma_i
    =
    \Bigl(
        \sigma_\text{max}^{1/\rho}
        + \frac{i}{n-1}
        \bigl(\sigma_\text{min}^{1/\rho} - \sigma_\text{max}^{1/\rho}\bigr)
    \Bigr)^{\!\rho},
    \qquad i = 0, 1, \ldots, n,
    \label{eq:karras}
\end{equation}
where $\rho$ controls the distribution of steps: larger $\rho$ concentrates more steps at low noise levels (fine detail), while $\rho \to 0$ gives uniform spacing in $\sigma$. We refer to \citet{karras2022edm} for the specific parameter values.

\paragraph{Stochastic churn.}
The deterministic ODE (Eq.~\ref{eq:edm_ode}) can be augmented with stochastic noise injection~\citep{karras2022edm}. Before each denoising step at noise level $\sigma_i$, a noise injection step increases the effective noise level,
\begin{equation}
    \hat{\sigma}_i = \sigma_i(1 + \gamma), \qquad
    \hat{\bm{y}} = \bm{y}_\sigma+ s_\text{noise}\sqrt{\hat{\sigma}_i^2 - \sigma_i^2}\;\bm\epsilon,
    \label{eq:churn}
\end{equation}
where $\gamma$ is the churn amplitude (set to 0 for deterministic sampling), $s_\text{noise}$ is a noise amplification factor, and $\bm\epsilon \sim \mathcal{N}(\bm{0}, \mathbf{I})$. Churn is applied only when $\sigma_i$ exceeds a threshold $\sigma_\text{min churn}$. The denoiser is then evaluated at $\hat{\sigma}_i$, and the Euler step proceeds from $\hat{\sigma}_i$ to $\sigma_{i+1}$.

\paragraph{Full algorithm.}
The complete EDM sampling procedure is given in Algorithm~\ref{alg:edm}. We use the following hyperparameters: $\sigma_\text{max} = 160$, $\sigma_\text{min} = 0.0004$, $\rho = 7$, $\gamma = 0.6$, $s_\text{noise} = 1.003$, $s_\text{step} = 1.5$, $\sigma_\text{min churn} = 1.0$.

\begin{algorithm}[t]
\caption{EDM sampling from a flow matching model}
\label{alg:edm}
\small
\begin{algorithmic}[1]
    \REQUIRE Velocity model $\bm{v}_\theta$, schedule $\{\sigma_i\}_{i=0}^{n}$ (Eq.~\ref{eq:karras}), churn $\gamma$, noise scale $s_\text{noise}$, step scale $s_\text{step}$, threshold $\sigma_\text{min churn}$
    \STATE Sample $\bm\epsilon \sim \mathcal{N}(\bm{0}, \mathbf{I})$; set $\bm{y}_\sigma\leftarrow \sigma_\text{max}\,\bm\epsilon$ \hfill \COMMENT{equivalently, $\xt = (1-t_0)\bm\epsilon$}
    \FOR{$i = 0$ \textbf{to} $n - 1$}
        \STATE $\hat{\sigma}_i \leftarrow \sigma_i$
        \IF{$\sigma_{i+1} > \sigma_\text{min churn}$}
            \STATE $\hat{\sigma}_i \leftarrow \sigma_i (1 + \gamma)$ \hfill \COMMENT{stochastic churn}
            \STATE $\bm{y}_\sigma\leftarrow \bm{y}_\sigma+ s_\text{noise}\sqrt{\hat{\sigma}_i^2 - \sigma_i^2}\;\bm\epsilon$, \quad $\bm\epsilon \sim \mathcal{N}(\bm{0}, \mathbf{I})$
        \ENDIF
        \STATE $t \leftarrow 1 / (1 + \hat{\sigma}_i)$ \hfill \COMMENT{EDM $\to$ flow time}
        \STATE $\xt \leftarrow t \cdot \bm{y}_\sigma$ \hfill \COMMENT{EDM $\to$ flow state}
        \STATE $D \leftarrow \xt + (1 - t) \cdot \bm{v}_\theta(\xt, t)$ \hfill \COMMENT{denoised estimate via velocity}
        \STATE $\bm{y}_\sigma\leftarrow \bm{y}_\sigma + s_\text{step} (\sigma_{i+1} - \hat{\sigma}_i)\, (\bm{y}_\sigma - D)/\hat{\sigma}_i$ \hfill \COMMENT{Euler step (Eq.~\ref{eq:edm_ode})}
    \ENDFOR
    \STATE \textbf{return} $D$ \hfill \COMMENT{final denoised estimate $\approx \hat{\bm{x}}$}
\end{algorithmic}
\end{algorithm}

\subsection{EDM vs.\ Euler sampling comparison}
\label{si:edm_comparison}

\begin{figure}[H]
    \centering
    \includegraphics[width=\linewidth]{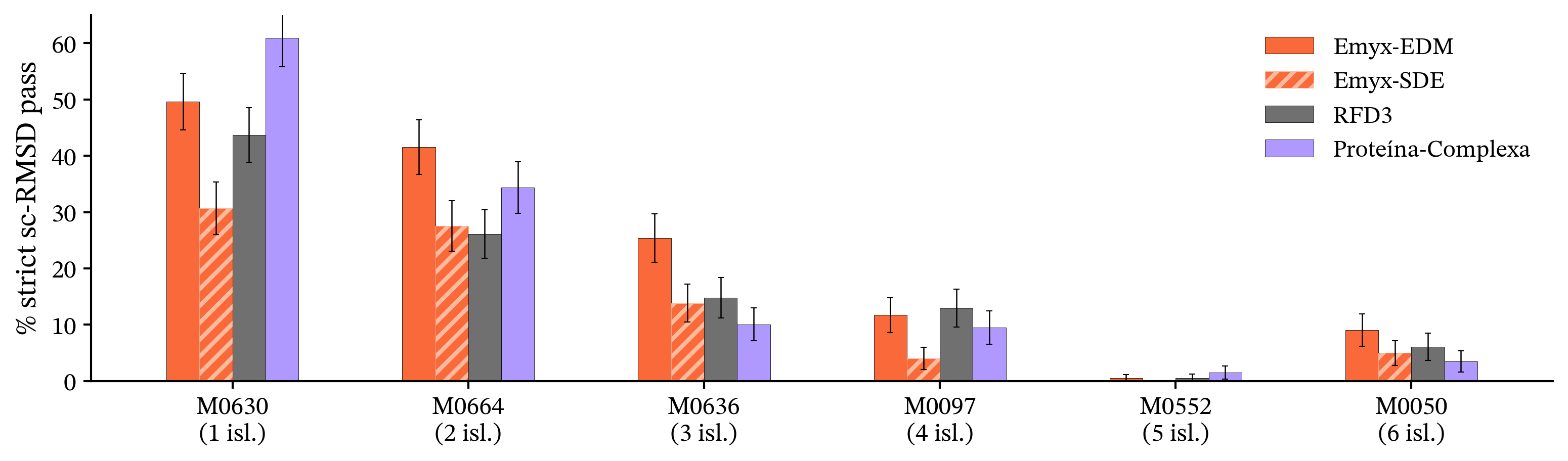}
    \caption{%
        \textbf{Per-target sampler comparison on AME} (strict sc-RMSD; 6 shared targets).
        Strict sc-RMSD success rate for EDM and SDE samplers, \rfdthree, and \proteinacomplexa on each of the 6 targets, ordered by residue island count.
        Bootstrap mean $\pm 1\sigma$ ($1{,}000$ resamples of 100 designs).
        The Euler ODE sampler (not shown) achieves $0\%$ on all 6 targets.
    }
    \label{fig:sampler_per_target}
\end{figure}

\section{Evaluation details}
\label{si:eval_details}

\subsection{AME benchmark}
\label{si:ame_benchmark}
\label{sec:benchmark}

We evaluate motif-conditioned scaffolding against the AME benchmark~\citep{Ahern2026-ob-rfdiffusion2, rfdiffusion3}. The benchmark comprises 41 catalytic active sites spanning Enzyme Commission (EC) classes 1--5, with between 1 and 7 \emph{residue islands} (contiguous segments of catalytic residues) serving as a proxy for scaffolding difficulty.\footnote{Active sites are derived from the Mechanism and Catalytic Site Atlas (M-CSA) cross-referenced with the PARITY dataset; see~\citet{rfdiffusion3} for the full target selection procedure.}

\subsection{Self-consistency protocol}
\label{si:sc_protocol}

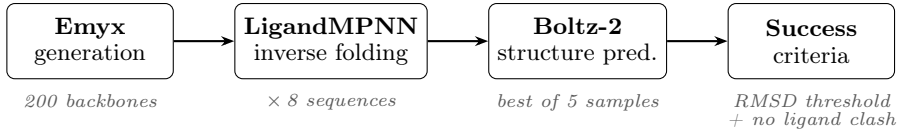
\begin{figure}[H]
    \centering
    \begin{tikzpicture}[
        node distance=0.6cm and 0.8cm,
        box/.style={draw, rounded corners=3pt, minimum height=1cm, minimum width=2.2cm,
                     align=center, font=\small},
        arr/.style={-{Stealth[length=5pt]}, thick},
        ann/.style={font=\scriptsize\itshape, text=gray!70!black, align=center},
    ]
        \node[box] (gen) {\textbf{\emyx}\\[-1pt] generation};
        \node[box, right=of gen] (inv) {\textbf{LigandMPNN}\\[-1pt] inverse folding};
        \node[box, right=of inv] (fold) {\textbf{Boltz-2}\\[-1pt] structure pred.};
        \node[box, right=of fold] (sc) {\textbf{Success}\\[-1pt] criteria};

        \draw[arr] (gen) -- (inv);
        \draw[arr] (inv) -- (fold);
        \draw[arr] (fold) -- (sc);

        \node[ann, below=2pt of gen] {200 backbones};
        \node[ann, below=2pt of inv] {$\times$\,8 sequences};
        \node[ann, below=2pt of fold] {best of 5 samples};
        \node[ann, below=2pt of sc] {RMSD threshold\\[-1pt] + no ligand clash};
    \end{tikzpicture}
    \caption{\textbf{Self-consistency evaluation pipeline} for motif-conditioned scaffolding.}
    \label{fig:eval_pipeline}
\end{figure}

We evaluate each generated backbone through a multi-stage self-consistency pipeline that tests whether the structure admits a foldable sequence capable of recovering the catalytic geometry upon structure re-prediction. Figure~\ref{fig:eval_pipeline} summarises the pipeline.

\begin{figure}[H]
    \centering
    \includegraphics[width=0.85\linewidth]{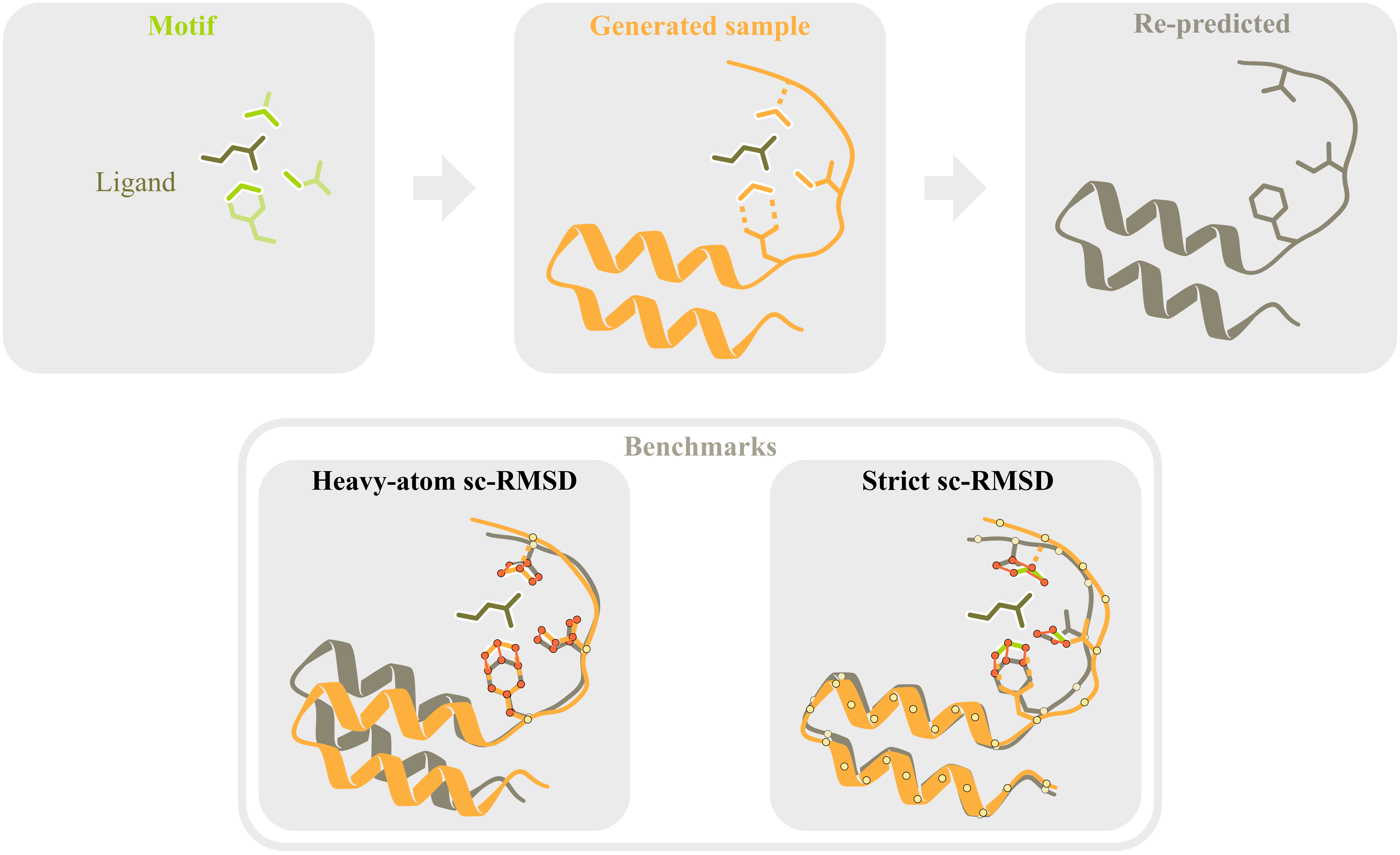}
    \caption{%
        \textbf{Self-consistency evaluation.} \emph{Top}: a motif and ligand are scaffolded with a generated structure, which is then sequence-redesigned and re-predicted. When the model fails to precisely overlap the motif, overwriting the motif tip atoms with the input motif creates unphysical bonds and angles (dashed lines). \emph{Bottom}: The heavy-atom sc-RMSD criterion aligns on the motif backbone only and computes the RMSD over the motif residues' heavy atoms; this biases the RMSD metric via good alignment on the backbone of the motif but poor alignment on the tip atoms. The strict sc-RMSD criterion aligns on the full protein backbone and computes the RMSD for the entire backbone and the motif tip atoms; this not only eliminates the bias in the motif RMSD but additionally tests for global fold recovery.
    }
    \label{fig:sc_rmsd_schematic}
\end{figure}

The pipeline proceeds as follows:

\begin{enumerate}
    \item \textbf{Backbone generation.} For each motif target, \emyx generates 200 backbone structures using the EDM sampler (Algorithm~\ref{alg:edm}) with 200 integration steps.\footnote{Two critical parameters for reproducing the benchmark are not explicitly specified in~\citep{rfdiffusion3} but only in their code: (i) the generated chain length is fixed at 180 residues, and (ii) the initial noise distribution is centred on the centre of mass of the motif atoms. We adopt both conventions.}

    \item \textbf{Inverse folding.} LigandMPNN~\citep{dauparas2025ligandmpnn} redesigns the sequence for each generated backbone, producing 8 candidate sequences per structure at a sampling temperature of $0.1$. The ligand and motif residue identities are held fixed; only scaffold positions are redesigned.

    \item \textbf{Structure prediction.} Each candidate sequence is folded 5 times by Boltz-2~\citep{passaro2025boltz2} with MSA input enabled. The prediction with the highest pTM score is retained as the representative fold for that sequence.

    \item \textbf{Success criteria.} As discussed in \S\ref{sec:experimental_setup}, we evaluate three approaches to the success critera that differ in (i) how the designed and re-predicted structures are aligned and (ii) which RMSD values are checked. All three additionally require \emph{no ligand clash}: every inter-atomic distance between HETATM (ligand) atoms and protein atoms in the generated structure exceeds $1.5$\,\angstrom.
    \begin{enumerate}
        \item \emph{Heavy-atom sc-RMSD.} Align the re-predicted structure to the design on the backbone atoms of \emph{motif residues only}. A design passes if (i) the RMSD between the generated and re-predicted structure motif residue heavy-atoms is $<1.5$\,\angstrom{} and (ii) no ligand clash. This is the standard AME criterion used in~\citet{Ahern2026-ob-rfdiffusion2, rfdiffusion3, geffner2026proteinacomplexa}; it tests only that local catalytic geometry is preserved.
        \item \emph{Tip-atom sc-RMSD.} The same alignment as the heavy-atom sc-RMSD. A design passes if (i) the RMSD between the motif tip atoms (the atoms specified in the input motif) of the re-predicted structure and the input motif atoms is $<1.5$\,\angstrom{} and (ii) no ligand clash. This elimiates the RMSD bias from the remaining motif atoms (see Figure~\ref{fig:sc_rmsd_schematic}).
        \item \emph{Strict sc-RMSD.} Align the re-predicted structure to the design on the entire protein backbone. A design passes if (i) the full-backbone RMSD $<2.0$\,\angstrom, (ii) the RMSD of the re-predicted structure and the input motif atoms is $<1.5$\,\angstrom{}, and (iii) no ligand clash. This tests that the global fold is recovered \emph{and} that local catalytic geometry is preserved simultaneously.
    \end{enumerate}
    See \S\ref{sec:backbone_eval} for the motivation behind strict sc-RMSD and Fig.~\ref{fig:sc_examples} for an illustration of designs that satisfy heavy-atom sc-RMSD but fail strict sc-RMSD. Figure~\ref{fig:sc_rmsd_schematic} provides a schematic overview of the self-consistency pipeline and the distinction between the two alignment strategies.
\end{enumerate}

\noindent This protocol yields $200 \times 8 = 1{,}600$ sequence--structure pairs per motif target. A generated backbone is considered to be successfully designed if one or more of its 8 LigandMPNN-redesigned sequences passes the success criterion. To compute confidence intervals we bootstrap: from each target's 200 samples we draw 100 with replacement, compute the success rate on that subsample, and repeat $1{,}000$ times to obtain mean $\pm 1\sigma$. The subsample size of 100 designs matches the standard heavy-atom sc-RMSD design budget~\citep{rfdiffusion3}.  The AME plots in \S\ref{si:ame_per_target} report the metrics for all three success criteria.

\subsection{Novelty and diversity metrics}
\label{si:novelty_diversity}

Beyond motif placement accuracy, we evaluate two properties critical for \emph{de novo} enzyme design: structural novelty (how similar designs are to known PDB proteins) and structural diversity (how many distinct solutions does the model produce per target). Both metrics are computed only on \emph{successful} designs (those passing the success criteria above).

\paragraph{TM-score.}
The template modelling score (TM-score)~\citep{zhang2004tmscore} measures global structural similarity between two protein structures on a length-normalised scale in $[0, 1]$, where $1$ indicates a perfect match and values below ${\approx}0.17$ correspond to unrelated folds. For a target of length $L_\text{target}$ aligned to a query,
\begin{equation}
    \text{TM-score} = \frac{1}{L_\text{target}} \sum_{i=1}^{L_\text{aligned}} \frac{1}{1 + (d_i / d_0)^2}.
    \label{eq:tmscore}
\end{equation}
Here $d_i$ is the C$_\alpha$ distance between aligned residue pair $i$ after optimal superposition and $d_0 = 1.24\,\sqrt[3]{L_\text{target} - 15} - 1.8$ is a length-dependent distance scale that normalises for protein size.

\paragraph{Metrics.}
We use TM-score in two ways: (i)~as a \emph{novelty} metric, where each successful design is searched against the PDB using Foldseek~\citep{vanKempen2024foldseek} and the TM-score to the closest hit is reported (lower = more novel); and (ii)~as a \emph{diversity} metric, where each model's $200$ designs per target are clustered with \texttt{foldseek easy-cluster}, and the number of unique clusters among successful designs is reported (higher = more diverse). The raw cluster count is biased by sample volume (models with more successful samples could yield more clusters by chance); we therefore also report the sample-volume-controlled \emph{unique-clusters per 100 designs} metric (\S\ref{si:clusters_per_100}). Per-model totals are in Table~\ref{tab:ame_results}.

\subsection{Unique-clusters per 100 designs}
\label{si:clusters_per_100}

\begin{figure}[H]
    \centering
    \includegraphics[width=\linewidth]{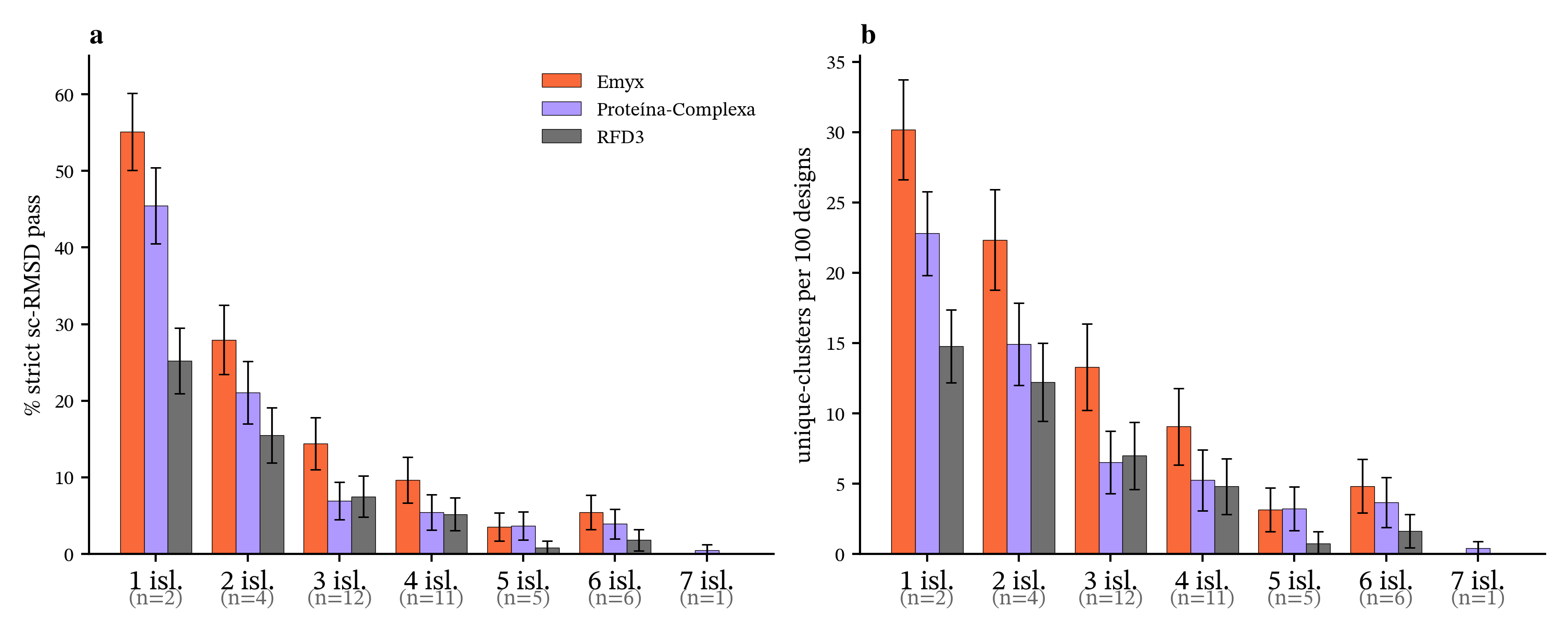}
    \caption{%
        \textbf{Metric comparison} for Figure~\ref{fig:ame_combined} panel (a) under both metrics (strict sc-RMSD).
        \textbf{a}, Raw success rate ($\%$ strict sc-RMSD pass), as reported in the main text.
        \textbf{b}, Unique-clusters per 100 designs (each Foldseek cluster credits at most once per bootstrap draw).
        Both panels show per-island-count groups, bootstrap mean $\pm 1\sigma$ over $1{,}000$ resamples of 100 designs. Model rankings track each other across all island groups, confirming that the two metrics give consistent conclusions in this regime.
    }
    \label{fig:ame_metric_comparison}
\end{figure}

As a diversity-aware companion to the raw success rate, we also report \emph{unique-clusters per 100 designs}, which credits each Foldseek cluster at most once per bootstrap draw so models cannot inflate scores by producing near-identical structures. The procedure is: (i)~for each model, cluster its $200$ designs per target with \texttt{foldseek easy-cluster}, producing per-target cluster labels that are independent across models; (ii)~pool all designs from the targets in an island-count group; (iii)~draw $100$ designs with replacement from the pool and count distinct (target, cluster-label) pairs among the successful draws; (iv)~repeat for $1{,}000$ draws and report mean $\pm 1\sigma$. In this regime the metric tracks the raw success rate closely (Figure~\ref{fig:ame_metric_comparison}), so we report it only in Table~\ref{tab:ame_results}. Under strict sc-RMSD, \emyx attains \edit{$13.1 \pm 3.2$} unique-clusters per 100 designs (raw $13.4\%$), \proteinacomplexa \edit{$8.4 \pm 2.6$} (raw $8.8\%$), and \rfdthree \edit{$6.6 \pm 2.4$} (raw $6.7\%$).

\subsection{Geometry validity}
\label{si:geom_validity}

\begin{figure}[H]
    \centering
    \includegraphics[width=0.5\linewidth]{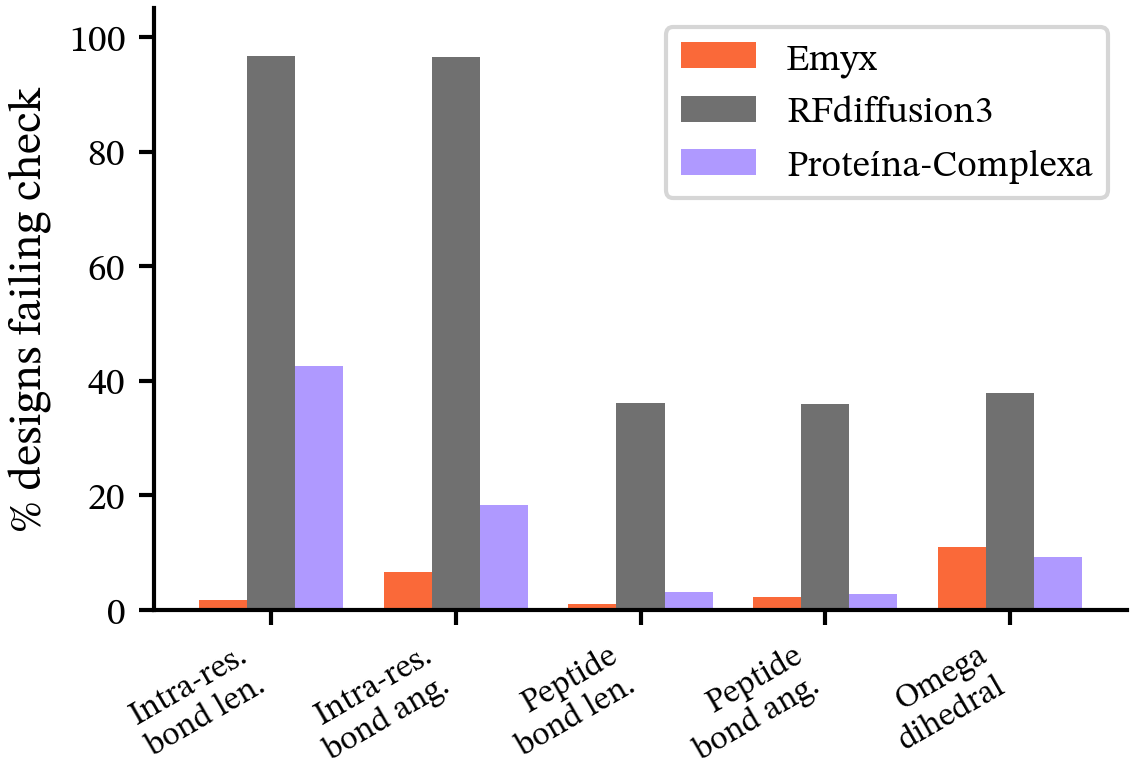}
    \caption{%
        \textbf{Geometry validity breakdown} for \emyx, \rfdthree, and \proteinacomplexa.
        Percentage of designs failing each geometry check. \emyx failures are distributed across omega dihedrals ($11.0\%$), intra-residue bond angles ($6.7\%$), and intra-residue bond lengths ($1.7\%$), with peptide bond checks rarely failing (${\approx}1\text{--}2\%$). \proteinacomplexa shows intermediate geometry quality, with $42.6\%$ of designs failing intra-residue bond lengths and $18.3\%$ failing bond angles, but peptide bond and omega checks near \emyx levels. \rfdthree fails $>96\%$ of designs on intra-residue checks and ${\approx}35\%$ on peptide checks.
    }
    \label{fig:geometry_validity}
\end{figure}

\begin{figure}[H]
    \centering
    \includegraphics[width=\linewidth]{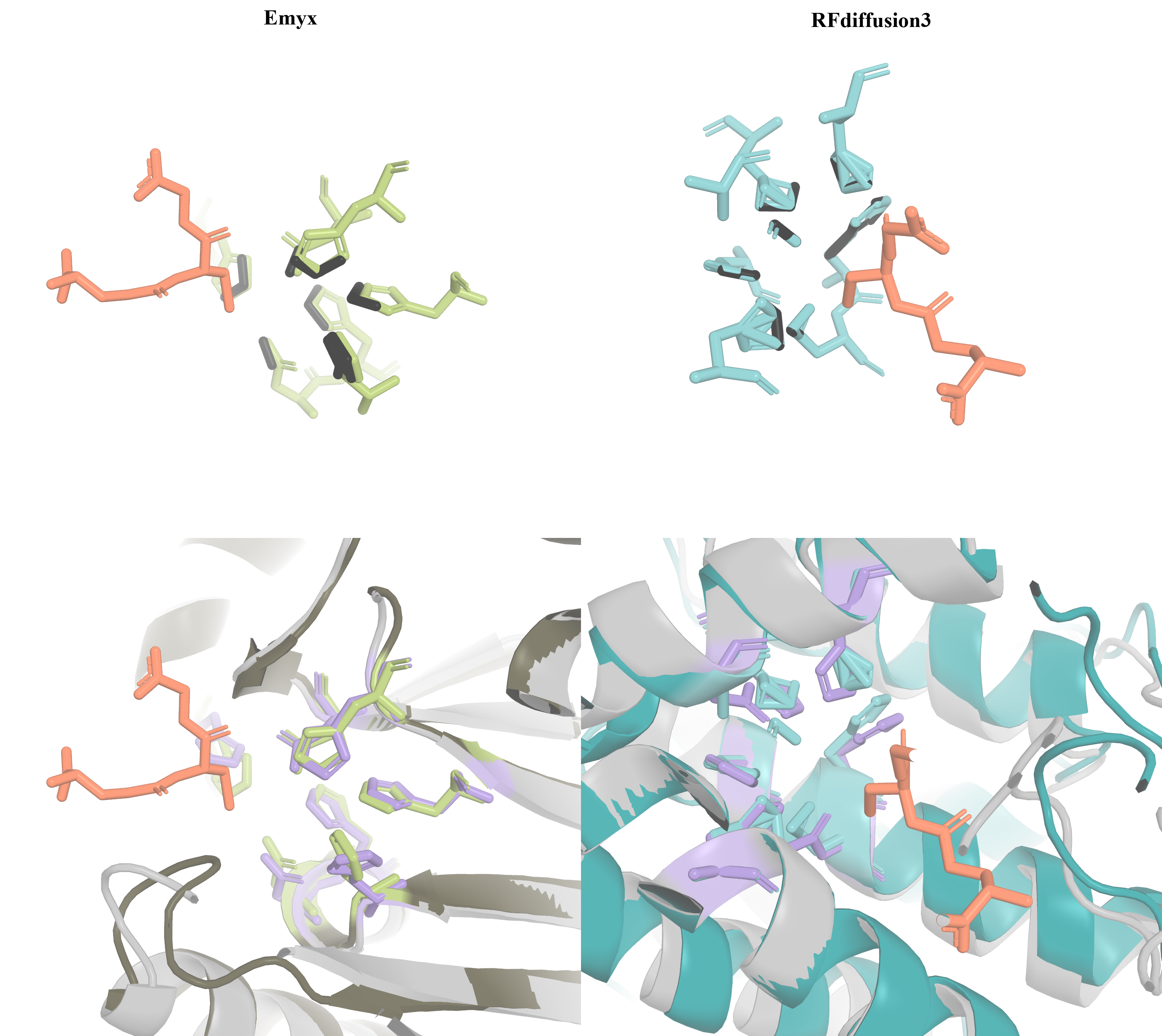}
    \caption{%
        \textbf{Motif geometry comparison.} Representative design for both models on the M0157 protein from the AME benchmark. Both designs pass the motif RMSD $< 1.5$\,\angstrom{} criterion.
        \textbf{Top row}, Input motif atoms (dark sticks) overlaid on the generated motif residues. \emyx (green, left) adheres closely to the specified motif, whilst \rfdthree shows multiple issues including broken residues, unphysical bond lengths, and non-planar aromatic rings.
        \textbf{Bottom row}, same designs with Boltz-2 re-prediction overlaid (grey cartoon, motif sidechains as purple sticks). \emyx and \rfdthree's structures both pass the motif-RMSD criterion.    }
    \label{fig:m0157_motif_closeup}
\end{figure}

Motif RMSD alone can pass even when a generator outputs distorted residues (non-planar aromatic rings, inverted C$_\alpha$ chirality, stretched peptide bonds). We therefore additionally audit motif-region geometry for each generated design against PeptideBuilder~\citep{peptidebuilder} per-residue ideal bond lengths and bond angles. The audit is restricted to a motif-window of residues: every motif residue plus one adjacent residue on each side. For that window we compute, per design, the maximum absolute deviation of (i)~intra-residue bond lengths, (ii)~intra-residue bond angles from their ideals, (iii)~peptide $\text{C}(i) - \text{N}(i+1)$ bond length deviation from $1.33$\,\angstrom, (iv)~peptide $\text{C}_\alpha$--C--N and C--N--$\text{C}_\alpha$ angles deviation from $116.6^\circ$ and $121.4^\circ$ respectively, and (v)~the absolute deviation of the omega torsion from $180^\circ$. A design passes the geometry check when \emph{every} max deviation is within tolerance,
\begin{equation}
\text{dist. deviation} \leq 0.2\,\angstrom, \qquad
\text{angle deviation} \leq 20^\circ, \qquad
\text{omega angle deviation} \leq 30^\circ.
\end{equation}
These thresholds are calibrated against crystal structure data from the PDB as a gold-standard geometric reference: against the AME benchmark, $96.9\%$ of Boltz-2 re-predicted structures pass in the motif window, $82.2\%$ of \emyx samples pass, $49.9\%$ of \proteinacomplexa samples pass, and $0.8\%$ of \rfdthree samples pass. \proteinacomplexa failures are dominated by intra-residue bond length violations ($42.6\%$) and bond angle violations ($18.3\%$), with peptide bond and omega dihedral checks comparable to \emyx. Most of the \rfdthree failures are due to unphysical intra-residue bond lengths, angle distortion in the motif region, and chain breaks at the motif--scaffold peptide joins. We suspect this is caused by the inference code overwriting the generated motif coordinates with the input motif coordinates when \rfdthree fails to precisely overlap the specified input motif. Figure~\ref{fig:m0157_motif_closeup} shows a representative side-by-side comparison.

\section{AME benchmark results}
\label{si:ame_summary}

This section provides per-target breakdowns of the AME benchmark results summarised in \S\ref{sec:res_benchmark}.

\subsection{Per-target success rates and sequence quality}
\label{si:ame_per_target}

\begin{figure}[H]
    \centering
    \includegraphics[width=\linewidth]{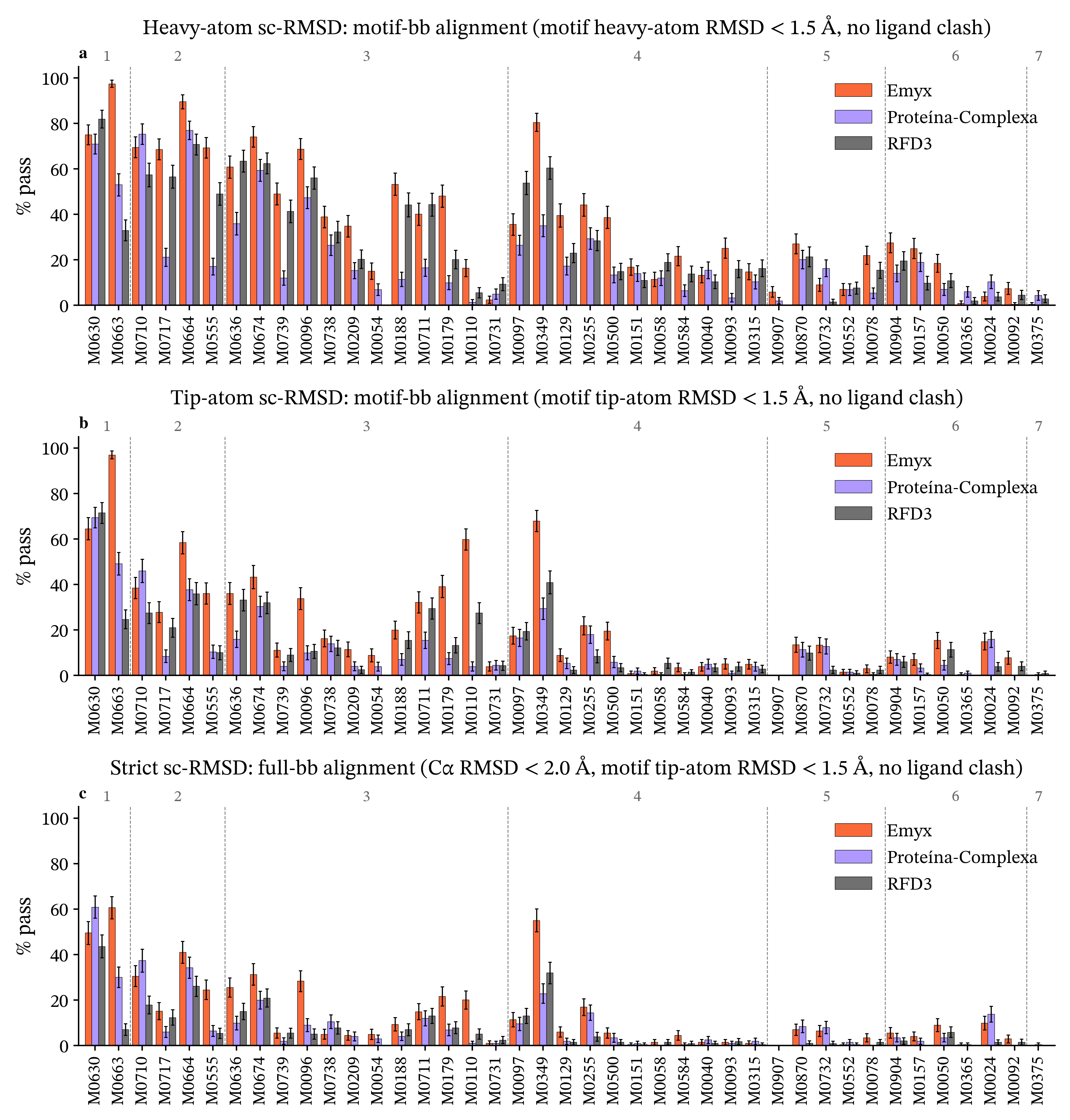}
    \caption{%
        \textbf{Per-target success rates under three evaluation metrics} for \emyx, \rfdthree, and \proteinacomplexa (LigandMPNN-redesigned sequence).
        \textbf{a}, \emph{Heavy-atom sc-RMSD}: motif-backbone alignment with success defined as the RMSD over all motif residue heavy atoms $< 1.5$\,\angstrom{} and no ligand clash.
        \textbf{b}, \emph{Tip-atom sc-RMSD}: motif-backbone alignment with success defined as the motif tip-atom RMSD $< 1.5$\,\angstrom{} and no ligand clash.
        \textbf{c}, \emph{Strict sc-RMSD}: full-backbone alignment with success defined as backbone RMSD $< 2.0$\,\angstrom, motif tip-atom RMSD $< 1.5$\,\angstrom, and no ligand clash.
        Bootstrap mean $\pm 1\sigma$ ($1{,}000$ resamples of 100 designs). Numbers above the bars index the catalytic-residue island groups.
    }
    \label{fig:ame_success_comparison}
\end{figure}

Table~\ref{tab:ame_results} presents the aggregate benchmark results. Under strict sc-RMSD (\S\ref{sec:backbone_eval}), \emyx achieves $13.4\%$ success ($39/41$ targets solved) compared to $8.8\%$ for \proteinacomplexa ($37/41$) and $6.7\%$ for \rfdthree ($35/41$). Figure~\ref{fig:ame_success_comparison} shows the per-target breakdown on the LigandMPNN-redesigned sequence under all three criteria: (\textbf{a}) heavy-atom sc-RMSD (motif-backbone alignment, all motif heavy-atom RMSD $< 1.5$\,\angstrom, no ligand clash), (\textbf{b}) tip-atom sc-RMSD (motif-backbone alignment, motif tip-atom RMSD $< 1.5$\,\angstrom, no ligand clash), and (\textbf{c}) strict sc-RMSD (full-backbone alignment, backbone RMSD $< 2.0$\,\angstrom{} and motif tip-atom RMSD $< 1.5$\,\angstrom, no ligand clash). \emyx is competitive or superior on every island-count group under all three definitions.

\begin{figure}[H]
    \centering
    \includegraphics[width=0.7\linewidth]{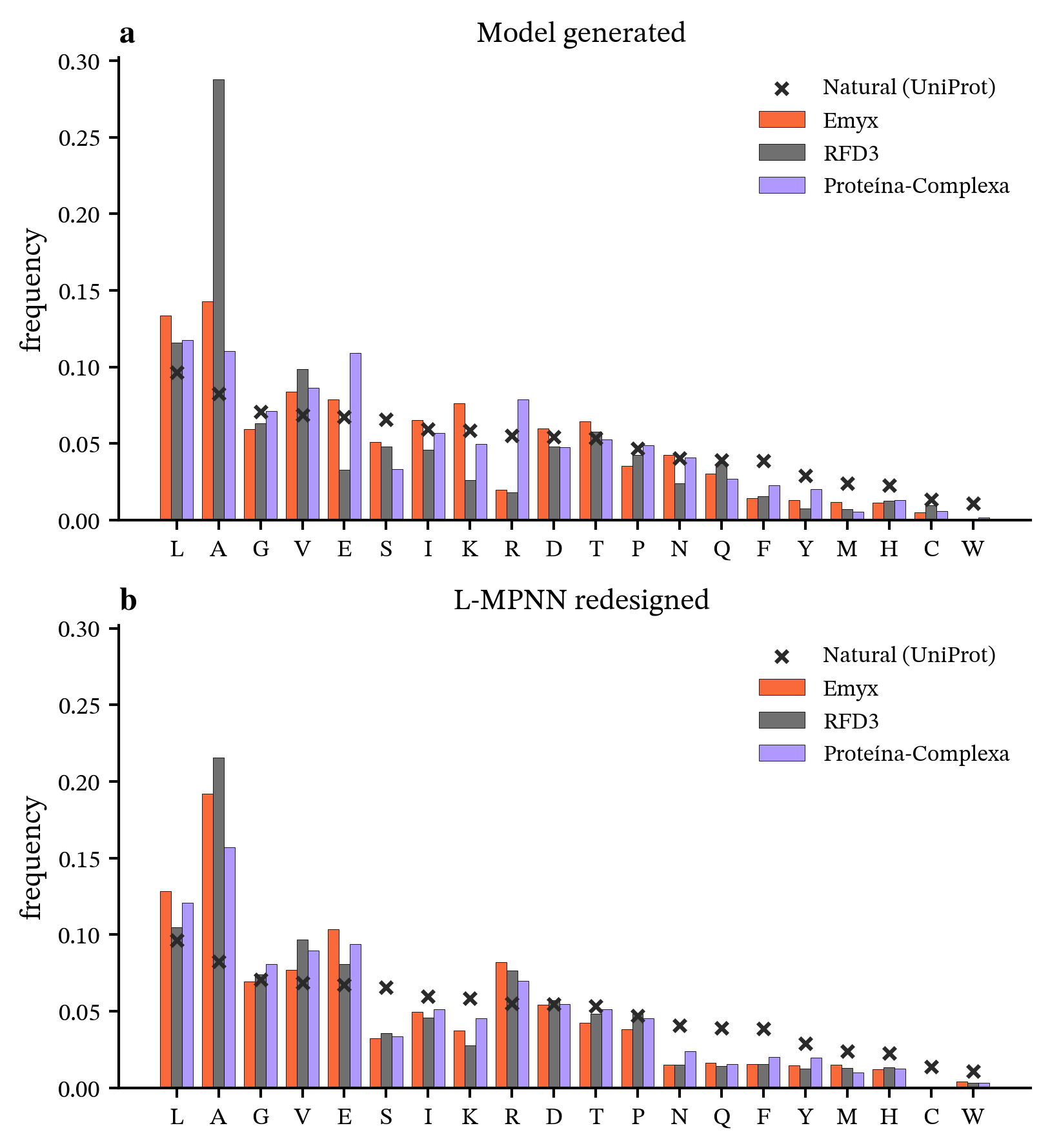}
    \caption{%
        \textbf{Amino acid distributions pooled over all AME targets} for \emyx, \rfdthree, and \proteinacomplexa. `Natural (UniProt)' is the Swiss-Prot background distribution. Residues are ordered by natural frequency.
    }
    \label{fig:ame_aa}
\end{figure}

Figure~\ref{fig:ame_aa} compares the amino acid composition of generated sequences against the UniProt/Swiss-Prot background. \emyx's generated sequence distribution tracks the natural PDB distribution closely across all 20 residues, including rare residues such as Trp, Cys, and Met, with the largest deviations being a modest over-representation of Leu, Glu, and Lys. \proteinacomplexa shows a similar profile to \emyx. LigandMPNN, by contrast, collapses towards ${\approx}20$--$30\%$ Ala (all three generators' backbones produce near-identical redesigned distributions), and depleted in Asn/Gln/Cys/Trp relative to natural proteins.

\begin{table}[H]
    \caption{%
        \textbf{AME benchmark results} (strict sc-RMSD; \S\ref{sec:backbone_eval}). Median TM-to-PDB and raw cluster counts are computed on successful designs only. Unique-clusters / 100 reports the companion metric defined in \S\ref{si:clusters_per_100} (mean $\pm 1\sigma$ over $1{,}000$ bootstrap draws of $100$ designs, pooled across all 41 targets).
    }
    \label{tab:ame_results}
    \centering
    \footnotesize
    \setlength{\tabcolsep}{4pt}
    \begin{tabular}{lcccccc}
        \toprule
        Method & Params & Solved & Success rate $\uparrow$ & Unique-clusters / 100 $\uparrow$ & Median TM-to-PDB $\downarrow$ & Raw clusters $\uparrow$ \\
        \midrule
        \rfdthree          & 168M & 35/41          & $6.7\%$          & \edit{$6.6 \pm 2.4$} & $0.512$ & \edit{$242$} \\
        \proteinacomplexa  & $298$M & 37/41          & $8.8\%$          & \edit{$8.4 \pm 2.6$} & \edit{$0.499$} & \edit{$292$} \\
        \emyx              & 140M & \textbf{39/41} & \textbf{13.4\%}  & \edit{$\mathbf{13.1 \pm 3.2}$} & \edit{\textbf{0.475}} & \edit{\textbf{441}} \\
        \bottomrule
    \end{tabular}
\end{table}

\subsection{Per-target novelty and diversity breakdown}
\label{si:ame_per_protein}

\begin{figure}[H]
    \centering
    \includegraphics[width=\linewidth]{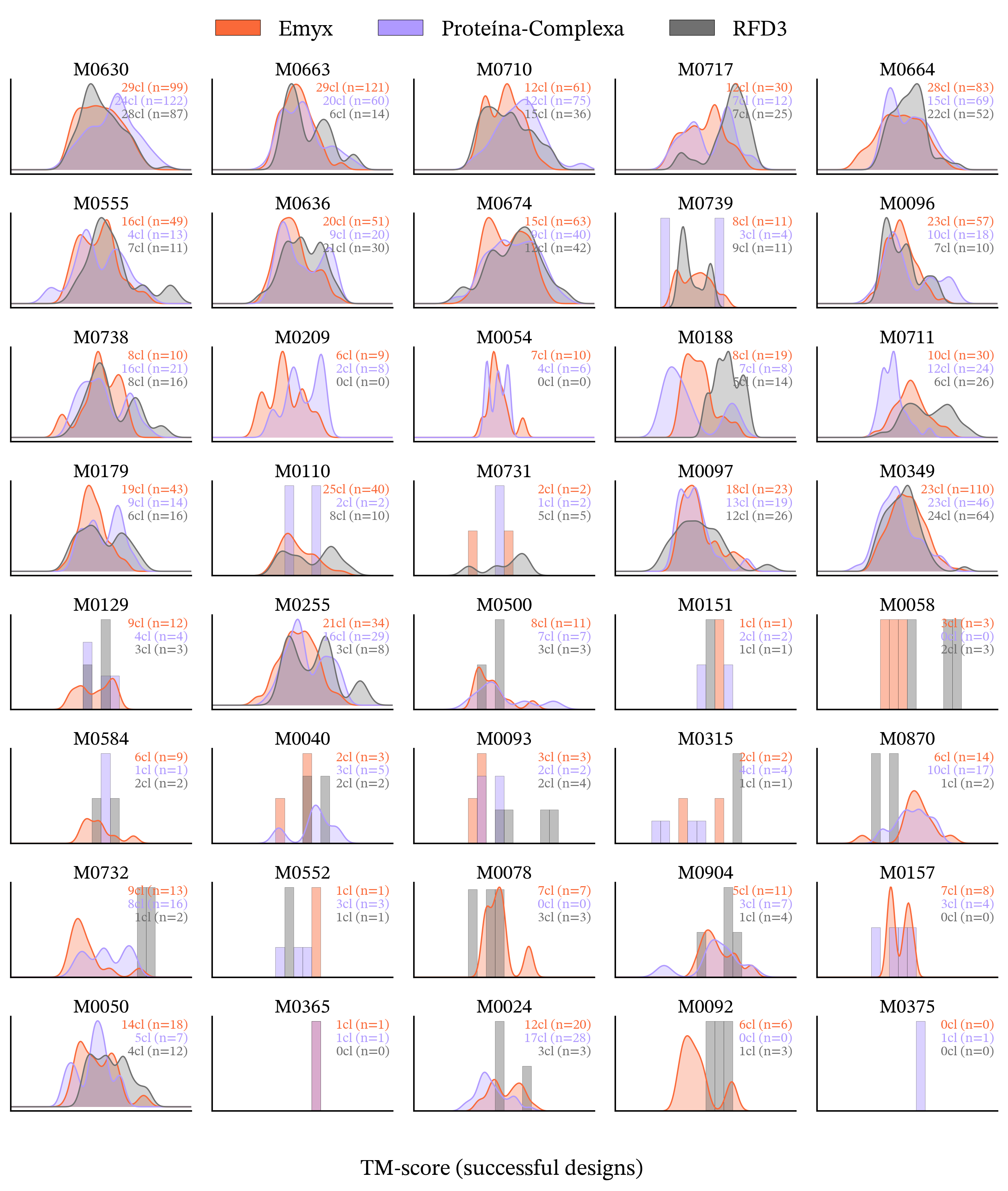}
    \caption{%
        \textbf{Per-target TM-score and cluster distributions} for \emyx, \rfdthree, and \proteinacomplexa (strict sc-RMSD; \S\ref{sec:backbone_eval}). TM-score to the closest PDB hit (Foldseek) for successful designs on all 41 AME targets (x-axis: TM-score $\in [0, 1]$). Each panel shows overlaid KDE densities; colour-coded annotations report the number of unique Foldseek clusters and successful designs $n$ per method.
    }
    \label{fig:ame_per_protein}
\end{figure}

Beyond aggregate metrics, per-target novelty and diversity reveal how consistently each model produces structurally distinct scaffolds. Figure~\ref{fig:ame_per_protein} shows the TM-score distribution and unique Foldseek cluster count for every AME target for \emyx, \rfdthree, and \proteinacomplexa, restricted to successful designs under strict sc-RMSD (\S\ref{sec:backbone_eval}). \emyx consistently produces designs with lower TM-scores (greater novelty) and more clusters (greater diversity) across the majority of targets.

\subsection{Original and redesigned sequence}
\label{si:orig_vs_redes}

\begin{figure}[H]
    \centering
    \includegraphics[width=\linewidth]{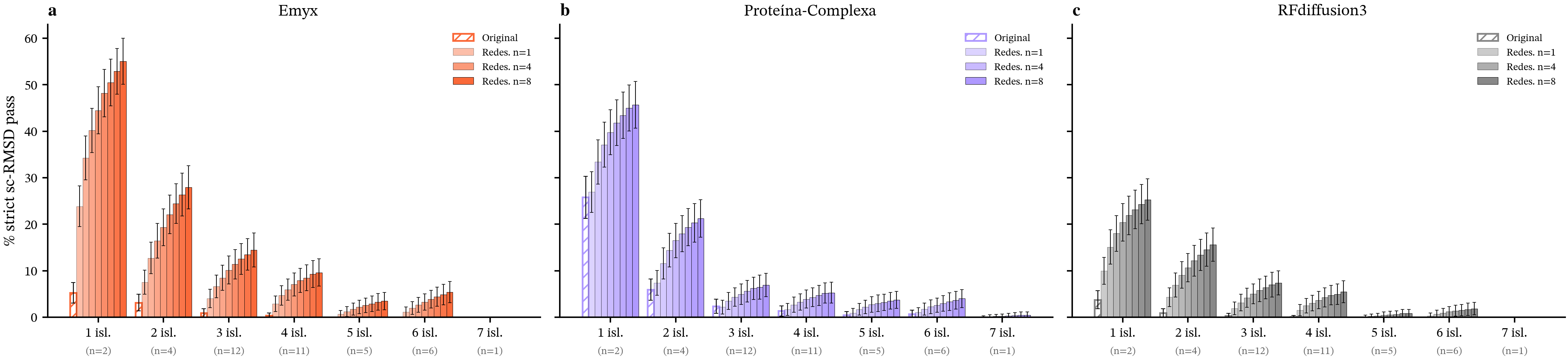}
    \caption{%
        \textbf{Original vs.\ redesigned sequence strict sc-RMSD success rate} (\S\ref{sec:backbone_eval}) by motif island count for \textbf{a}~\emyx, \textbf{b}~\proteinacomplexa, and \textbf{c}~\rfdthree. The white hatched bar is the original (single) sequence; the model-coloured bars show LigandMPNN best-of-$n$ for $n=1,2,\dots,8$ with darker shading indicating larger $n$. Bootstrap mean $\pm 1\sigma$ ($1{,}000$ resamples of 100 designs).
    }
    \label{fig:orig_vs_redes}
\end{figure}

All three models benefit from LigandMPNN sequence redesign: strict sc-RMSD success rates are consistently higher with redesigned sequences than with the original model-generated sequences across every island-count group (Figure~\ref{fig:orig_vs_redes}). This effect is most pronounced for \emyx, where the gap between original and redesigned rates widens on targets with fewer motif islands. We sweep the LigandMPNN best-of-$n$ from $n=1$ to $n=8$ to show how each additional redesign contributes; for \proteinacomplexa the single-redesign rate sits close to the original rate, with most of the gain only appearing once several redesigns are available.

\section{Conditional representation analysis}
\label{si:conditioning}

\begin{figure}[H]
    \centering
    \includegraphics[width=0.9\linewidth]{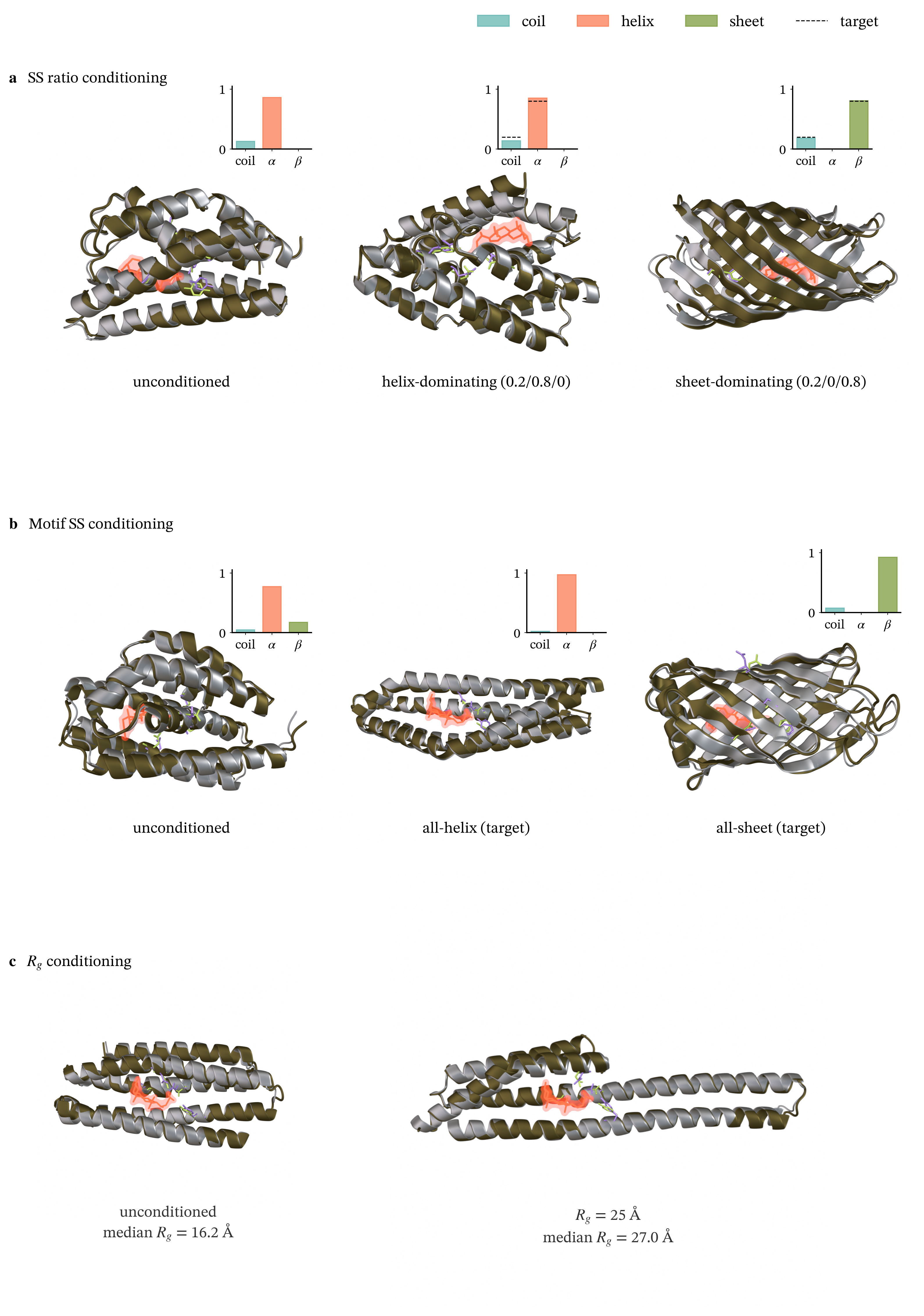}
    \caption{%
        \textbf{Conditional generation.} Samples conditioned on the M0349 motif, with global conditioning;
        \textbf{a}, Global secondary-structure (SS) ratio conditioning (unconditioned, helix-dominating, and sheet-dominating targets),
        \textbf{b}, per-motif-residue secondary-structure conditioning (unconditioned, all-helix, and all-sheet targets), and
        \textbf{c}, radius of gyration conditioning (unconditional and $R_g = 25$\,\angstrom{}). The inserted bar charts show the global distribution in samples secondary structure.
    }
    \label{fig:cond_all}
\end{figure}

\begin{figure}[H]
    \centering
    \includegraphics[width=0.95\linewidth]{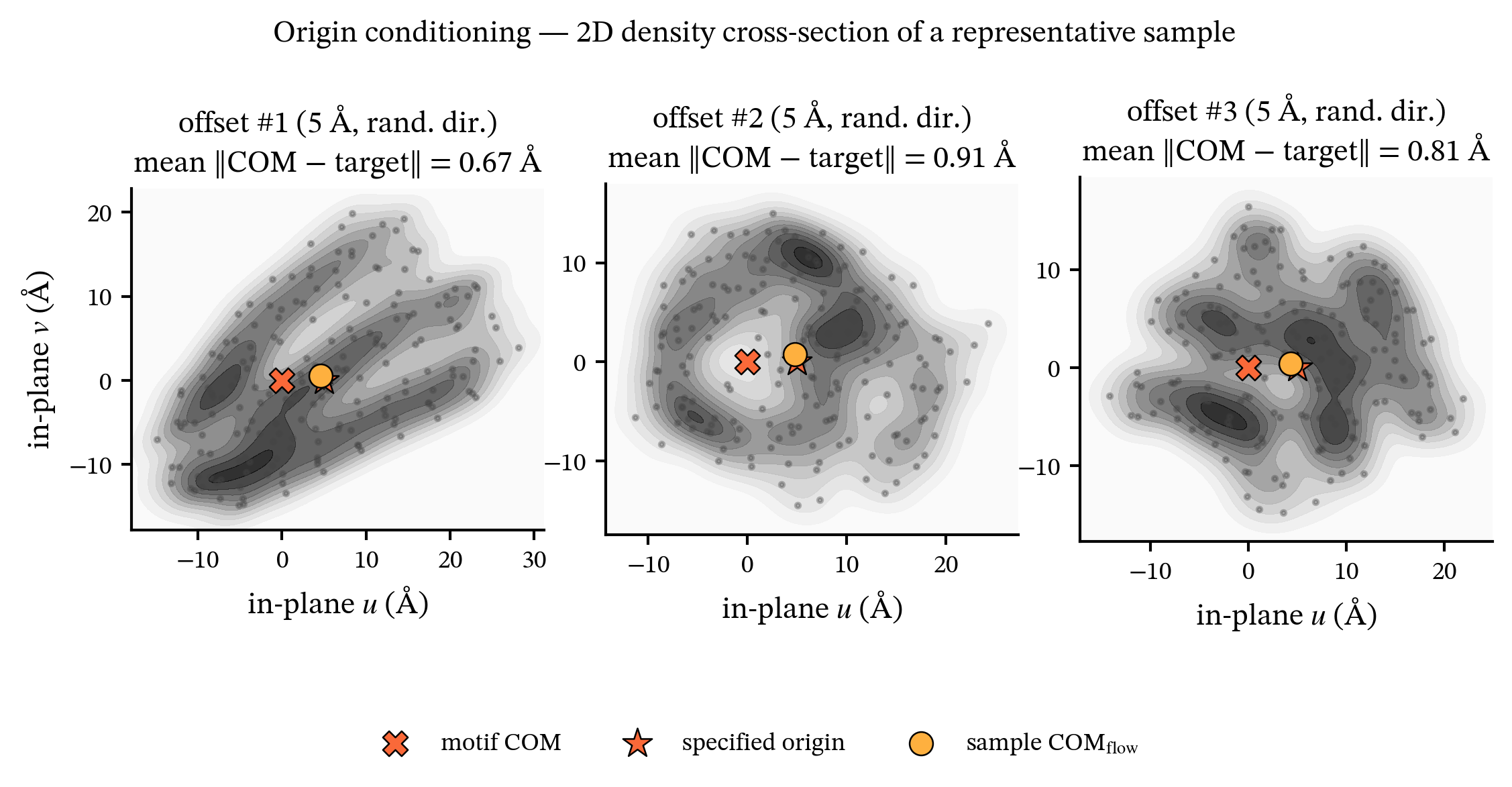}
    \caption{%
        \textbf{Noise-origin (centre-of-mass) placement conditioning on the M0349 motif.} For each subpanel the specified noise origin is offset from the motif COM by $5$\,\angstrom{} along a random direction. The contours show the in-plane Gaussian KDE of C$_\alpha$ atoms, with the motif COM (orange X), the specified origin (orange star), and the realised sample COM$_\text{flow}$ (yellow circle).
    }
    \label{fig:cond_origin}
\end{figure}

\emyx conditions generation on the global and per-residue structural features described in \S\ref{si:features_global}, each stochastically masked during training for regularisation (\S\ref{sec:training}). We evaluate each signal in isolation on 10 samples generated for the M0349 active site, varying global secondary-structure ratio, per-motif-residue secondary structure, radius of gyration $R_g$, and noise origin placement. These properties span the most commonly specified conditions in enzyme design: fold topology, scaffold compactness, and positioning relative to the motif. Figures \ref{fig:cond_all} and \ref{fig:cond_origin} illustrate that \emyx responds to all four conditioning signals on this representative target, producing structures whose measured properties shift in the expected direction. This is only intended for illustration and a systematic quantitative evaluation of conditioning fidelity across multiple targets is left for future work.

\section{Spectral analysis of weight utilisation}
\label{si:spectral_analysis}

We compare effective rank and power-law exponents across three generators (\emyx, \rfdthree, \proteinacomplexa~\citep{geffner2026proteinacomplexa}) and three structure predictors (Boltz-1~\citep{wohlwend2024boltz1}, Chai-1~\citep{chaidiscovery2024chai1}, Boltz-2~\citep{passaro2025boltz2}). The analysis uses the weightwatcher framework~\citep{martin2021heavytailed}. For each weight matrix $\bm{W} \in \R^{m \times n}$ with singular values $\sigma_1 \geq \sigma_2 \geq \cdots \geq \sigma_r$, the nuclear-to-spectral rank ratio is
\begin{equation}
    \mathcal{R}(\bm{W}) = \frac{\lVert \bm{W} \rVert_*}{\lVert \bm{W} \rVert_2} = \frac{\sum_i \sigma_i}{\sigma_1},
    \label{eq:stable_rank}
\end{equation}
and the effective rank percentage is
\begin{equation}
    \text{EffRank}(\bm{W}) = 100 \times \frac{\mathcal{R}(\bm{W})}{\min(m, n)}.
    \label{eq:eff_rank_pct}
\end{equation}
Intuitively, $\mathcal{R}(\bm{W})$ counts the effective number of significant singular values of $\bm{W}$: a full-rank matrix with uniformly distributed singular values has $\mathcal{R} = \min(m,n)$, while a rank-1 matrix has $\mathcal{R} = 1$ regardless of the matrix dimensions. Normalising by $\min(m,n)$ gives $\text{EffRank} \in (0, 100]$, which we interpret as the fraction of the available representational capacity that the layer actually uses. A layer at 15\% effective rank maps its inputs through a 15\%-dimensional subspace of the full weight space; the remaining 85\% of the singular spectrum contributes negligibly. For a transformer module, low effective rank across all layers indicates that the module is operating in a compressed regime, either because the input signal is inherently low-dimensional (as with noised coordinate embeddings in generators) or because training has collapsed the representation. High effective rank indicates that the module is utilising a richer subspace, typically because the input features (MSA covariation, template geometry) provide a diverse signal that the weights must preserve. Layers are grouped by transformer block index and averaged per component (QKV projections, output projections, and feed-forward networks).

\begin{figure}[H]
    \centering
    \includegraphics[width=\linewidth]{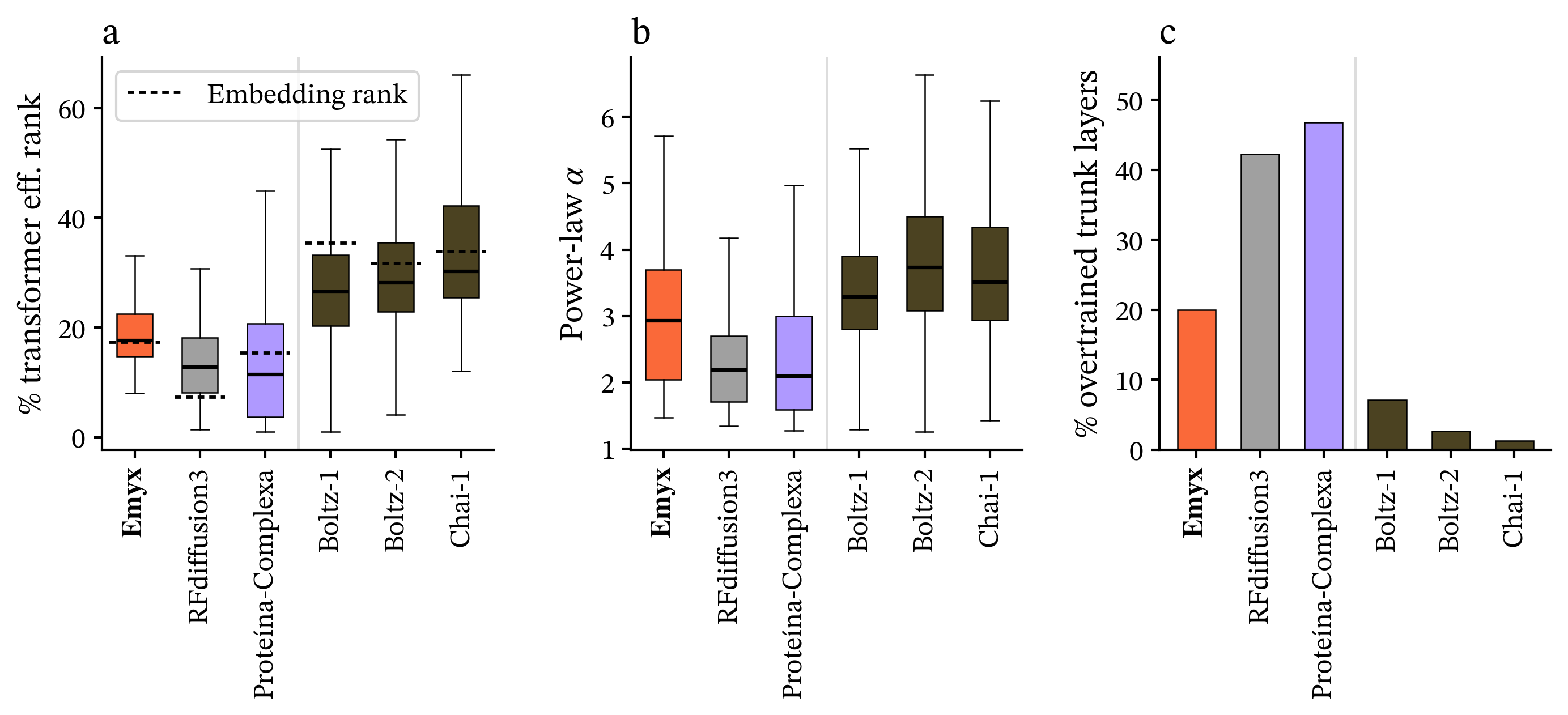}
    \caption{%
        \textbf{Spectral analysis of protein structure models (main transformer trunk).} \textbf{a}, Effective rank of the main transformer block, with dashed lines indicating the embedding rank from Table~\ref{tab:trunk_gap}. \textbf{b}, Power-law exponent $\alpha$. Generators (grey, \emyx highlighted in salmon) cluster at 15--19\% effective rank and lower $\alpha$; predictors (dark olive) occupy 29--35\% rank with higher $\alpha$. \textbf{c}, Percentage of trunk layers with $\alpha < 2.0$ (overtrained). \emyx has the lowest overtraining fraction among generators (20\%), while \proteinacomplexa and \rfdthree range from 42--47\%.
    }
    \label{fig:spectral_combined}
\end{figure}

\paragraph{Trunk effective rank and MSA enrichment.}
\label{si:spectral_embedding}
Table~\ref{tab:trunk_gap} and Figure~\ref{fig:spectral_combined}a show the trunk effective rank across all six models. Generators cluster at 15.0--18.5\% while predictors occupy 29.3--35.4\%. The three generator trunks span $140$--$170$M parameters and all cluster in this narrow low-rank regime, while predictor trunks span $316$--$521$M and all cluster in the high-rank regime. The gap likely originates in the initialisation signal, where generator coordinate embeddings achieve 7--17\% effective rank (Table~\ref{tab:trunk_gap}), while predictor sequence projections start comparably sparse (Boltz-1: 5.6\%, Boltz-2: 13.2\%). The divergence occurs because predictors have dedicated enrichment stages, MSA modules that extract co-evolutionary signal and template modules that encode distances and orientations from known homologous structures, that transform these sparse initialisations into rich representations before the main transformer module: Boltz-1's MSA module (35.4\% effective rank), Boltz-2's MSA module (31.7\%), and Chai-1's MSA module (33.9\%).

\begin{table}[H]
    \caption{\textbf{Effective rank (\%) of embedding and trunk blocks.} For generators, the embedding rank is the coordinate embedding rank; for predictors, it is the post-MSA embedding rank. \rfdthree does not embed coordinates directly. Predictors transform sparse sequence projections into dense representations via MSA covariation features and template-derived pair representations. \proteinacomplexa parameter count is the latent denoiser only (the analysed module); excludes the 128M all-atom decoder (inference-active) and 128M autoencoder encoder (training-only).}
    \label{tab:trunk_gap}
    \centering
    \footnotesize
    \setlength{\tabcolsep}{4pt}
    \begin{tabular}{llcccc}
        \toprule
        Model & Type & Params (M) & Trunk blocks & Embedding rank \% & Mean trunk rank \% \\
        \midrule
        \emyx         & Generator & 140 & 18 & 17.3 & 18.5 \\
        \rfdthree     & Generator & 168 & 18 & 7.3 & 15.0 \\
        Prote\'{i}na-Complexa      & Generator & 170 & 14 & 15.4 & 15.3 \\
        \midrule
        Boltz-1       & Predictor & 446 & 48 & 35.4 & 29.3 \\
        Boltz-2       & Predictor & 521 & 64 & 31.7 & 31.9 \\
        Chai-1        & Predictor & 316 & 48 & 33.9 & 35.4 \\
        \bottomrule
    \end{tabular}
\end{table}

\paragraph{Generators are prone to overtraining.} Generators lack the rich post-MSA and template embeddings that naturally regularise predictor trunks, making their weights more susceptible to overtraining. We quantify this using the power-law exponent $\alpha$ fitted to the empirical spectral density (ESD) of each weight matrix~\citep{martin2021heavytailed}. Given the ordered singular values $\sigma_1 \geq \sigma_2 \geq \cdots \geq \sigma_r$ of $\bm{W}$, the ESD of eigenvalues $\lambda_i = \sigma_i^2$ follows
\begin{equation}
    p(\lambda) \propto \lambda^{-\alpha},
    \label{eq:power_law}
\end{equation}
where $\alpha$ is estimated via maximum likelihood on the tail. Lower $\alpha$ indicates tighter fitting to the data, with $\alpha < 2$ traditionally considered overtrained~\citep{martin2021heavytailed}. We note that this threshold is a convention from the weightwatcher literature rather than a sharp phase boundary; the relative ordering of models is robust to the choice of threshold (at $\alpha < 1.5$, the generator-predictor gap persists, and at $\alpha < 2.5$ the ranking is unchanged, though absolute percentages shift).
\begin{table}[H]
    \caption{\textbf{Power-law exponent statistics for main transformer module.} Generators have lower mean $\alpha$ and a higher proportion of overtrained layers than predictors. Statistics are computed over weight matrices in the main transformer trunk, consistent with the effective rank analysis in Table~\ref{tab:trunk_gap}. \emph{Well-trained} ($2 \leq \alpha \leq 6$) and \emph{overtrained} ($\alpha < 2$) percentages need not sum to $100\%$: the remainder is the under-trained regime ($\alpha > 6$, fits not converged).}
    \label{tab:alpha_stats}
    \centering
    \small
    \begin{tabular}{llccc}
        \toprule
        Model & Type & Mean $\alpha$ & Well-trained \% & Overtrained \% \\
        \midrule
        \emyx         & Generator & 2.97 & 80.0 & 20.0 \\
        \rfdthree     & Generator & 2.45 & 55.9 & 42.2 \\
        Prote\'{i}na-Complexa      & Generator & 2.36 & 53.2 & 46.8 \\
        \midrule
        Boltz-1       & Predictor & 3.48 & 89.6 & 7.1 \\
        Chai-1        & Predictor & 3.01 & 82.0 & 16.4 \\
        Boltz-2       & Predictor & 4.07 & 87.7 & 2.7 \\
        \bottomrule
    \end{tabular}
\end{table}

Generators have systematically lower $\alpha$ (mean 2.59 across three models) than predictors (mean 3.52), with 20--47\% of trunk layers falling below the $\alpha = 2$ overtrained boundary versus 3--16\% for predictors (Table~\ref{tab:alpha_stats}, Figure~\ref{fig:spectral_combined}b,c). \proteinacomplexa is the most overtrained generator (47\% of trunk layers below $\alpha = 2.0$), followed by \rfdthree (42\%). \emyx achieves the lowest overtrained fraction ($20.0\%$, mean $\alpha = 2.97$), consistent with its use of stochastic depth (drop-path) and low-dimensional bottleneck modulation layers (\S\ref{si:training_hyperparameters}), though this analysis cannot establish causation without controlled ablation. Notably, \proteinacomplexa has the highest overtrained fraction yet produces reasonably diverse designs (median TM \edit{$0.50$}, \edit{$292$} clusters), suggesting that compensating factors, plausibly its ${\approx}2\times$ larger training data and ${\sim}\,3\times$ total training parameter count (426M; cf.\ \emyx 140M), can offset poor weight regularisation. The spectral analysis therefore characterises regularisation quality across the model family rather than directly predicting generation diversity.

\section{Sample gallery}
\label{si:sample_gallery}

\begin{figure}[H]
    \centering
    \includegraphics[width=0.85\linewidth]{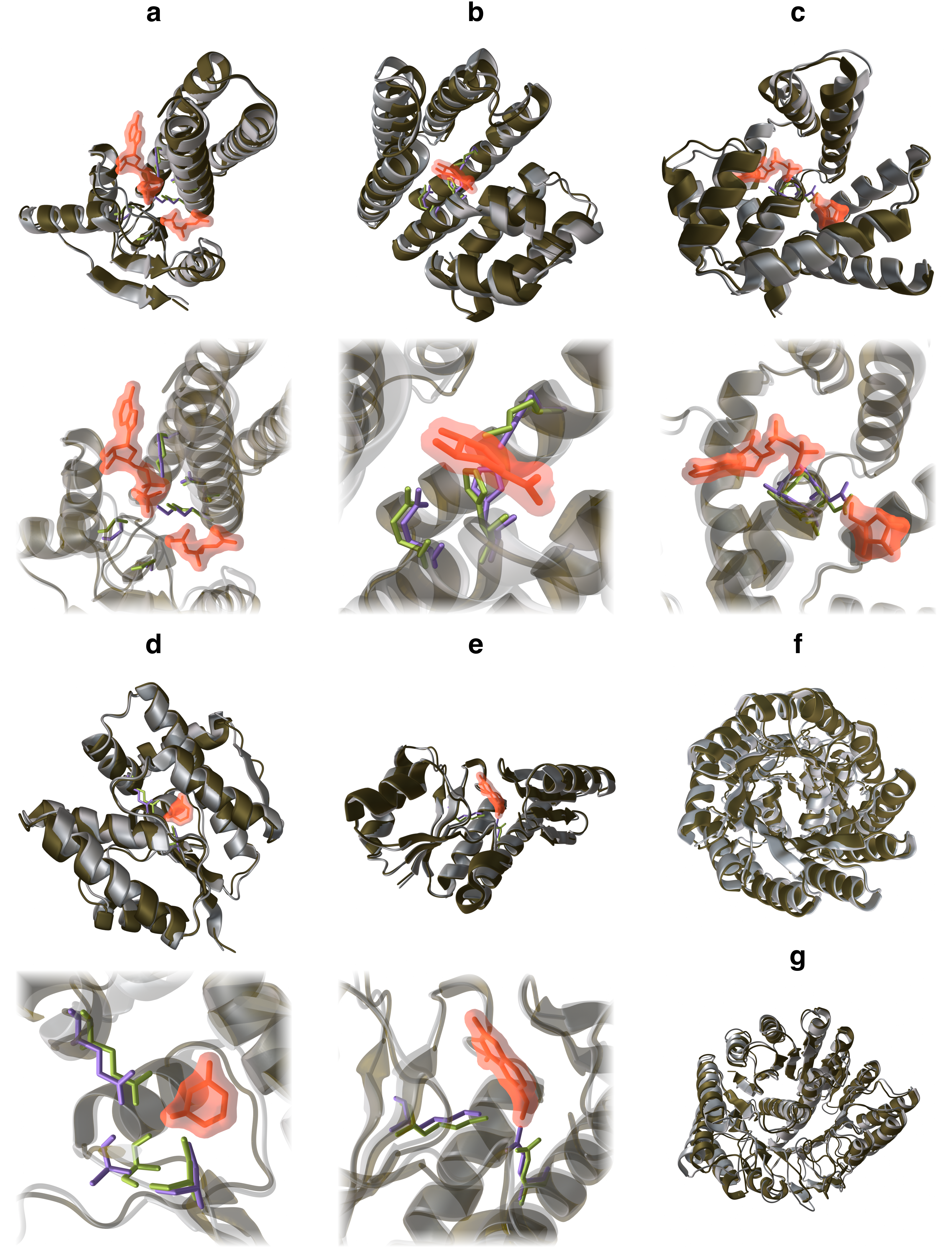}
    \caption{%
        \textbf{Gallery of \emyx-generated structures.} \textbf{a--e}, AME benchmark scaffolds (180 residues), each with full scaffold (top) and active-site detail (bottom).
        \textbf{a}, M0870.
        \textbf{b}, M0054.
        \textbf{c}, M0663.
        \textbf{d}, M0636.
        \textbf{e}, M0664.
        \textbf{f--g}, Large unconditional samples (500 residues each).
    }
    \label{fig:sample_gallery}
\end{figure}

Figure~\ref{fig:sample_gallery} presents representative \emyx-generated structures spanning both motif-conditioned and unconditional generation. Panels \textbf{a}--\textbf{e} show five AME benchmark scaffolds (180 residues each), displayed at two scales: the full scaffold (top) and a close-up of the active-site region (bottom). Motif residues and ligands are highlighted. Panels \textbf{f}--\textbf{g} show two large unconditional samples (500 residues each), demonstrating that the model generalises to extended structures well beyond the typical AME benchmark chain length.

\section{Architecture}
\label{si:architecture}

This section gives the full component definitions and pseudocode for the \emyx architecture summarised in \S\ref{sec:arch}. Each component is described in prose immediately followed by the corresponding pseudocode. All operations use RMSNorm (eps=$10^{-6}$) and Linear layers without bias unless otherwise noted.

\begin{figure}[H]
    \centering
    \includegraphics[width=\linewidth]{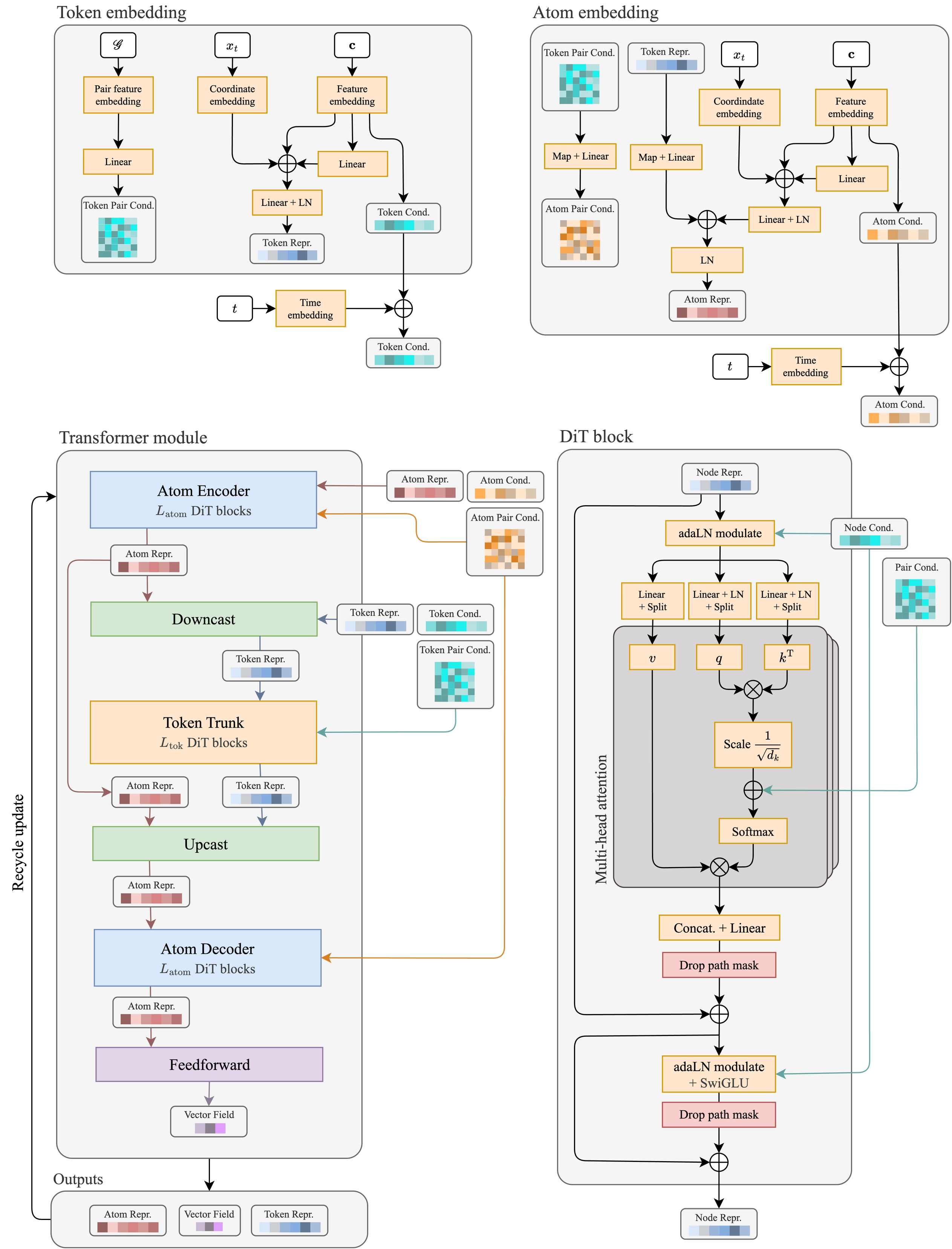}
    \caption{%
        \textbf{Detailed \emyx architecture.} Expanded version of Figure~\ref{fig:samples_efficiency}\textbf{a} showing the full component graph, including embedding inputs, pair conditioning paths, sparse edge construction, atom/token-level transformer stacks, gated cross-attention bridges (downcast/upcast), and the recycling loop. Tensor shapes and per-block modulation signals are annotated where space permits. Component definitions and pseudocode follow.
    }
    \label{fig:architecture_detailed}
\end{figure}

\subsection{Notation}
\label{si:arch_notation}

{%
\footnotesize
\hfuzz=4pt
\renewcommand{\iconnode}[1]{\includegraphics[height=1.8em]{\repdir/#1}}%
\renewcommand{\iconpair}[1]{\includegraphics[height=4em]{\repdir/#1}}%
\renewcommand{\graphrep}{\raisebox{-1.2ex}{\includegraphics[height=3.6em]{\repdir/representations_graph}}}%
\begin{longtable}{@{}m{0.32\linewidth}m{0.38\linewidth}m{0.10\linewidth}@{}}
    \caption{\textbf{Notation used in the pseudocode algorithms.}}
    \label{tab:notation} \\
    \toprule
    \multicolumn{3}{c}{\textit{General terminology}} \\
    \midrule
    \endfirsthead
    \caption[]{Notation (continued).} \\
    \toprule
    \endhead
    \midrule
    \multicolumn{3}{r}{\footnotesize\textit{Continued on next page}} \\
    \bottomrule
    \endfoot
    \bottomrule
    \endlastfoot
    Token & \multicolumn{2}{l}{Residue or ligand unit, each containing up to $A{=}14$ atoms} \\
    Rep14 & \multicolumn{2}{l}{Fixed 14-atom representation per token; ghost atoms use N/O coords} \\
    \midrule
    \multicolumn{3}{c}{\textit{Dimensions}} \\
    \midrule
    $N_\text{atom}$ & \multicolumn{2}{l}{Total number of atoms in the batch} \\
    $N_\text{tok}$ & \multicolumn{2}{l}{Total number of tokens in the batch} \\
    $A = 14$ & \multicolumn{2}{l}{Maximum atoms per token} \\
    $d_\text{atom}$ & \multicolumn{2}{l}{Atom representation dimension} \\
    $d_\text{tok}$ & \multicolumn{2}{l}{Token representation dimension} \\
    $E_\text{atom}$, $E_\text{tok}$ & \multicolumn{2}{l}{Number of sparse edges (atom, token level)} \\
    $H_\text{atom}$, $H_\text{tok}$ & \multicolumn{2}{l}{Number of attention heads (atom, token level)} \\
    $H_\text{cross}$ & \multicolumn{2}{l}{Number of cross-attention heads} \\
    $d_h$ & \multicolumn{2}{l}{Per-head dimension ($= d / H$)} \\
    $L_\text{atom}$, $L_\text{tok}$ & \multicolumn{2}{l}{Number of transformer layers (atom, token level)} \\
    $d_\text{time}$ & \multicolumn{2}{l}{Time embedding dimension} \\
    $d_\text{edge}$ & \multicolumn{2}{l}{Edge embedding dimension} \\
    $n_\text{freq}$ & \multicolumn{2}{l}{Number of Fourier frequencies} \\
    $n_\text{dup}$ & \multicolumn{2}{l}{Number of duplicate copies in upcast} \\
    $n_\text{seq}$ & \multicolumn{2}{l}{Sequence neighbour radius} \\
    $n_\text{budget}$ & \multicolumn{2}{l}{Edge budget per node} \\
    $N_r$ & \multicolumn{2}{l}{Number of recycling iterations} \\
    $p_\text{drop}$ & \multicolumn{2}{l}{Dropout probability (applied to attention output and SwiGLU hidden)} \\
    $p_\text{path}$ & \multicolumn{2}{l}{Maximum stochastic depth (DropPath) probability; linearly scaled $0 \to p_\text{path}$} \\
    $d_\text{bn}$ & \multicolumn{2}{l}{Bottleneck dimension in adaLN modulation} \\
    $\mathcal{N}(i)$ & \multicolumn{2}{l}{Sparse neighbourhood of node $i$ in graph $G$} \\
    \midrule
    \multicolumn{3}{c}{\textit{Inputs}} \\
    \midrule
    $\bm{x}_t \in \R^{N_\text{atom} \times 3}$ & \multicolumn{2}{l}{Noisy atom coordinates at time $t$} \\
    $\bm{c}$ & \multicolumn{2}{l}{Conditioning (categorical features, masks, motif flags, residue indices)} \\
    $G$ & Sparse connectivity graph (built by Alg.~\ref{alg:edges}) & \graphrep \\
    $G^\text{atom}$, $G^\text{tok}$ & \multicolumn{2}{l}{Atom-level and token-level subgraphs of $G$} \\
    $\bm{M} \in \mathbb{N}^{3 \times N_\text{atom}}$ & \multicolumn{2}{l}{Rep14 mapping (atom index, token index, position within token)} \\
    $\bm{m}_\text{valid} \in \{0,1\}^{N_\text{tok} \times 14}$ & \multicolumn{2}{l}{Rep14 validity mask (masks empty slots)} \\
    $t \in [0, 1]$ & \multicolumn{2}{l}{Flow matching time} \\
    \midrule
    \multicolumn{3}{c}{\textit{Outputs}} \\
    \midrule
    $\bm{v}_\theta$ & \multicolumn{2}{l}{Learned velocity model (parameterised by $\theta$)} \\
    $\bm{v} \in \R^{N_\text{atom} \times 3}$ & \multicolumn{2}{l}{Predicted velocity field (output of $\bm{v}_\theta$)} \\
    $\hat{\bm{x}} \in \R^{N_\text{atom} \times 3}$ & \multicolumn{2}{l}{Predicted structure ($= \bm{x}_t + (1-t)\,\bm{v}$); recycled via detach} \\
    \midrule
    \multicolumn{3}{c}{\textit{Learned representations}} \\
    \midrule
    $\bm{h}^\text{atom} \in \R^{N_\text{atom} \times d_\text{atom}}$ & Atom-level node representation & \atomrep \\
    $\bm{h}^\text{tok} \in \R^{N_\text{tok} \times d_\text{tok}}$ & Token-level node representation & \tokenrep \\
    $\bm{c}^\text{atom} \in \R^{N_\text{atom} \times d_\text{atom}}$ & Atom-level conditioning & \atomcond \\
    $\bm{c}^\text{tok} \in \R^{N_\text{tok} \times d_\text{tok}}$ & Token-level conditioning & \tokencond \\
    $\bm{b}^\text{atom} \in \R^{E_\text{atom} \times H_\text{atom}}$ & Atom-level pair bias & \atompair \\
    $\bm{b}^\text{tok} \in \R^{E_\text{tok} \times H_\text{tok}}$ & Token-level pair bias & \tokenpair \\
    $\bm{t}^\text{atom} \in \R^{N_\text{atom} \times d_\text{atom}}$ & \multicolumn{2}{l}{Atom-level time embedding} \\
    $\bm{t}^\text{tok} \in \R^{N_\text{tok} \times d_\text{tok}}$ & \multicolumn{2}{l}{Token-level time embedding} \\
    \midrule
    \multicolumn{3}{c}{\textit{Recycling (Alg.~\ref{alg:forward_pass})}} \\
    \midrule
    $\bm{h}^\text{tok}_\text{rec}$, $\bm{h}^\text{atom}_\text{rec}$ & \multicolumn{2}{l}{Detached node representations from previous recycling iteration} \\
    \midrule
    \multicolumn{3}{c}{\textit{Transformer internals (Algs.~\ref{alg:dit_block}--\ref{alg:swiglu})}} \\
    \midrule
    $\bm{Q}, \bm{K}, \bm{V}$ & \multicolumn{2}{l}{Query, key, value matrices in attention} \\
    $(\gamma, \beta, \alpha)$ & \multicolumn{2}{l}{adaLN scale, shift, and gate; subscript 1/2 for attention/MLP sub-layers} \\
    $d_\text{hidden}$ & \multicolumn{2}{l}{SwiGLU hidden dimension ($= \lceil \tfrac{8}{3} d \,/\, 256 \rceil \cdot 256$)} \\
    \midrule
    \multicolumn{3}{c}{\textit{EDM sampling (Alg.~\ref{alg:edm})}} \\
    \midrule
    $\bm{\epsilon}$ & \multicolumn{2}{l}{Gaussian noise sample ($\sim \mathcal{N}(\bm{0}, \sigma_\text{data}^2\,\mathbf{I})$)} \\
    $\sigma$ & \multicolumn{2}{l}{EDM noise level} \\
    $\sigma_\text{data}$ & \multicolumn{2}{l}{Data noise standard deviation (from coordinate scale)} \\
    $\sigma_\text{max}$, $\sigma_\text{min}$ & \multicolumn{2}{l}{Karras schedule endpoints} \\
    $\rho$ & \multicolumn{2}{l}{Karras schedule exponent (controls step distribution)} \\
    $\{\sigma_i\}_{i=0}^{n-1}$ & \multicolumn{2}{l}{Karras noise schedule (Eq.~\ref{eq:karras})} \\
    $n$ & \multicolumn{2}{l}{Number of sampling steps} \\
    $\bm{y}$ & \multicolumn{2}{l}{EDM noisy state ($= \xt / t$)} \\
    $D$ & \multicolumn{2}{l}{Denoised estimate ($= \xt + (1-t)\,\bm{v}_\theta$)} \\
    $\gamma$ & \multicolumn{2}{l}{Stochastic churn amplitude} \\
    $s_\text{noise}$ & \multicolumn{2}{l}{Noise amplification factor} \\
    $s_\text{step}$ & \multicolumn{2}{l}{Step scale factor} \\
    $\sigma_\text{min churn}$ & \multicolumn{2}{l}{Churn threshold (disabled for final steps)} \\
\end{longtable}
}

\subsection{Forward pass}
\label{si:alg_forward}

The top-level forward pass (Algorithm~\ref{alg:forward_pass}) computes
embeddings once and then runs the trunk through $N_r{+}1$ recycling iterations.
On each iteration we inject the previous pass's token representations, atom representations,
and a distogram built from the current coordinate estimate. Only the final iteration carries gradients.

\begin{algorithm}[H]
\caption{\emyx forward pass with recycling.\hfill$\bm{v} \leftarrow (\bm{x}_t,\, \bm{c},\, t,\, N_r)$}
\label{alg:forward_pass}
\begin{algorithmic}[1]
\STATE \textbf{def} ForwardPass($\bm{x}_t \in \R^{N_\text{atom} \times 3}$,
    $\bm{c}$, $t \in [0,1]$, $N_r \in \mathbb{N}$):
\begin{algindent}
    \vspace{6pt}
    \STATE $(G, \bm{h}^\text{atom}_0, \bm{b}^\text{atom}, \bm{c}^\text{atom},
           \bm{h}^\text{tok}_0, \bm{b}^\text{tok}_0, \bm{c}^\text{tok})
           \leftarrow$ Embed$(\bm{x}_t, \bm{c}, t)$
        \hfill $\triangleright$ Alg.~\ref{alg:init_embed}
    \vspace{6pt}
    \STATE $\bm{h}^\text{tok}_\text{rec} \leftarrow \bm{0}$;
           \quad $\bm{h}^\text{atom}_\text{rec} \leftarrow \bm{0}$
        \hfill $\triangleright$ Recycling state
    \vspace{6pt}
    \FOR{$r = 0, \ldots, N_r$}
        \STATE \textbf{if} $r < N_r$: disable gradients
        \vspace{6pt}
        \STATE $\bm{h}^\text{tok} \leftarrow \bm{h}^\text{tok}_0
                + \text{Linear}(\text{RMSNorm}(\bm{h}^\text{tok}_\text{rec}))$
            \hfill $\triangleright$ Token recycle: ADD to initial embedding
        \vspace{6pt}
        \STATE $\bm{h}^\text{atom} \leftarrow \bm{h}^\text{atom}_0
                + \text{Linear}(\text{RMSNorm}(\bm{h}^\text{atom}_\text{rec}))$
            \hfill $\triangleright$ Atom recycle: ADD to initial embedding
        \vspace{6pt}
        \STATE $\bm{b}^\text{tok} \leftarrow \bm{b}^\text{tok}_0
                + [r > 0] \cdot \text{DistogramEmbed}\hat{\bm{x}}, \bm{M}, G^\text{tok})$
            \hfill $\triangleright$ Pair bias recycle: ADD when $r > 0$;
        \vspace{6pt}
        \STATE $(\bm{h}^\text{atom}, \bm{h}^\text{tok})
                \leftarrow$ TransformerModule$(\bm{h}^\text{atom}, \bm{b}^\text{atom}, \bm{c}^\text{atom},
                \bm{h}^\text{tok}, \bm{b}^\text{tok}, \bm{c}^\text{tok}, G)$
            \hfill $\triangleright$ Alg.~\ref{alg:trunk}
        \vspace{6pt}
        \STATE $\bm{v} \leftarrow$ OutputHead$(\bm{h}^\text{atom}, \bm{c}^\text{atom})$
            \hfill $\triangleright$ Alg.~\ref{alg:output}
        \vspace{6pt}
        \STATE $\bm{v}[\text{is\_motif}] \leftarrow \bm{0}$
        \vspace{6pt}
        \STATE $\hat{\bm{x}} \leftarrow \bm{x}_t + (1 - t) \, \bm{v}$
            \hfill $\triangleright$ Predicted structure
        \vspace{6pt}
        \STATE $\bm{h}^\text{tok}_\text{rec} \leftarrow \text{detach}(\bm{h}^\text{tok})$;
               \quad $\bm{h}^\text{atom}_\text{rec} \leftarrow \text{detach}(\bm{h}^\text{atom})$
    \ENDFOR
    \vspace{6pt}
    \RETURN $\bm{v} \in \R^{N_\text{atom} \times 3}$
        \hfill $\triangleright$ Predicted velocity
\end{algindent}
\end{algorithmic}
\end{algorithm}

\subsection{Embedding initialisation}
\label{si:alg_embed}

Algorithm~\ref{alg:init_embed} constructs the sparse connectivity graph, token-, atom-, and time-level embeddings, and sums the time contribution into the per-level conditioning signal. The individual embedding subroutines are described in the subsequent subsections.

\begin{algorithm}[H]
\caption{Embedding initialisation.\newline\mbox{}\hfill\graphrep\;\atomrep\;\atompair\;\atomcond\;\tokenrep\;\tokenpair\;\tokencond\;$\leftarrow (\bm{x}_t,\, \bm{c},\, t)$}
\label{alg:init_embed}
\begin{algorithmic}[1]
\STATE \textbf{def} Embed($\bm{x}_t \in \R^{N_\text{atom} \times 3}$,
    $\bm{c}$, $t \in [0,1]$):
\begin{algindent}
    \vspace{6pt}
    \STATE $G \leftarrow$ EdgeModule$(\bm{x}_t, \bm{c})$
        \hfill $\triangleright$ Alg.~\ref{alg:edges}; atom and token edge sets
    \vspace{6pt}
    \STATE $(\bm{h}^\text{tok}, \bm{b}^\text{tok}, \bm{c}^\text{tok})
           \leftarrow$ TokenEmbed$(\bm{x}_t, \bm{c}, G^\text{tok})$
        \hfill $\triangleright$ Alg.~\ref{alg:token_embed}
    \vspace{6pt}
    \STATE $(\bm{h}^\text{atom}, \bm{b}^\text{atom}, \bm{c}^\text{atom})
           \leftarrow$ AtomEmbed$(\bm{x}_t, \bm{c}, \bm{h}^\text{tok}, \bm{b}^\text{tok}, G)$
        \hfill $\triangleright$ Alg.~\ref{alg:atom_embed}
    \vspace{6pt}
    \STATE $(\bm{t}^\text{atom}, \bm{t}^\text{tok})
           \leftarrow$ TimeEmbed$(t, \bm{c})$
        \hfill $\triangleright$ Alg.~\ref{alg:time_embed}
    \vspace{6pt}
    \STATE $\bm{c}^\text{atom} \leftarrow (\bm{c}^\text{atom} + \bm{t}^\text{atom}) \,/\, 2$
    \vspace{6pt}
    \STATE $\bm{c}^\text{tok} \leftarrow (\bm{c}^\text{tok} + \bm{t}^\text{tok}) \,/\, 2$
    \vspace{6pt}
    \RETURN
    \STATE \quad $G \in \mathbb{N}^{E \times 2}$,
    \STATE \quad $\bm{h}^\text{atom} \in \R^{N_\text{atom} \times d_\text{atom}}$,
    \STATE \quad $\bm{b}^\text{atom} \in \R^{E_\text{atom} \times H_\text{atom}}$,
    \STATE \quad $\bm{c}^\text{atom} \in \R^{N_\text{atom} \times d_\text{atom}}$,
    \STATE \quad $\bm{h}^\text{tok} \in \R^{N_\text{tok} \times d_\text{tok}}$,
    \STATE \quad $\bm{b}^\text{tok} \in \R^{E_\text{tok} \times H_\text{tok}}$,
    \STATE \quad $\bm{c}^\text{tok} \in \R^{N_\text{tok} \times d_\text{tok}}$
\end{algindent}
\end{algorithmic}
\end{algorithm}

\subsection{Transformer trunk}
\label{si:alg_trunk}

The core of \emyx is a five-stage transformer module (Figure~\ref{fig:samples_efficiency}\textbf{a}).
The TransformerModule (Algorithm~\ref{alg:trunk}) implements this patchify--process--unpatchify pipeline, illustrated below.

\begin{figure}[H]
\centering
\begin{tikzpicture}[
    stage/.style={draw, rounded corners=3pt, minimum width=5.2cm,
                  minimum height=0.8cm, align=center, font=\small},
    arr/.style={-{Stealth[length=5pt]}, thick, gray!70},
    desc/.style={font=\scriptsize, anchor=west, text width=6.2cm, align=left},
]
\node[stage] (enc)   {\textbf{Atom encoder}};
\node[stage, below=0.45cm of enc]   (down)  {\textbf{Downcast}};
\node[stage, below=0.45cm of down]  (trunk) {\textbf{Token trunk}};
\node[stage, below=0.45cm of trunk] (up)    {\textbf{Upcast}};
\node[stage, below=0.45cm of up]    (dec)   {\textbf{Atom decoder}};

\draw[arr] (enc)   -- (down);
\draw[arr] (down)  -- (trunk);
\draw[arr] (trunk) -- (up);
\draw[arr] (up)    -- (dec);

\node[desc] at ([xshift=0.4cm]enc.east)   {$L_\text{atom}$ DiT blocks at atom resolution, capturing local atomic detail};
\node[desc] at ([xshift=0.4cm]down.east)  {Gated cross-attention pooling atom information into tokens via Rep14};
\node[desc] at ([xshift=0.4cm]trunk.east) {$L_\text{tok}$ DiT blocks at token resolution; main processing stage with the vast majority of parameters};
\node[desc] at ([xshift=0.4cm]up.east)    {Gated cross-attention broadcasting token information back to individual atoms};
\node[desc] at ([xshift=0.4cm]dec.east)   {$L_\text{atom}$ DiT blocks recovering localised atomic detail};
\end{tikzpicture}
\end{figure}

\begin{algorithm}[H]
\caption{TransformerModule.\newline\mbox{}\hfill\atomrep\;\tokenrep\;$\leftarrow$\;\atomrep\;\atompair\;\atomcond\;\tokenrep\;\tokenpair\;\tokencond\;\graphrep}
\label{alg:trunk}
\begin{algorithmic}[1]
\STATE \textbf{def} TransformerModule(
    $\bm{h}^\text{atom} \in \R^{N_\text{atom} \times d_\text{atom}}$,
    $\bm{b}^\text{atom} \in \R^{E_\text{atom} \times H_\text{atom}}$,
    $\bm{c}^\text{atom} \in \R^{N_\text{atom} \times d_\text{atom}}$,
    $\bm{h}^\text{tok} \in \R^{N_\text{tok} \times d_\text{tok}}$,
    $\bm{b}^\text{tok} \in \R^{E_\text{tok} \times H_\text{tok}}$,
    $\bm{c}^\text{tok} \in \R^{N_\text{tok} \times d_\text{tok}}$,
    $G \in \mathbb{N}^{E \times 2}$):
\begin{algindent}
    \vspace{6pt}
    \STATE $\bm{h}^\text{atom} \leftarrow$ DiTBlock$(\bm{h}^\text{atom},\;
        \bm{c}^\text{atom},\; \bm{b}^\text{atom},\;
        G^\text{atom},\; L_\text{atom})$
        \hfill $\triangleright$ Alg.~\ref{alg:dit_block}; atom encoder
    \vspace{6pt}
    \STATE $\bm{h}^\text{tok} \leftarrow$ Downcast$(\bm{h}^\text{atom},\;
        \bm{h}^\text{tok},\; \bm{M})$
        \hfill $\triangleright$ Alg.~\ref{alg:downcast}
    \vspace{6pt}
    \STATE $\bm{h}^\text{tok} \leftarrow$ DiTBlock$(\bm{h}^\text{tok},\;
        \bm{c}^\text{tok},\; \bm{b}^\text{tok},\;
        G^\text{tok},\; L_\text{tok})$
        \hfill $\triangleright$ Token trunk
    \vspace{6pt}
    \STATE $\bm{h}^\text{atom} \leftarrow$ Upcast$(\bm{h}^\text{atom},\;
        \bm{h}^\text{tok},\; \bm{M})$
        \hfill $\triangleright$ Alg.~\ref{alg:upcast}
    \vspace{6pt}
    \STATE $\bm{h}^\text{atom} \leftarrow$ DiTBlock$(\bm{h}^\text{atom},\;
        \bm{c}^\text{atom},\; \bm{b}^\text{atom},\;
        G^\text{atom},\; L_\text{atom})$
        \hfill $\triangleright$ Atom decoder
    \vspace{6pt}
    \RETURN $\bm{h}^\text{atom} \in \R^{N_\text{atom} \times d_\text{atom}}$,\;
            $\bm{h}^\text{tok} \in \R^{N_\text{tok} \times d_\text{tok}}$
\end{algindent}
\end{algorithmic}
\end{algorithm}

\subsection{DiT block}
\label{si:alg_dit}

Each transformer layer follows the DiT~\citep{peebles2023dit} design with adaptive layer normalisation (adaLN-Zero) conditioned on the diffusion timestep, sparse self-attention with per-head pair bias, and a SwiGLU~\citep{shazeer2020glu} feed-forward network. The same block design is used at both atom and token resolutions (with different dimensions and head counts). The conditioning signal $\bm{c}$ drives a modulation network that produces shift, scale, and gate parameters for both the attention and feed-forward sub-layers; the gate parameters modulate the residual connections, controlling how much each sub-layer contributes to the residual stream.

We introduce four modifications to the standard DiT design. First, a \emph{bottleneck projection}: the conditioning is first projected to a small intermediate dimension before expansion to the full modulation vector, reducing the parameter count of the modulation pathway and improving weight utilisation as measured by spectral analysis (\S\ref{sec:res_weight_quality}). Second, we apply a \emph{sigmoid activation} to the gate parameters, bounding them to $[0, 1]$ and stabilising the residual stream. Third, we apply \emph{dropout} to both the attention output projection and the SwiGLU hidden activations. Fourth, we use \emph{stochastic depth} (DropPath)~\citep{huang-stochastic-dropout}, where entire attention or feed-forward sub-layers are bypassed during training with probability that increases linearly from $0$ to $p_\text{path}$ across the layer stack, providing regularisation that grows with depth.

\begin{algorithm}[H]
\caption{DiT-style transformer block with adaLN-Zero modulation, dropout, and stochastic depth.\newline\mbox{}\hfill\tokenrep\;$\leftarrow$\;\tokenrep\;\tokencond\;\tokenpair\;\graphrep}
\label{alg:dit_block}
\begin{algorithmic}[1]
\STATE \textbf{def} DiTBlock(
    $\bm{h} \in \R^{N \times d}$,
    $\bm{c} \in \R^{N \times d}$,
    $\bm{b} \in \R^{E \times H}$,
    $G \in \mathbb{N}^{E \times 2}$,
    $L \in \mathbb{N}$):
\begin{algindent}
    \FOR{$\ell = 1, \ldots, L$}
        \vspace{6pt}
        \STATE $p_\ell \leftarrow p_\text{path} \cdot (\ell - 1) \,/\, \max(L - 1,\, 1)$
            \hfill $\triangleright$ Linear drop path schedule
        \vspace{6pt}
        \STATE $(\gamma_1, \beta_1, \alpha_1, \gamma_2, \beta_2, \alpha_2)
                \leftarrow$ AdaLN$_\ell(\bm{c})$
            \hfill $\triangleright$ Alg.~\ref{alg:dit_mod}
        \vspace{6pt}
        \STATE $\bm{h}' \leftarrow \text{RMSNorm}_\ell(\bm{h}) \odot (1 + \gamma_1) + \beta_1$
            \hfill $\triangleright$ adaLN modulate
        \vspace{6pt}
        \STATE $\bm{h} \leftarrow \bm{h} + \text{DropPath}(\alpha_1 \odot \text{SparseAttn}_\ell(\bm{h}',\; \bm{b},\; G),\; p_\ell)$
            \hfill $\triangleright$ Alg.~\ref{alg:sparse_attn}
        \vspace{6pt}
        \STATE $\bm{h}' \leftarrow \text{RMSNorm}_\ell(\bm{h}) \odot (1 + \gamma_2) + \beta_2$
            \hfill $\triangleright$ adaLN modulate
        \vspace{6pt}
        \STATE $\bm{h} \leftarrow \bm{h} + \text{DropPath}(\alpha_2 \odot \text{SwiGLU}_\ell(\bm{h}'),\; p_\ell)$
            \hfill $\triangleright$ Alg.~\ref{alg:swiglu}
    \ENDFOR
    \vspace{6pt}
    \RETURN $\bm{h} \in \R^{N \times d}$
\end{algindent}
\end{algorithmic}
\end{algorithm}

\subsection{Sparse self-attention with pair bias}
\label{si:alg_attn}

Attention is computed only over edges in the sparse graph $G$. Following~\citet{wang2025simplefold}, we apply QK-normalisation (normalising queries and keys but not values). For each edge, the attention logit incorporates a learned per-head pair bias from the edge representations, and a sparse softmax over each node's neighbourhood aggregates weighted values. The output is scaled by a layer-dependent factor to stabilise the residual stream in deep networks.

\begin{algorithm}[H]
\caption{Sparse self-attention with per-head pair bias.\hfill\tokenrep\;$\leftarrow$\;\tokenrep\;\tokenpair\;\graphrep}
\label{alg:sparse_attn}
\begin{algorithmic}[1]
\STATE \textbf{def} SparseAttn($\bm{h} \in \R^{N \times d}$,
    $\bm{b} \in \R^{E \times H}$, $G \in \mathbb{N}^{E \times 2}$):
\begin{algindent}
    \vspace{6pt}
    \STATE $\bm{Q}, \bm{K}, \bm{V} \leftarrow \text{split}(\text{Linear}(\bm{h}),\; 3)$
        \hfill $\bm{Q}, \bm{K}, \bm{V} \in \R^{N \times H \times d_h}$
    \vspace{6pt}
    \STATE $\bm{Q} \leftarrow \text{RMSNorm}(\bm{Q})$;
           \quad $\bm{K} \leftarrow \text{RMSNorm}(\bm{K})$
        \hfill $\triangleright$ QK-norm only (not V)
    \vspace{6pt}
    \STATE $a_{ij} \leftarrow (\bm{Q}_i \cdot \bm{K}_j) \,/\, \sqrt{d_h} + \bm{b}_{ij}$
        \hfill $\forall\, (i, j) \in G$;\; $i$ = dst, $j$ = src
    \vspace{6pt}
    \STATE $w_{ij} \leftarrow \text{Softmax}_{j \in \mathcal{N}(i)}(a_{ij})$
        \hfill $\triangleright$ Normalise over src neighbours of dst $i$
    \vspace{6pt}
    \STATE $\bm{o}_i \leftarrow \sum_{j \in \mathcal{N}(i)} w_{ij} \, \bm{V}_j$
        \hfill $\triangleright$ Aggregate src values into dst
    \vspace{6pt}
    \STATE $\bm{h} \leftarrow \text{Dropout}(\text{Linear}(\bm{o}),\; p_\text{drop})$
        \hfill $\triangleright$ Output projection with dropout
    \vspace{6pt}
    \RETURN $\bm{h} \in \R^{N \times d}$
\end{algindent}
\end{algorithmic}
\end{algorithm}

\subsection{Gated cross-attention: downcast and upcast}
\label{si:alg_cross}

The downcast and upcast modules bridge between atom-level ($d_\text{atom}$) and token-level ($d_\text{tok}$) representations using gated cross-attention through the Rep14 mapping $\bm{M}$. In the \emph{downcast} (Algorithm~\ref{alg:downcast}), each token queries its 14 constituent atom representations, with a validity mask suppressing invalid (empty) positions in ligand and metal tokens, pooling atom-level information into a single token representation. In the \emph{upcast} (Algorithm~\ref{alg:upcast}), each of the 14 atom slots queries the token representation. To provide distinct information to each atom position, the token representations are expanded into $n_\text{dup}$ learned copies, so that each atom slot attends to a position-specific variant of the token representation. Both modules apply a sigmoid gate to the output to ensure conservative updates at the start of training.

\begin{algorithm}[H]
\caption{Downcast: atoms to tokens.\hfill\tokenrep\;$\leftarrow$\;\atomrep\;\tokenrep}
\label{alg:downcast}
\begin{algorithmic}[1]
\STATE \textbf{def} Downcast($\bm{h}^\text{atom} \in \R^{N_\text{atom} \times d_\text{atom}}$,
    $\bm{h}^\text{tok} \in \R^{N_\text{tok} \times d_\text{tok}}$,
    $\bm{M} \in \mathbb{N}^{3 \times N_\text{atom}}$):
\begin{algindent}
    \vspace{6pt}
    \STATE $\bm{A}, \bm{m}_\text{valid} \leftarrow$ AtomToToken$(\bm{h}^\text{atom}, \bm{M})$
        \hfill $\bm{A} \in \R^{N_\text{tok} \times 14 \times d_\text{atom}}$
    \vspace{6pt}
    \STATE $\bm{Q} \leftarrow \text{Linear}(\text{RMSNorm}(\bm{h}^\text{tok}))$
        \hfill $\in \R^{N_\text{tok} \times 1 \times H_\text{cross} \times d_h}$
    \vspace{6pt}
    \STATE $\bm{K}, \bm{V} \leftarrow \text{Linear}(\text{RMSNorm}(\bm{A}))$
        \hfill $\in \R^{N_\text{tok} \times 14 \times H_\text{cross} \times d_h}$
    \vspace{6pt}
    \STATE $\bm{a} \leftarrow (\bm{Q} \cdot \bm{K}^\top) / \sqrt{d_h}$;
           \quad $\bm{a}[\lnot\bm{m}_\text{valid}] \leftarrow -\infty$
    \vspace{6pt}
    \STATE $\bm{w} \leftarrow \text{Softmax}(\bm{a})$
    \vspace{6pt}
    \STATE $\bm{o} \leftarrow \bm{w} \cdot \bm{V}$
    \vspace{6pt}
    \STATE $g \leftarrow \mathrm{sigmoid}(\text{Linear}(\bm{o}))$
        \hfill $\triangleright$ Sigmoid gate
    \vspace{6pt}
    \STATE $\bm{h}^\text{tok} \leftarrow \bm{h}^\text{tok} + \text{Linear}(g \odot \bm{o})$
        \hfill $\triangleright$ Gated residual update
    \vspace{6pt}
    \RETURN $\bm{h}^\text{tok} \in \R^{N_\text{tok} \times d_\text{tok}}$
\end{algindent}
\end{algorithmic}
\end{algorithm}

\begin{algorithm}[H]
\caption{Upcast: tokens to atoms.\hfill\atomrep\;$\leftarrow$\;\atomrep\;\tokenrep}
\label{alg:upcast}
\begin{algorithmic}[1]
\STATE \textbf{def} Upcast($\bm{h}^\text{atom} \in \R^{N_\text{atom} \times d_\text{atom}}$,
    $\bm{h}^\text{tok} \in \R^{N_\text{tok} \times d_\text{tok}}$,
    $\bm{M} \in \mathbb{N}^{3 \times N_\text{atom}}$):
\begin{algindent}
    \vspace{6pt}
    \STATE $\bm{s} \leftarrow \text{Linear}(\text{RMSNorm}(\bm{h}^\text{tok}))$
        \hfill $\in \R^{N_\text{tok} \times n_\text{dup}}$;\; per-duplicate scalar bias
    \STATE $\bm{T} \leftarrow \bm{h}^\text{tok}[:, \text{None}, :] + \bm{s}[:, :, \text{None}]$
        \hfill $\in \R^{N_\text{tok} \times n_\text{dup} \times d_\text{tok}}$
    \vspace{6pt}
    \STATE $\bm{A}, \bm{m}_\text{valid} \leftarrow$ AtomToToken$(\bm{h}^\text{atom}, \bm{M})$
        \hfill $\bm{A} \in \R^{N_\text{tok} \times 14 \times d_\text{atom}}$
    \vspace{6pt}
    \STATE $\bm{Q} \leftarrow \text{Linear}(\text{RMSNorm}(\bm{A}))$
        \hfill $\in \R^{N_\text{tok} \times 14 \times H_\text{cross} \times d_h}$
    \vspace{6pt}
    \STATE $\bm{K}, \bm{V} \leftarrow \text{Linear}(\text{RMSNorm}(\bm{T}))$
        \hfill $\in \R^{N_\text{tok} \times n_\text{dup} \times H_\text{cross} \times d_h}$
    \vspace{6pt}
    \STATE $\bm{a} \leftarrow (\bm{Q} \cdot \bm{K}^\top) / \sqrt{d_h}$
    \vspace{6pt}
    \STATE $\bm{w} \leftarrow \text{Softmax}(\bm{a})$
    \vspace{6pt}
    \STATE $\bm{o} \leftarrow \bm{w} \cdot \bm{V}$;
           \quad $\bm{o}[\lnot\bm{m}_\text{valid}] \leftarrow \bm{0}$
    \vspace{6pt}
    \STATE $g \leftarrow \mathrm{sigmoid}(\text{Linear}(\bm{o}))$
    \vspace{6pt}
    \STATE $\bm{h}^\text{atom} \leftarrow \bm{h}^\text{atom} + \text{TokenToAtom}(\text{Linear}(g \odot \bm{o}),\; \bm{M})$
        \hfill $\triangleright$ Gated residual update
    \vspace{6pt}
    \RETURN $\bm{h}^\text{atom} \in \R^{N_\text{atom} \times d_\text{atom}}$
\end{algindent}
\end{algorithmic}
\end{algorithm}

\subsection{Output head}
\label{si:alg_output}

The output layer predicts a per-atom 3D velocity vector $\bm{v}_\theta$ from the atom-decoder representations using adaLN-Zero modulation, with both the modulation and output Linear layers zero-initialised so that the model predicts zero velocity at initialisation (identity function). After the output layer, the predicted velocity for all motif atoms is explicitly set to zero.

\begin{algorithm}[H]
\caption{adaLN-Zero output head. Predicts 3D velocity vectors.\hfill$\bm{v}$\;$\leftarrow$\;\atomrep\;\atomcond}
\label{alg:output}
\begin{algorithmic}[1]
\STATE \textbf{def} OutputHead($\bm{h} \in \R^{N_\text{atom} \times d_\text{atom}}$,
    $\bm{c} \in \R^{N_\text{atom} \times d_\text{atom}}$):
\begin{algindent}
    \vspace{6pt}
    \STATE $(\beta, \gamma) \leftarrow \text{chunk}(\text{Linear}(\text{SiLU}(\bm{c})),\; 2)$
        \hfill $\triangleright$ Zero-init Linear
    \vspace{6pt}
    \STATE $\bm{h} \leftarrow \text{RMSNorm}(\bm{h}) \odot (1 + \gamma) + \beta$
        \hfill $\triangleright$ adaLN modulate
    \vspace{6pt}
    \STATE $\bm{v} \leftarrow \text{Linear}(\bm{h})$
        \hfill $\triangleright$ Zero-init projection $d_\text{atom} \to 3$
    \vspace{6pt}
    \RETURN $\bm{v} \in \R^{N_\text{atom} \times 3}$
\end{algindent}
\end{algorithmic}
\end{algorithm}

\subsection{adaLN modulation and SwiGLU}
\label{si:alg_components}

The adaLN modulation (Algorithm~\ref{alg:dit_mod}) projects the conditioning signal $\bm{c}$ through a factorised bottleneck: a single Linear layer with SiLU activations on both sides ($\text{SiLU} \to \text{Linear}_{d \to d_\text{bn}} \to \text{SiLU}$) projects to an intermediate dimension $d_\text{bn} \ll d$ (preventing rank collapse), followed by a zero-initialised expansion to the six per-sub-layer scale, shift, and gate parameters. Gate parameters ($\alpha_1, \alpha_2$) are passed through a sigmoid activation, bounding them to $[0, 1]$ and stabilising residual stream growth. The SwiGLU feed-forward network (Algorithm~\ref{alg:swiglu}) follows~\citet{shazeer2020glu}: the hidden dimension is set to $\frac{2}{3}\cdot 4 d$ rounded up to a multiple of 256, and the output is the element-wise product of a SiLU-gated branch with a linear branch, with dropout applied on the gated hidden activations before the output projection.

\begin{algorithm}[H]
\caption{adaLN modulation with bottleneck and sigmoid gating. Generates shift, scale, and gate parameters.\hfill$(\gamma, \beta, \alpha)$\;$\leftarrow$\;\tokencond}
\label{alg:dit_mod}
\begin{algorithmic}[1]
\STATE \textbf{def} AdaLN($\bm{c} \in \R^{N \times d}$, $n_\text{params}{=}6$, $d_\text{bn}$):
\begin{algindent}
    \vspace{6pt}
    \STATE $\bm{z} \leftarrow \text{SiLU}(\text{Linear}(\text{SiLU}(\bm{c})))$
        \hfill $\in \R^{N \times d_\text{bn}}$;\; bottleneck projection
    \vspace{6pt}
    \STATE $\bm{p} \leftarrow \text{Linear}(\bm{z})$
        \hfill $\in \R^{N \times (n_\text{params} \cdot d)}$;\; zero-init
    \vspace{6pt}
    \STATE $(\gamma_1, \beta_1, \alpha_1, \gamma_2, \beta_2, \alpha_2)
            \leftarrow \text{chunk}(\bm{p},\; n_\text{params})$
        \hfill each $\in \R^{N \times d}$
    \vspace{6pt}
    \STATE $\alpha_1 \leftarrow \mathrm{sigmoid}(\alpha_1)$;
           \quad $\alpha_2 \leftarrow \mathrm{sigmoid}(\alpha_2)$
        \hfill $\triangleright$ Gate $\in [0, 1]$
    \vspace{6pt}
    \RETURN $(\gamma_1, \beta_1, \alpha_1, \gamma_2, \beta_2, \alpha_2)$
        \hfill $\triangleright$ scale, shift, gate $\times$ 2 (attn, MLP)
\end{algindent}
\end{algorithmic}
\end{algorithm}

\begin{algorithm}[H]
\caption{SwiGLU feed-forward network with dropout.\hfill\tokenrep\;$\leftarrow$\;\tokenrep}
\label{alg:swiglu}
\begin{algorithmic}[1]
\STATE \textbf{def} SwiGLU($\bm{h} \in \R^{N \times d}$):
\begin{algindent}
    \vspace{6pt}
    \STATE $d_\text{hidden} \leftarrow \lceil \tfrac{2}{3} \cdot 4d \,/\, 256 \rceil \cdot 256$
        \hfill $\triangleright$ Rounded to multiple of 256
    \vspace{6pt}
    \STATE $\bm{g} \leftarrow \text{SiLU}(\bm{W}_1 \bm{h})$;
           \quad $\bm{u} \leftarrow \bm{W}_3 \bm{h}$
        \hfill $\bm{g}, \bm{u} \in \R^{N \times d_\text{hidden}}$
    \vspace{6pt}
    \STATE $\bm{h} \leftarrow \text{Dropout}(\bm{g} \odot \bm{u},\; p_\text{drop})$
        \hfill $\triangleright$ Dropout on gated hidden
    \vspace{6pt}
    \RETURN $\bm{W}_2 \bm{h} \in \R^{N \times d}$
\end{algindent}
\end{algorithmic}
\end{algorithm}

\subsection{Feature embedding}
\label{si:feature_embedding}

All categorical features, both per-atom and per-token, are embedded through a lightweight feature embedding. Each feature is independently embedded and the resulting vectors are concatenated and projected through a bottleneck MLP whose intermediate dimension is substantially smaller than the output dimension. This bottleneck prevents rank collapse in the embedding space, a phenomenon we identified through spectral analysis of trained models: without the bottleneck, learned embeddings occupy a low-rank subspace that limits the effective capacity of downstream attention layers. We demonstrate the impact of this design choice via spectral analysis in \S\ref{sec:res_weight_quality}.

\subsection{Token embedding}
\label{si:alg_embeddings}

Token-level representations are initialised from three sources: (1)~per-token categorical features (residue type, terminus status, motif membership, secondary structure, and residue/HETATM flags) processed through the feature embedding; (2)~per-structure global features (secondary-structure ratios, radius of gyration, amino acid group ratios, centering mode) embedded and broadcast to every token; and (3)~a Fourier coordinate embedding of the token's atom positions, averaged over valid atom slots. These three sources are summed and normalised to produce the initial token representations, which also serve as the token conditioning signal. Token-level pair representations combine a relative position embedding (encoding the signed residue-index difference along the chain) and a chemical bond embedding (encoding the bond type along each edge that represents a bond). These are projected to produce a per-head pair bias along the sparse edges.

\begin{algorithm}[H]
\caption{Token embedding.\hfill\tokenrep\;\tokenpair\;\tokencond\;$\leftarrow$\;$\bm{x}_t, \bm{c}$}
\label{alg:token_embed}
\begin{algorithmic}[1]
\STATE \textbf{def} TokenEmbed($\bm{x}_t \in \R^{N_\text{atom} \times 3}$,
    $\bm{c}$, $G^\text{tok} \in \mathbb{N}^{E_\text{tok} \times 2}$):
\begin{algindent}
    \vspace{6pt}
    \STATE $\bm{f}_\text{coord} \leftarrow$ FourierProj$(\bm{x}_t,\; \bm{M},\; \text{pool over 14 atoms})$
        \hfill $\in \R^{N_\text{tok} \times d_\text{tok}}$
    \vspace{6pt}
    \STATE $\bm{f}_\text{cat} \leftarrow$ FeatureEmbed$(\bm{c}_\text{tok})$
        \hfill $\triangleright$ token type, termini, is\_motif, is\_residue, SS
    \vspace{6pt}
    \STATE $\bm{f}_\text{graph} \leftarrow$ FeatureEmbed$(\bm{c}_\text{graph})[\text{batch\_idx}]$
        \hfill $\triangleright$ SS ratios, $R_g$, surface/core ratios
    \vspace{6pt}
    \STATE $\bm{h}^\text{tok} \leftarrow \text{RMSNorm}(\bm{f}_\text{coord} + \bm{f}_\text{cat} + \bm{f}_\text{graph})$
    \vspace{6pt}
    \STATE $\bm{c}^\text{tok} \leftarrow \bm{h}^\text{tok}$
        \hfill $\triangleright$ Conditioning snapshot before pair bias
    \vspace{6pt}
    \STATE $\bm{e}_\text{pos} \leftarrow$ RelPosEmbed$(\text{residue indices},\; G^\text{tok})$
        \hfill $\in \R^{E_\text{tok} \times d_\text{edge}}$
    \vspace{6pt}
    \STATE $\bm{e}_\text{bond} \leftarrow$ BondEmbed$(\text{bonds},\; G^\text{tok})$
        \hfill $\in \R^{E_\text{tok} \times d_\text{edge}}$
    \vspace{6pt}
    \STATE $\bm{b}^\text{tok} \leftarrow \text{Linear}((\bm{e}_\text{pos} + \bm{e}_\text{bond}) / 2)$
        \hfill $\in \R^{E_\text{tok} \times H_\text{tok}}$
    \vspace{6pt}
    \RETURN $\bm{h}^\text{tok} \in \R^{N_\text{tok} \times d_\text{tok}}$,\;
            $\bm{b}^\text{tok} \in \R^{E_\text{tok} \times H_\text{tok}}$,\;
            $\bm{c}^\text{tok} \in \R^{N_\text{tok} \times d_\text{tok}}$
\end{algindent}
\end{algorithmic}
\end{algorithm}

\subsection{Atom embedding}
\label{si:atom_embedding}

Atom-level representations are built from per-atom categorical features (element type, atom index, RASA, terminus, motif, residue/HETATM flags) via the feature embedding, combined with a Fourier coordinate embedding. Token-level context is injected by broadcasting token representations to their constituent atoms via the Rep14 mapping. Atom-level pair biases are derived from the token-level pair bias by mapping token edges to atom edges through the Rep14 representation.

\begin{algorithm}[H]
\caption{Atom embedding.\newline\mbox{}\hfill\atomrep\;\atompair\;\atomcond\;$\leftarrow$\;$\bm{x}_t, \bm{c}$\;\tokenrep\;\tokenpair}
\label{alg:atom_embed}
\begin{algorithmic}[1]
\STATE \textbf{def} AtomEmbed(
    $\bm{x}_t \in \R^{N_\text{atom} \times 3}$,
    $\bm{c}$,
    $\bm{h}^\text{tok} \in \R^{N_\text{tok} \times d_\text{tok}}$,
    $\bm{b}^\text{tok} \in \R^{E_\text{tok} \times H_\text{tok}}$,
    $G \in \mathbb{N}^{E \times 2}$):
\begin{algindent}
    \vspace{6pt}
    \STATE $\bm{f}_\text{coord} \leftarrow$ FourierProj$(\bm{x}_t)$
        \hfill $\in \R^{N_\text{atom} \times d_\text{atom}}$
    \vspace{6pt}
    \STATE $\bm{f}_\text{cat} \leftarrow$ FeatureEmbed$(\bm{c}_\text{atom})$
        \hfill $\triangleright$ element, atom name, RASA, termini, is\_motif
    \vspace{6pt}
    \STATE $\bm{h}^\text{atom} \leftarrow \bm{f}_\text{cat} + \text{Linear}(\bm{f}_\text{cat}) + \bm{f}_\text{coord}$
    \vspace{6pt}
    \STATE $\bm{h}^{\text{tok} \to \text{atom}} \leftarrow$ TokenToAtom$(\bm{h}^\text{tok},\; \bm{M})$
        \hfill $\triangleright$ Broadcast token representations to atoms
    \vspace{6pt}
    \STATE $\bm{h}^\text{atom} \leftarrow \text{RMSNorm}(\bm{h}^\text{atom}
        + \text{Linear}(\bm{h}^{\text{tok} \to \text{atom}}))$
    \vspace{6pt}
    \STATE $\bm{c}^\text{atom} \leftarrow \bm{h}^\text{atom}$
    \vspace{6pt}
    \STATE $\bm{b}^{\text{tok} \to \text{atom}} \leftarrow$ EdgeTokenToAtom$(\bm{b}^\text{tok},\;
        G,\; \bm{M})$
        \hfill $\in \R^{E_\text{atom} \times H_\text{tok}}$
    \vspace{6pt}
    \STATE $\bm{b}^\text{atom} \leftarrow \text{Linear}(\bm{b}^{\text{tok} \to \text{atom}})$
        \hfill $\in \R^{E_\text{atom} \times H_\text{atom}}$
    \vspace{6pt}
    \RETURN $\bm{h}^\text{atom} \in \R^{N_\text{atom} \times d_\text{atom}}$,\;
            $\bm{b}^\text{atom} \in \R^{E_\text{atom} \times H_\text{atom}}$,\;
            $\bm{c}^\text{atom} \in \R^{N_\text{atom} \times d_\text{atom}}$
\end{algindent}
\end{algorithmic}
\end{algorithm}

\subsection{Time embedding}
\label{si:alg_time}

A sinusoidal embedding with log-spaced frequencies encodes the flow time $t$ for each atom. Motif atoms always receive $t = 1$ regardless of the current timestep, encoding the information that their coordinates are already at the target. The time embedding is combined with the representation-based conditioning (via averaging in $\mathrm{Embed}$, Algorithm~\ref{alg:init_embed}) to form the final conditioning signal $\bm{c}$ that drives the adaptive layer normalisation throughout the network.

\begin{algorithm}[H]
\caption{Sinusoidal time embedding. Motif atoms always see $t{=}1$.\hfill$\bm{t}^\text{atom},\,\bm{t}^\text{tok}$\;$\leftarrow$\;$t$}
\label{alg:time_embed}
\begin{algorithmic}[1]
\STATE \textbf{def} TimeEmbed($t \in [0,1]$, $\bm{c}$):
\begin{algindent}
    \vspace{6pt}
    \STATE $\bm{\omega} \leftarrow \text{logspace}(2\pi,\; 128\pi,\; d_\text{time}/2)$
        \hfill $\triangleright$ Log-spaced frequencies
    \vspace{6pt}
    \STATE $t_\ell \leftarrow t[\text{batch\_idx}]$;
           \quad $t_\ell[\text{is\_motif}] \leftarrow 1$
        \hfill $\triangleright$ Motif atoms always at $t{=}1$
    \vspace{6pt}
    \STATE $\bm{e} \leftarrow [\sin(t_\ell \cdot \bm{\omega}),\;
           \cos(t_\ell \cdot \bm{\omega})]$
        \hfill $\in \R^{N_\text{atom} \times d_\text{time}}$
    \vspace{6pt}
    \STATE $\bm{t}^\text{atom} \leftarrow \text{RMSNorm}(\text{Linear}(\text{Linear}(\bm{e})))$
        \hfill $\in \R^{N_\text{atom} \times d_\text{atom}}$
    \vspace{6pt}
    \STATE $\bm{t}^\text{tok} \leftarrow \text{RMSNorm}(\text{Linear}(\text{Linear}(\bm{e}^\text{tok})))$
        \hfill $\triangleright$ Same construction at token level
    \vspace{6pt}
    \RETURN $\bm{t}^\text{atom} \in \R^{N_\text{atom} \times d_\text{atom}}$,\;
            $\bm{t}^\text{tok} \in \R^{N_\text{tok} \times d_\text{tok}}$
\end{algindent}
\end{algorithmic}
\end{algorithm}

\subsection{Edge construction algorithm}
\label{si:alg_edges}

$\mathrm{EdgeModule}$ (Algorithm~\ref{alg:edges}) implements the priority scheme described in \S\ref{si:features_edges}. Token-level distances are computed as the minimum over all $14 \times 14$ atom pairs between two tokens. At the atom level, each atom connects to its $\pm 1$ sequence neighbours and up to 128 total edges (including 32 ligand $k$-NN); at the token level, each token connects to $\pm 32$ sequence neighbours and up to 128 total edges (including 32 ligand $k$-NN), with all tokens connected to motif tokens when present. The edge list is sorted by destination node and a CSR index pointer is built for use by the sparse attention kernels.

\begin{algorithm}[H]
\caption{Sparse edge construction with priority ordering.\hfill$G^\text{atom},\, G^\text{tok}$\;$\leftarrow$\;$\bm{x}_t, \bm{c}$}
\label{alg:edges}
\begin{algorithmic}[1]
\STATE \textbf{def} EdgeModule($\bm{x}_t \in \R^{N_\text{atom} \times 3}$, $\bm{c}$):
\begin{algindent}
    \FOR{level $\in \{\text{atom}, \text{token}\}$}
        \vspace{6pt}
        \STATE $G_\text{seq} \leftarrow \{(i,j) : |\text{resi}_i - \text{resi}_j| \leq n_\text{seq},\;
            \text{chain}_i = \text{chain}_j\}$
            \hfill $\triangleright$ Sequence neighbours
        \vspace{6pt}
        \STATE $G_\text{bond} \leftarrow$ chemical bonds from $\bm{c}$
        \vspace{6pt}
        \STATE $G_\text{lig} \leftarrow$ k-NN among ligand atoms
        \vspace{6pt}
        \STATE $G_\text{motif} \leftarrow \{(i,j) : \text{is\_motif}_j = \text{true}\}$
            \hfill $\triangleright$ Token level only
        \vspace{6pt}
        \STATE $G_\text{prev}(i) \leftarrow$ per-node degree from $G_\text{seq} \cup G_\text{bond} \cup G_\text{lig} \cup G_\text{motif}$
        \STATE $G_\text{knn}(i) \leftarrow$ k-NN$_{n_\text{budget} - |G_\text{prev}(i)|}(\bm{x}_t,\; i)$
            \hfill $\triangleright$ Per-node fill to edge budget
        \vspace{6pt}
        \STATE $G^\text{level} \leftarrow G_\text{seq} \cup G_\text{bond}
            \cup G_\text{lig} \cup G_\text{motif} \cup G_\text{knn}$
            \hfill $\triangleright$ Priority-ordered union
    \ENDFOR
    \vspace{6pt}
    \RETURN $G^\text{atom} \in \mathbb{N}^{E_\text{atom} \times 2}$,\;
            $G^\text{tok} \in \mathbb{N}^{E_\text{tok} \times 2}$
\end{algindent}
\end{algorithmic}
\end{algorithm}

\subsection{Recycling and distogram embedding}
\label{si:alg_distogram}

The full trunk is executed $N_r + 1$ times, with gradients carried only by the final iteration. After each pass, the predicted endpoint $\hat{\bm{x}}$ is used to construct a C$_\alpha$ distogram from pairwise distances between C$_\alpha$ atoms along the token-level sparse edges. This distogram is embedded and added to the token-level pair bias on the next pass, providing geometric feedback from the previous prediction (Algorithm~\ref{alg:distogram}). The previous iteration's atom and token node representations are similarly fed back into the initial embeddings through zero-initialised Linear projections (see the recycling block at the start of each iteration in Algorithm~\ref{alg:forward_pass}). On the first pass, all recycling contributions are zeroed out. During training, the number of recycling iterations is sampled uniformly from $\{0, 1, \ldots, N_r\}$ for regularisation. At inference, we use $N_r = 2$; ablation of the recycling depth is left for future work.

\begin{algorithm}[H]
\caption{Distogram token embedding for recycling.\hfill\tokenpair\;$\leftarrow$\;$\hat{\bm{x}}$}
\label{alg:distogram}
\begin{algorithmic}[1]
\STATE \textbf{def} DistogramEmbed(
    $\hat{\bm{x}} \in \R^{N_\text{atom} \times 3}$,
    $\bm{M} \in \mathbb{N}^{3 \times N_\text{atom}}$,
    $G^\text{tok} \in \mathbb{N}^{E_\text{tok} \times 2}$):
\begin{algindent}
    \vspace{6pt}
    \STATE $\bm{x}_\text{CA} \leftarrow$ AtomToToken$(\hat{\bm{x}}, \bm{M})[:, 1, :]$
        \hfill $\triangleright$ C$_\alpha$ from Rep14 position 1
    \vspace{6pt}
    \STATE $d_{ij} \leftarrow \|\bm{x}_{\text{CA},i} - \bm{x}_{\text{CA},j}\|_2$
        \hfill $\forall\, (i,j) \in G^\text{tok}$
    \vspace{6pt}
    \STATE $\bm{b}_\text{idx} \leftarrow \text{bucketize}(d_{ij},\; \text{boundaries})$
        \hfill $\triangleright$ $n_\text{bins}{=}65$ bins from $1/\sigma_\text{data}$--$30/\sigma_\text{data}$
    \vspace{6pt}
    \STATE $\bm{b}^\text{tok} \leftarrow \text{Embed}(\bm{b}_\text{idx})$
        \hfill $\in \R^{E_\text{tok} \times H_\text{tok}}$
    \vspace{6pt}
    \RETURN $\bm{b}^\text{tok} \in \R^{E_\text{tok} \times H_\text{tok}}$
\end{algindent}
\end{algorithmic}
\end{algorithm}

\subsection{Fourier coordinate projection}
\label{si:alg_fourier}

Coordinate representations are injected into both token and atom embeddings via a Fourier projection (Algorithm~\ref{alg:fourier}): each coordinate is multiplied by a bank of log-spaced frequencies (wavelengths from $0.1/\sigma_\text{data}$ to $50/\sigma_\text{data}$), concatenated with its sines and cosines and the raw value, and mapped through a two-layer MLP to the target embedding dimension. For token embeddings, coordinates are first pooled over the 14 Rep14 atom slots with the validity mask.

\begin{algorithm}[H]
\caption{Fourier coordinate projection.\hfill\tokenrep\;$\leftarrow$\;$\bm{x}_t$}
\label{alg:fourier}
\begin{algorithmic}[1]
\STATE \textbf{def} FourierProj($\bm{x} \in \R^{N \times 3}$, $d_\text{out}$):
\begin{algindent}
    \vspace{6pt}
    \STATE $\bm{\omega} \leftarrow \text{logspace}(2\pi\sigma_\text{data}/50,\; 2\pi\sigma_\text{data}/0.1,\; n_\text{freq})$
        \hfill $\triangleright$ $\sigma_\text{data} = 10$
    \vspace{6pt}
    \STATE $\bm{f} \leftarrow \bm{x} \cdot \bm{\omega}^\top$
        \hfill $\in \R^{N \times 3 \times n_\text{freq}}$
    \vspace{6pt}
    \STATE $\bm{f} \leftarrow [\sin(\bm{f}),\; \cos(\bm{f})]$
        \hfill $\in \R^{N \times 6 n_\text{freq}}$
    \vspace{6pt}
    \STATE $\bm{f} \leftarrow [\bm{f},\; \bm{x}]$
        \hfill $\triangleright$ Concatenate raw coordinates
    \vspace{6pt}
    \STATE $\bm{h} \leftarrow \text{MLP}(\bm{f})$
        \hfill $\triangleright$ Two-layer projection
    \vspace{6pt}
    \RETURN $\bm{h} \in \R^{N \times d_\text{out}}$
\end{algindent}
\end{algorithmic}
\end{algorithm}

\end{document}